\documentclass[10pt,twocolumn,letterpaper]{article}

\usepackage[pagenumbers]{cvpr} %

\usepackage[utf8]{inputenc} %
\usepackage[T1]{fontenc}    %
\usepackage{url}            %
\usepackage{booktabs}       %
\usepackage{amsfonts}       %
\usepackage{nicefrac}       %
\usepackage{microtype}      %
\usepackage[dvipsnames]{xcolor}

\usepackage{multirow}
\usepackage{tikz}
\usepackage{caption}
\usepackage{graphicx}
\usepackage{amsmath}
\usepackage{amssymb}
\usepackage{booktabs}
\usepackage{siunitx}
\usepackage{enumitem}

\usepackage{tcolorbox}
\usepackage{makecell}
\usepackage{graphbox}

\newcommand\todocite[1]{\textcolor{red}{[Citation needed]}}

\usepackage[accsupp]{axessibility} %

\definecolor{cvprblue}{rgb}{0.21,0.49,0.74}
\usepackage[pagebackref,breaklinks,colorlinks,citecolor=cvprblue]{hyperref}

\def\paperID{10239} %
\def\confName{CVPR}
\def\confYear{2024}

\title{\vspace{-8pt}Physical Property Understanding from Language-Embedded Feature Fields}

\author{Albert J. Zhai ~~~~ Yuan Shen ~~~~ Emily Y. Chen ~~~~  Gloria X. Wang ~~~~ Xinlei Wang \\  Sheng Wang ~~~~  Kaiyu Guan ~~~~  Shenlong Wang\\
{University of Illinois at Urbana-Champaign}
}

\begin{document}

\maketitle
\begin{abstract}
Can computers perceive the physical properties of objects solely through vision? Research in cognitive science and vision science has shown that humans excel at identifying materials and estimating their physical properties based purely on visual appearance. In this paper, we present a novel approach for dense prediction of the physical properties of objects using a collection of images. Inspired by how humans reason about physics through vision, we leverage large language models to propose candidate materials for each object. We then construct a language-embedded point cloud and estimate the physical properties of each 3D point using a zero-shot kernel regression approach. Our method is accurate, annotation-free, and applicable to any object in the open world. Experiments demonstrate the effectiveness of the proposed approach in various physical property reasoning tasks, such as estimating the mass of common objects, as well as other properties like friction and hardness. 
Code is available at {\url{https://ajzhai.github.io/NeRF2Physics}}.
\end{abstract}
\section{Introduction}

Imagine that you are shopping in a home improvement store. Even though there is a huge variety of tools and furniture items in the store, you can most likely make a reasonable estimate of how heavy most of the objects are in a glance. Now imagine that you are hiking in a forest and trying to cross a stream by stepping on some stones in the water. Simply by looking at the stones, you are probably able to identify which stones have enough friction to walk on and which stones would cause you to slip and fall into the water.

Humans are remarkably adept at predicting physical properties of objects based on visual information. Research in cognitive science and human vision has shown that humans make such predictions by associating visual appearances with materials that we have encountered before and have rich, grounded knowledge about \cite{FLEMING201462, fleming2013perceptual}.

\begin{figure}[t]
\centering
\includegraphics[width=\linewidth,  clip]{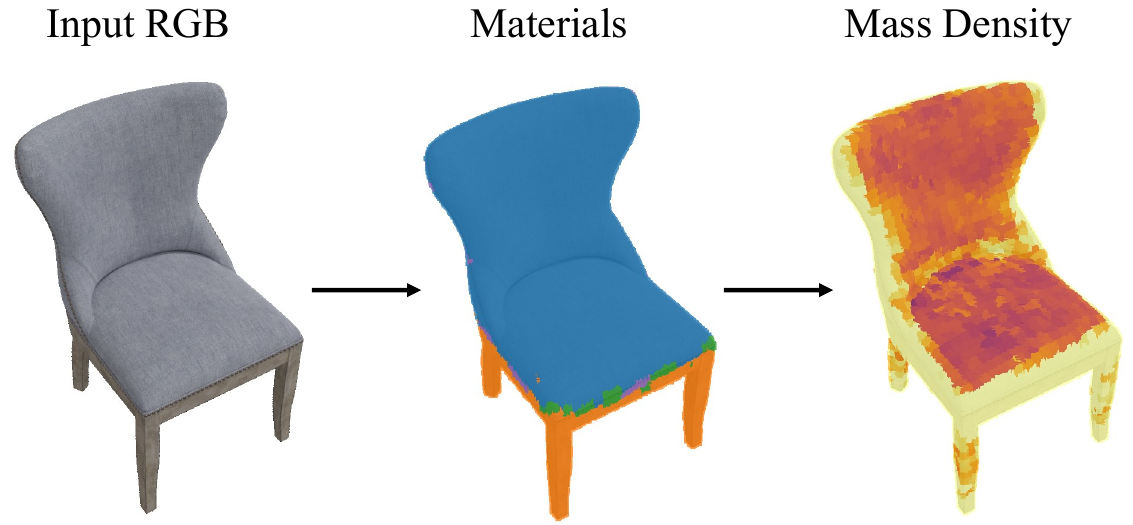}
\caption{\textbf{Estimating physical properties from images.} Humans can predict physical properties of objects by associating visual appearances with grounded knowledge about materials. We propose to equip computers with this capability by combining language-embedded feature fields with LLM-based material reasoning.}
\vspace{-10pt}
\label{fig:teaser}
\end{figure}

It is highly desirable for computers to be equipped with similar or even better capabilities of material perception. Having computational models for perceiving physics from visual data is crucial for various applications, including robotics, agriculture, urban planning, graphics, and many other domains. Nevertheless, challenges remain. One is the difficulty of acquiring labeled ground-truth data. For example, consider the endeavor of measuring the mass of a tree, or measuring the thermal conductivity densely throughout a coffee machine. Another is the highly uncertain nature of the prediction task due to having limited observations -- there is simply no way to know with certainty what is in the interior of an object without additional information.

\begin{figure*}[!t]
\centering
\includegraphics[width=\linewidth,  clip]{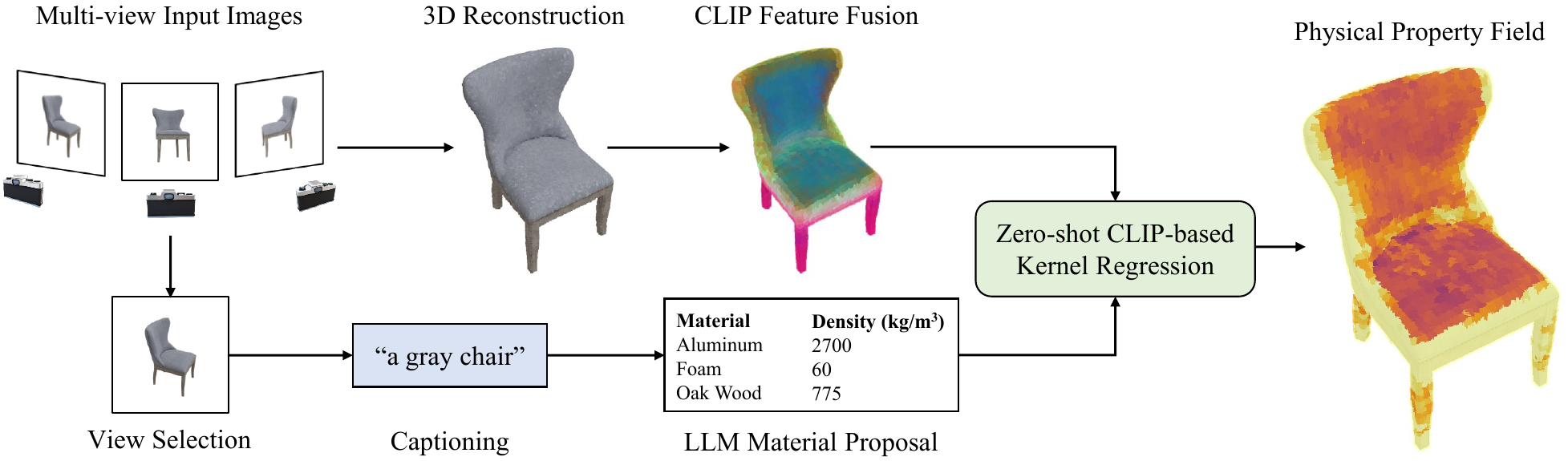}
\caption{\textbf{Overview of NeRF2Physics.} Given a collection of posed images, we first train a neural radiance field to capture the 3D geometry of the scene. Then, we fuse vision-language features into a point cloud extracted from the field. Next, we use a captioning model to provide a text description of the scene and prompt an LLM to produce a dictionary of possible materials in the scene, along with their physical properties. From here, physical properties can be estimated at any query point using zero-shot CLIP-based kernel regression within the dictionary. The kernel regression process is illustrated in more detail in Fig.~\ref{fig:zeroshot_overview}.}
\vspace{-10pt}
\label{fig:full_overview}
\end{figure*}

 This paper presents a training-free approach towards uncertainty-aware dense prediction of physical properties from a collection of images. Our method, named NeRF2Physics, integrates object-level semantic reasoning with point-level appearance reasoning. First, we use a neural radiance field to extract a set of 3D points on the object's surface and fuse 2D vision-language features into each point. Then, inspired by how humans reason about physics through vision, we draw upon the semantic knowledge contained within large language models to obtain a set of candidate materials for each object.  Finally, we estimate the physical properties of each point using a zero-shot retrieval-based approach and propagate the estimates across the entire object via spatial interpolation. Our method is accurate, annotation-free, and applicable to any object in the open world. 

We evaluate NeRF2Physics on the task of mass estimation using the ABO dataset, as well as our own dataset of real-world objects with manually measured friction and hardness values. The results demonstrate that our method outperforms other zero-shot baselines and even supervised baselines for mass estimation. Visualizations of the predicted fields show that our approach can produce reasonable predictions of a variety of physical properties without supervision. In our ablation study, we compare our method with alternative approaches for language-driven physical property estimation.

\section{Related Work}

\paragraph{Visual physics reasoning.}
Reasoning about physical properties from visual data is a longstanding problem~\cite{adelson2001seeing, NIPS2015_d09bf415, fragkiadaki2015learning}. Studies have shown that deep learning models can potentially carry similar physical reasoning capabilities as humans, including estimation of object mass, friction, electric conductivity, and other material properties~\cite{wu2016physics, NIPS2015_d09bf415, yildiriminterpreting, varma2008statistical}. Most of the prior work focuses on dynamic reasoning of object properties by either observing target dynamics~\cite{NIPS2015_d09bf415, yildiriminterpreting, li2022pac} or directly interacting with the target in a 3D physical engine~\cite{pinto2016curious, yaoestimating}. A few studies have also tackled estimating material properties from static images directly~\cite{pmlr-v78-standley17a, bell14intrinsic, bell2015material, sharan2013recognizing}. Although promising, existing work mostly tackles specific types of material properties, e.g., mass or tenderness, by collecting corresponding task-dependent data. In contrast, our method can generate diverse physical properties like mass density, friction, and hardness in a zero-shot manner from a single language-embedded feature field. Our work is greatly inspired by pioneering work from vision and cognitive science. Studies have found that humans are good at recognizing many material properties, e.g., thermal conductivity and hardness, from only visual inputs, even with only a brief demonstration~\cite{FLEMING201462, sharan2009material, fleming2013perceptual}. Our work seeks to empower computational models with such perception capabilities of recognizing a diverse range of material properties.

\paragraph{Language grounding.}
CLIP is a large pre-trained vision-language model that can efficiently learn visual concepts from natural language supervision~\cite{radford2021learning}. For training, CLIP jointly trains its image and text encoders to predict the correct pairings of image and text examples in a self-supervised fashion. Due to its success in capturing diverse visual concepts, CLIP has been widely used for zero-shot and few-shot tasks, spanning many applications from scene understanding~\cite{peng2022openscene, shafiullah2022clip, jatavallabhula2023conceptfusion,  chen2023open, lerf2023} to texture generation~\cite{michel2022text2mesh} to language-grounded reasoning~\cite{gadre2022clip, wu2022casa, li2022languagedriven, lerftogo2023, hong20233d}. The models are trained via contrastive learning with a large amount of data and are thus capable of associating meaningful language-driven semantics with image patches. Our work leverages CLIP embeddings of small patches to provide a solid foundation for physical property reasoning in a zero-shot manner.

\section{Neural Physical Property Fields}

\begin{figure*}[!t]
\centering
\includegraphics[width=\linewidth,  clip]{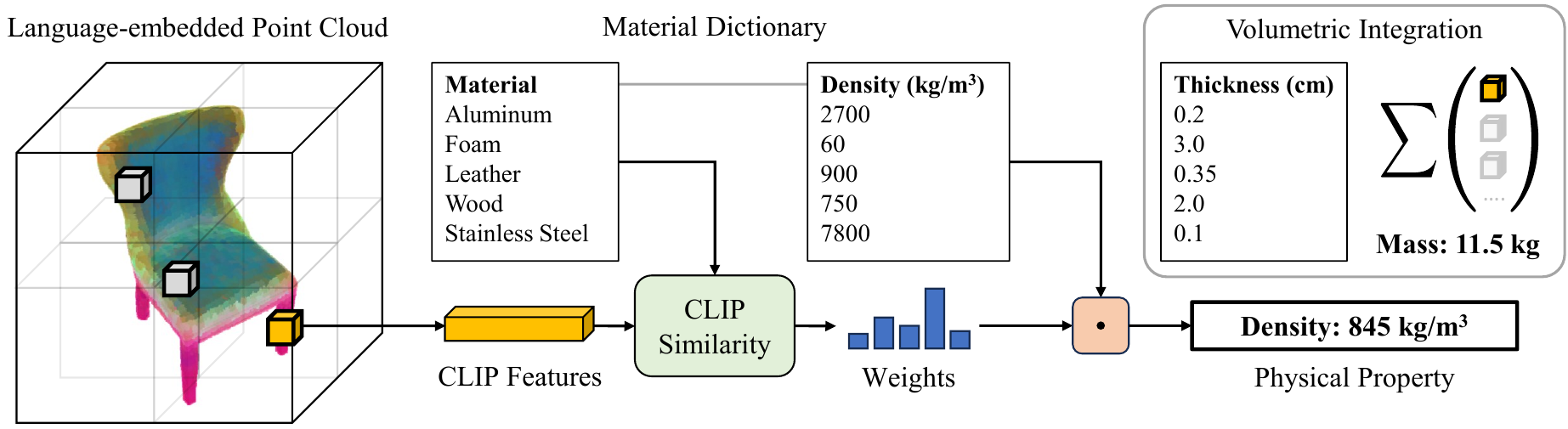}

\caption{\textbf{Overview of zero-shot physical property prediction.} To predict physical property values from the language-embedded point cloud, we extract CLIP features and perform kernel regression using the predicted dictionary of materials and their properties. To predict the total mass of an object, we then integrate the predicted mass density across cuboids on the surface of the object. The thickness of each cuboid is estimated in the same way as the other physical properties. }
\vspace{-12pt}
\label{fig:zeroshot_overview}
\end{figure*}

\subsection{Overview}

Our method takes as input a collection of posed images $\mathcal{I}$ and produces a physical property field $\rho(\mathbf{x})$ that can be queried to obtain physical property estimates at any occupied point within the scene.
In this work, we focus on single-object scenes, but our approach can be extended to multiple-object scenes as long as segmentation masks are available. 
Inspired by how humans perceive and reason about physical properties of the objects they encounter, we propose to leverage language-vision embeddings as well as large language models (LLMs)
to achieve this goal.  Fig.~\ref{fig:full_overview} depicts an overview of our approach.
First, we build a language-embedded point cloud
from which per-point semantic features can be queried (Sec.~\ref{sec:feature_field}). We then prompt an LLM to propose a dictionary $\mathcal{M}$ of candidate materials based on object semantics and apply zero-shot CLIP-based retrieval for reasoning about physical properties based on $\mathcal{M}$ (Sec.~\ref{sec:physics_reasoning}). Finally, for physical properties that require volumetric integration, such as mass, we propose an integration method using LLM-based estimates of surface thickness (Sec.~\ref{sec:integration}).

\subsection{Language-embedded point cloud} \label{sec:feature_field}

Accurate estimation of physical properties, such as object mass, requires both accurate geometric and material understanding of the scene. To capture the 3D geometry of a scene, we train a neural radiance field~\cite{mildenhall2020nerf} and extract a set of 3D ``source'' points via depth rendering.
We then fuse per-patch CLIP~\cite{radford2021learning} features into the source points using a simple visibility-aware averaging scheme.

\paragraph{Neural radiance field.}
For each scene, we first train a neural radiance field (NeRF)~\cite{mildenhall2020nerf} on the given images and camera poses. We choose to use NeRF because it tends to give higher-quality, more robust depth maps compared to other 3D reconstruction methods, especially for reflective surfaces. 
We directly use the Nerfacto method from Nerfstudio~\cite{tancik2023nerfstudio}, which combines several state-of-the-art techniques for improving performance. Once the NeRF is trained, we randomly sample a total of $N$ rays from the input views and estimate the depth along each ray via median depth rendering~\cite{tancik2023nerfstudio}. Finally, we convert the depths into 3D points and perform voxel down-sampling to remove redundant points. The result is a point cloud $\mathcal{S} \subset \mathbb{R}^3$, where $|\mathcal{S}| \leq N$. We will refer to these points, which should cover all of the visible surfaces in the scene, as the \textit{source points} for the scene.

\paragraph{3D language feature fusion.\label{sec_method:feature_fusion}}
CLIP features have been shown to perform well in several zero-shot image classification tasks~\cite{radford2021learning, novack2023chils}, giving reason to believe that they can be successfully applied for material recognition. In order to enable 3D reasoning based on CLIP, one must aggregate CLIP features within a 3D representation of the scene. A number of works~\cite{peng2022openscene, shafiullah2022clip, jatavallabhula2023conceptfusion,  chen2023open} have proposed methods for 3D fusion of CLIP features that are conducive to object-level segmentation due to the use of object-level region proposals. However, these methods are not well-suited for discriminating between different materials \textit{within the same object}, which is required for our use case. 

In this work, we use simple averaging to fuse CLIP embeddings of small patches in the input images, which usually contain enough appearance information to discriminate between different materials. 
For each source point $\mathbf{s} \in \mathcal{S}$ and input image $\mathbf{I} \in \mathcal{I}$, we determine the pixel coordinates $(u, v)$ of the point projected using the camera parameters of the image. We then test for occlusion using the NeRF-estimated depth to determine if the point is visible in the image. 
If it is visible, we extract a patch of size $P \times P$ centered at $(u, v)$, and apply a CLIP image encoder to obtain a 512-dimensional feature vector for that patch. 
If segmentation masks are available, they can be applied here to focus on an object of interest. Once this is done for all of the input images and source points, the patch features are average-pooled to create a fused feature vector $\mathbf{z}$ for each source point $\mathbf{s}$.

\subsection{Physical property reasoning} \label{sec:physics_reasoning}
Our language-embedded point cloud contains rich semantic features for every source point in the 3D scene. Such features are usually tightly related to the physical properties of the object(s) in the scene (see Fig.~\ref{fig:example_results}).
In this section, we propose a two-stage approach for estimating physical property values from the semantic features of the source points, which can then be propagated to any point in the continuous space by spatial interpolation. In the first stage, we prompt a VQA model to propose a dictionary of candidate materials along with their physical properties based on the input images. In the second stage, we perform a kernel regression over the materials in the dictionary for each source point using CLIP similarity in a zero-shot manner. Formally, for a given point $\mathbf{s}$ with semantic embedding $\mathbf{z}$, we formulate its physical property as 
\begin{equation}
    \rho(\mathbf{s}) = F(\mathbf{z}, \mathcal{M}) = F(\mathbf{z}, {G(\mathcal{I})}),
\end{equation}
where $G$ is the mapping from the input images $\mathcal{I}$ to candidate materials $\mathcal{M}$, and $F$ is the mapping from semantic features and candidate materials to our desired physical property.

\paragraph{LLM-based material proposal.}
The set of different materials that exist in the world is extremely large and difficult to define. Furthermore, many materials look identical and thus cannot be distinguished by local appearance alone. Despite this, humans are able to guess the material composition of objects through high-level reasoning about object semantics on top of low-level appearance cues. Inspired by this, our method calls upon LLMs for open-vocabulary semantic reasoning about materials and their physical properties.

Given a set of input images $\mathcal{I}$, we first select a canonical view $\mathbf{I}_0 \in \mathcal{I}$. If segmentation masks are available, we calculate the area of the mask in each frame and select the frame at the 75th percentile (as a heuristic method to obtain an informative view). If masks are not available, we select a view uniformly at random. We then use a VQA model (BLIP-2~\cite{li2023blip}) to produce a text description of $\mathbf{I}_0$. Finally, we pass this description to an LLM (GPT 3.5) and prompt it to return a dictionary of $K$ candidate materials  $\mathcal{M} = \{ (\texttt{key}_k, y_k)\}$, where $\texttt{key}_k$ is the material name text and $y_k$ is the value of the physical property, e.g.  $\{ (\texttt{Aluminum}, 2700 \texttt{kg/m}^3), (\texttt{Oak Wood}, 650-900 \texttt{kg/m}^3)\}$. Note that $y_k$ may be a range of values due to the inherent uncertainty in the task.

Although it is theoretically possible for a VQA model such as BLIP-2~\cite{li2023blip}, LLaVA~\cite{liu2023llava}, or GPT-4V~\cite{yang2023dawn} to propose the materials directly from the image, we find that decomposing the task into two parts produces more reliable results in our experiments. In the future, as VQA models become more powerful, one model may be sufficient.

\paragraph{Zero-shot CLIP-based kernel regression.} 
Once the material dictionary $\mathcal{M}$ has been obtained, we use CLIP features to perform material retrieval for each source point.
We predict the physical property values of each point by taking its fused CLIP features and conducting a CLIP-based kernel regression using the material dictionary. Formally speaking, the physical property field is equal to:
\begin{equation}
    \rho(\mathbf{s}) = F(\mathbf{z}, \mathcal{M})
    = \frac{\sum_{k=1}^K \exp (w_k[\mathbf{s}] /T) y_k}{\sum_{k=1}^{K} \exp (w_k[\mathbf{s}]/T)},
\end{equation}
where $w_k[\mathbf{s}] = \psi_\text{CLIP}(\mathbf{z}, \texttt{key}_k)$ is the cosine similarity between semantic feature $\mathbf{z}$ and the language CLIP feature of the material name $\texttt{key}_k$, and $T$ is a temperature parameter chosen through validation. The physical property values can be propagated from the source points to any 3D query point via nearest-neighbor interpolation:
\begin{equation}
    \rho(\mathbf{x}) = \rho(\mathrm{argmin}_{\mathbf{s} \in \mathcal{S}} ||\mathbf{s} - \mathbf{x}||).
\end{equation}

\subsection{Object-level physical property aggregation} \label{sec:integration}
So far, we have discussed dense prediction of physical properties in a per-point manner. In practice, one may also be interested in object-level physical properties (e.g. mass) that require integration over volumes.
Although NeRF gives us an estimate of the geometry of the visible surfaces in the scene, most objects contain a large amount of empty space in their interior, which heavily affects volumetric integration but cannot be captured by NeRF due to occlusion. To circumvent this, we prompt the large language model again to estimate the thickness $t_k$ of each material in $\mathcal{M}$, and then use the estimated thickness to define a set of cuboids sampled on the object's surface. Similar to the source point sampling, the cuboid locations are sampled by voxel-downsampling surface points, and we define their size to be $d \times d \times \tau$, where $d$ is the voxel size and $\tau$ is the predicted thickness at that point. Formally, the prediction $\hat{m}$ for the integral over the cuboids is given by 
\begin{equation}
\hat{m} = \sum_{\mathbf{x} \in \mathcal{V}} \frac{\sum_{k=1}^K \exp (w_k[\mathbf{x}]/T) y_k t_k }{\sum_{k=1}^{K} \exp (w_k[\mathbf{x}]/T)} d^2,
\end{equation}
where $\mathcal{V}$ is the set of voxel-downsampled points. Since this volume estimation can be biased depending on the geometry of the object, we introduce a scalar multiplication factor $c$ determined by validation. We also clamp the total volume to an upper bound estimated by depth carving to avoid wildly inaccurate thickness predictions.

\section{Experiments}
We evaluate NeRF2Physics across two datasets. First, we compare NeRF2Physics with existing methods for per-object mass estimation on a set of 500 objects from the Amazon Berkeley Objects (ABO) dataset~\cite{collins2022abo}, which we refer to as ABO-500. To evaluate the dense prediction capabilities of NeRF2Physics, we collect our own dataset of real-world objects with per-point friction and hardness measurements.

\begin{table*}[!t]
    \centering
    \resizebox{\linewidth}{!}{
\setlength{\tabcolsep}{0.2em} %
\renewcommand{\arraystretch}{1.}
    \begin{tabular}{ccccc}
    
    Input RGB &
    CLIP Features &
    \multicolumn{2}{c}{Material Segmentation} &
    Mass Density (kg/m$^3$)
    \\
         \tikz{
        \node[draw=white, line width=0mm, inner sep=0pt] 
        {\includegraphics[width=.21\linewidth, trim={4cm 5cm 4cm 3cm}, clip]{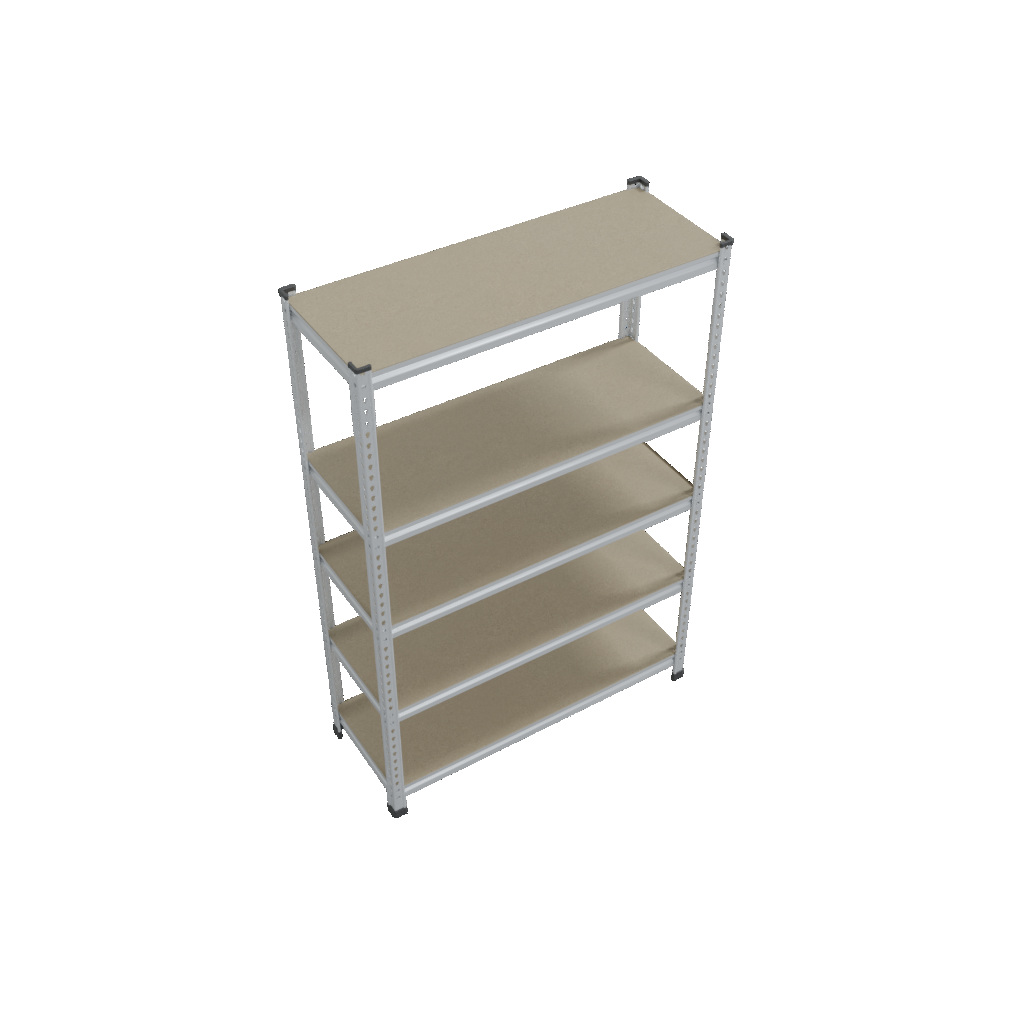}}
        }&
        \tikz{
        \node[draw=white, line width=0mm, inner sep=0pt] 
        {\includegraphics[width=.21\linewidth, trim={4cm 5cm 4cm 3cm}, clip]{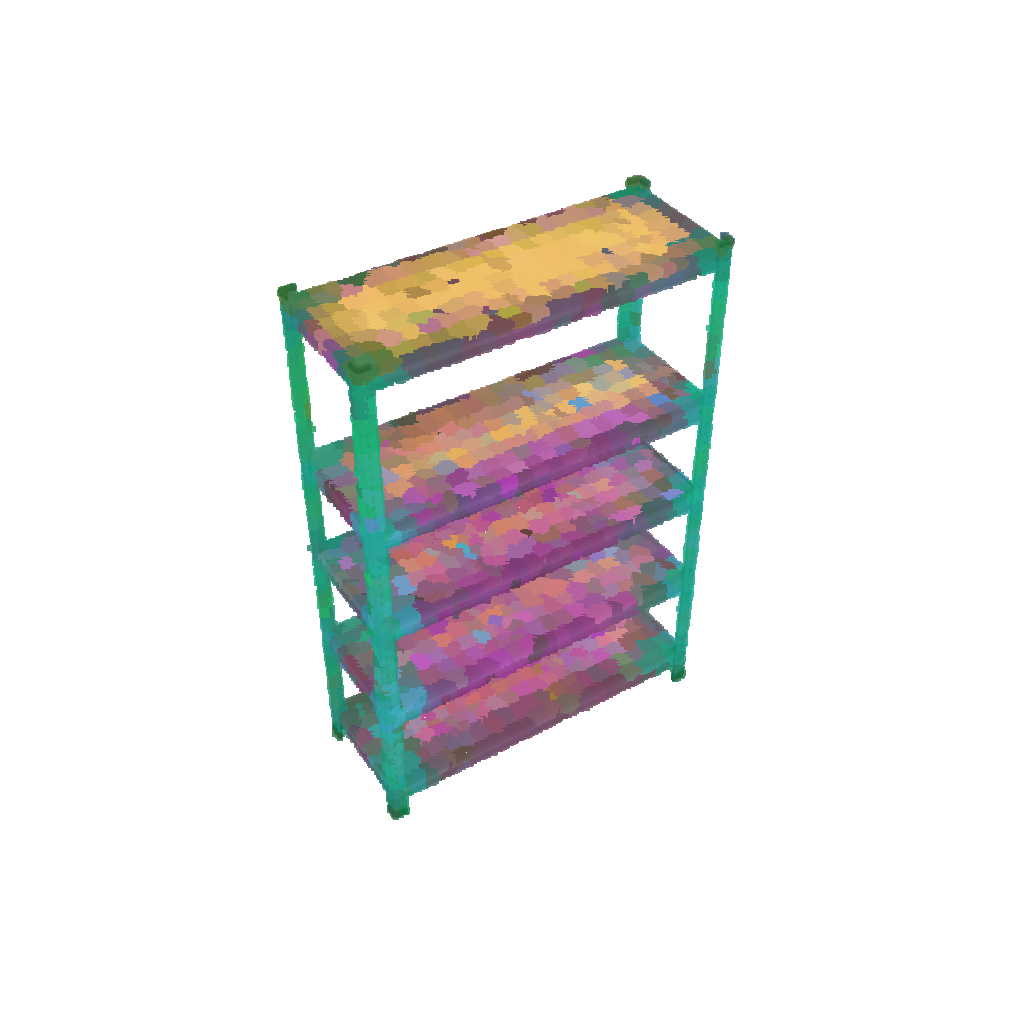} }
        }&
         \tikz{
        \node[draw=white, line width=0mm, inner sep=0pt] 
        {\includegraphics[width=.21\linewidth, trim={4cm 5cm 4cm 3cm}, clip]{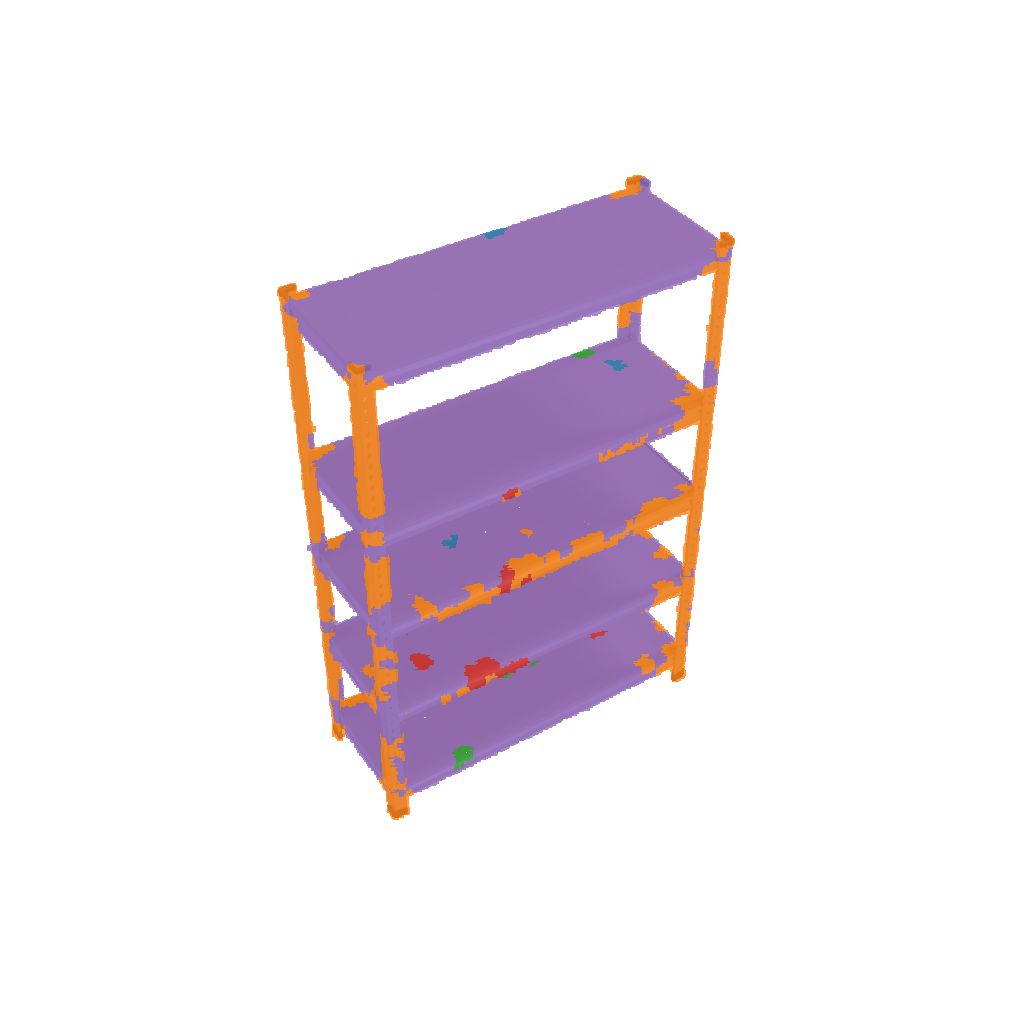}}
        }&
         \tikz{
        \node[draw=white, line width=0mm, inner sep=0pt] 
        {\includegraphics[width=.16\linewidth, trim={0cm 0cm 0cm 0cm}, clip]{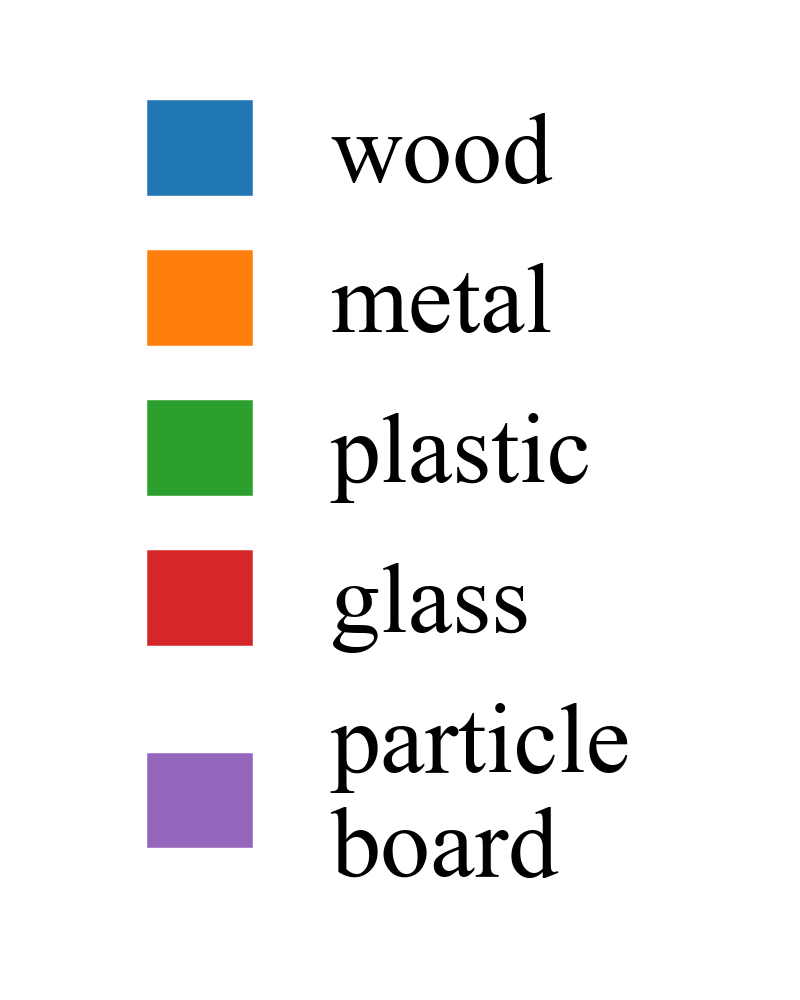}}
        }&
        \tikz{
        \node[draw=white, line width=0mm, inner sep=0pt] 
        { \includegraphics[width=.21\linewidth, trim={4cm 5cm 4cm 3cm}, clip]{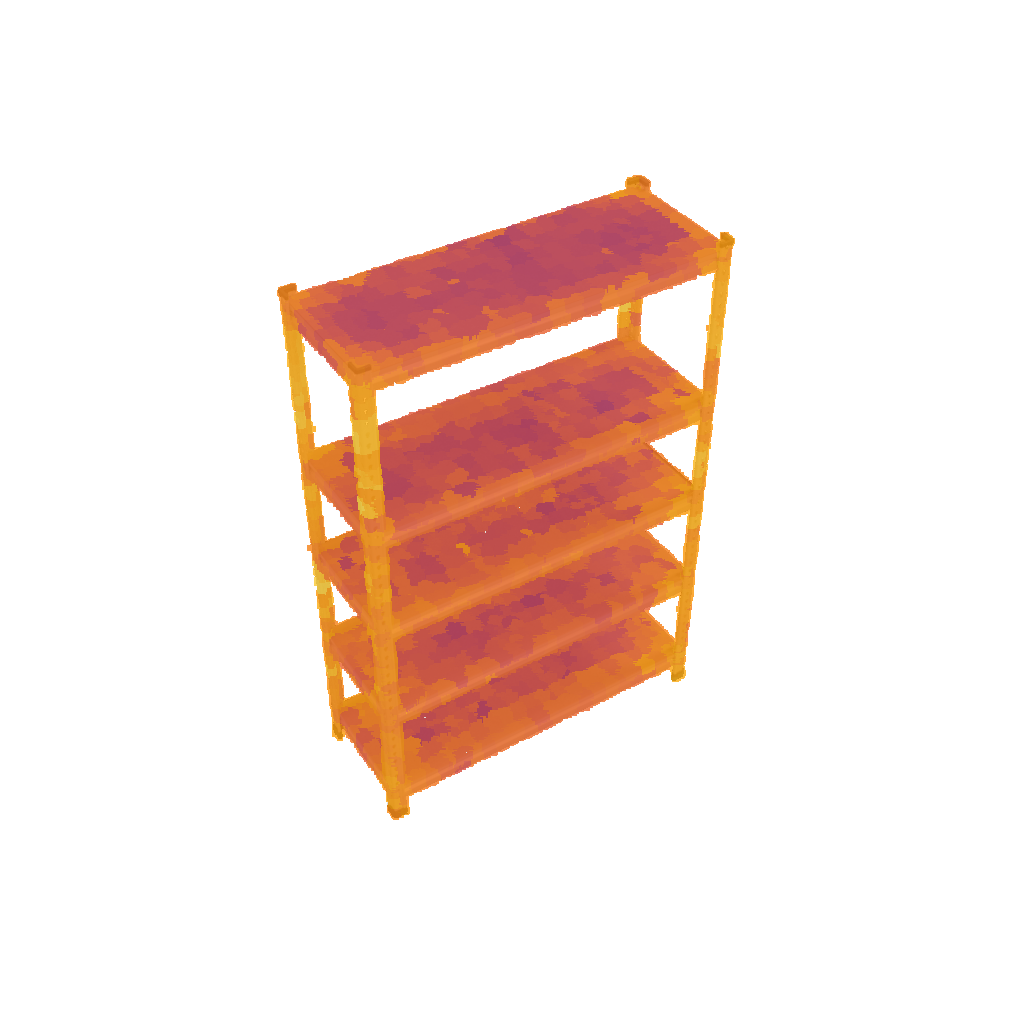}}
        }
        \vspace{-5pt}
        \\
        
        \tikz{
        \node[draw=white, line width=0mm, inner sep=0pt] 
        {\includegraphics[width=.21\linewidth, trim={4cm 3.3cm 4cm 4.7cm}, clip]{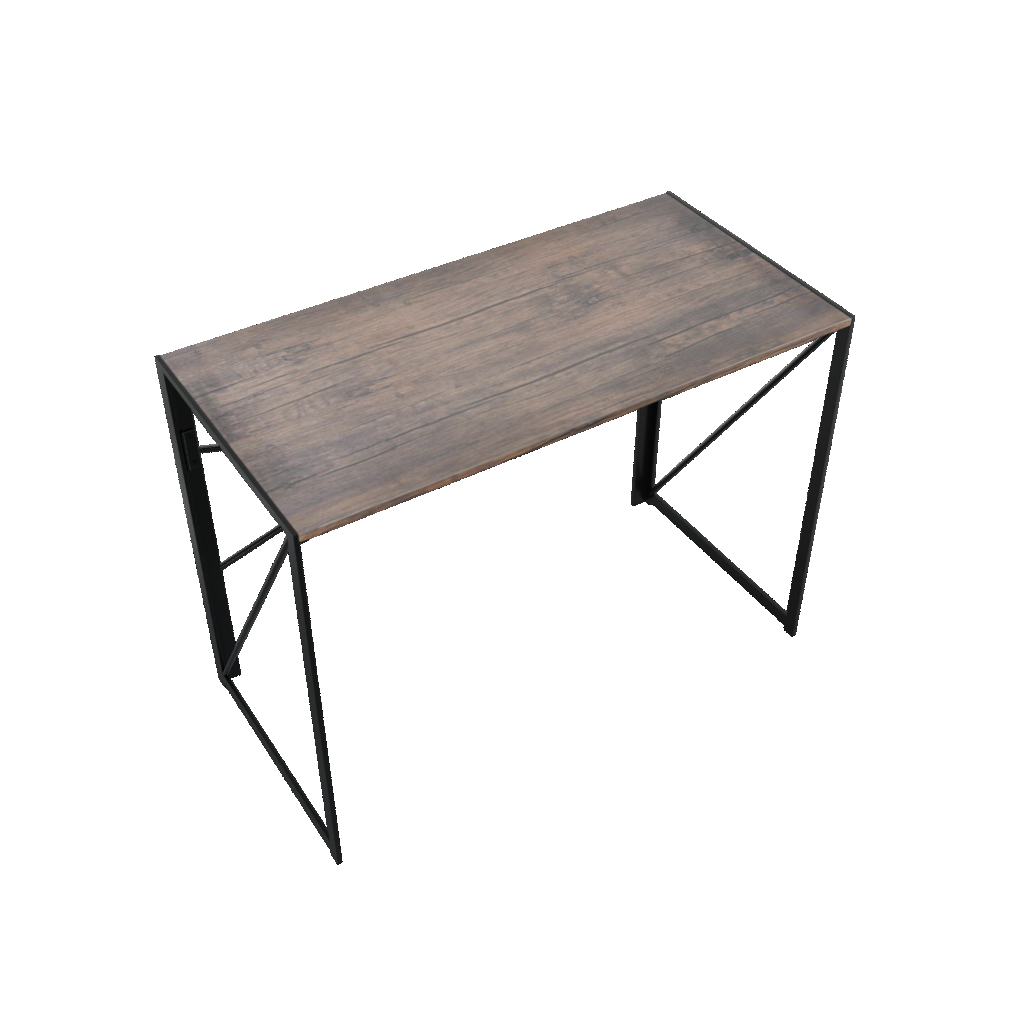}}
        }&
        \tikz{
        \node[draw=white, line width=0mm, inner sep=0pt] 
        {\includegraphics[width=.21\linewidth, trim={4cm 3.3cm 4cm 4.7cm}, clip]{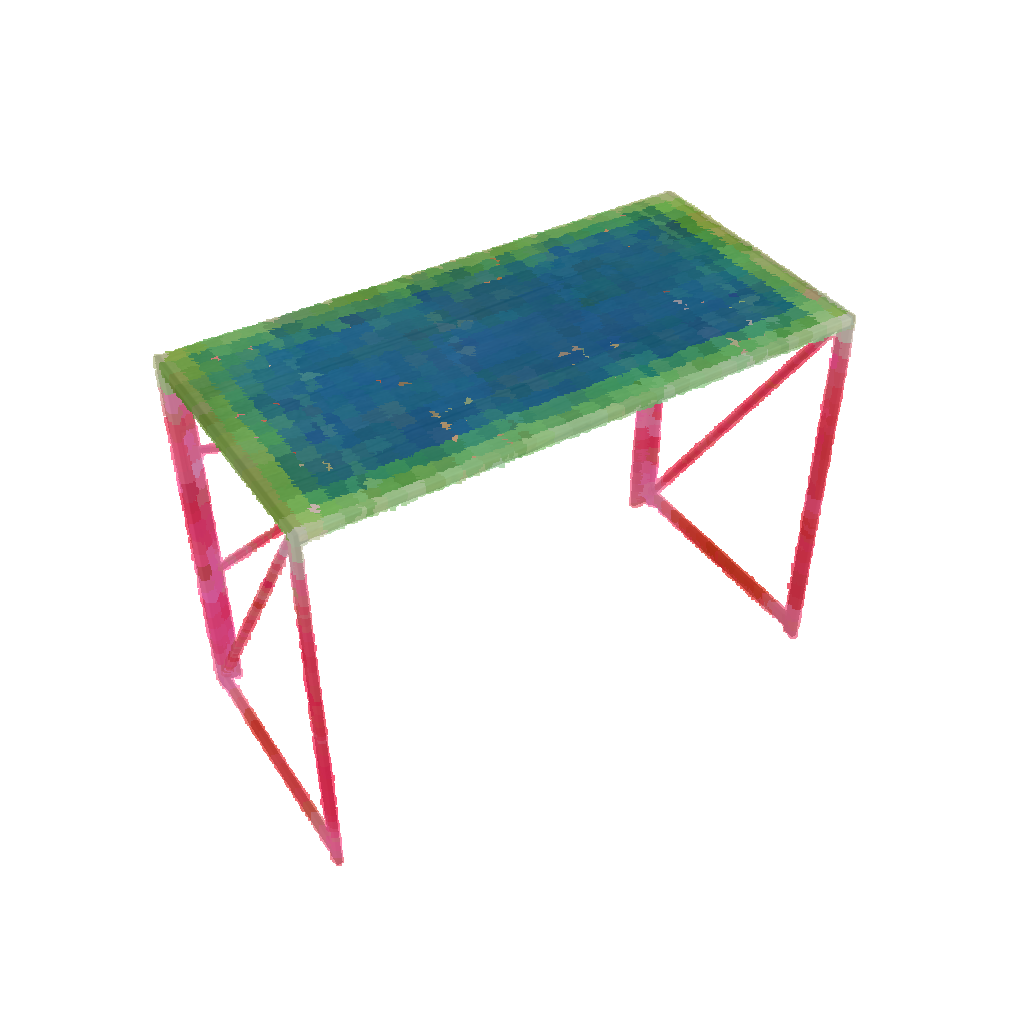} }
        }&
         \tikz{
        \node[draw=white, line width=0mm, inner sep=0pt] 
        {\includegraphics[width=.21\linewidth, trim={4cm 3.3cm 4cm 4.7cm}, clip]{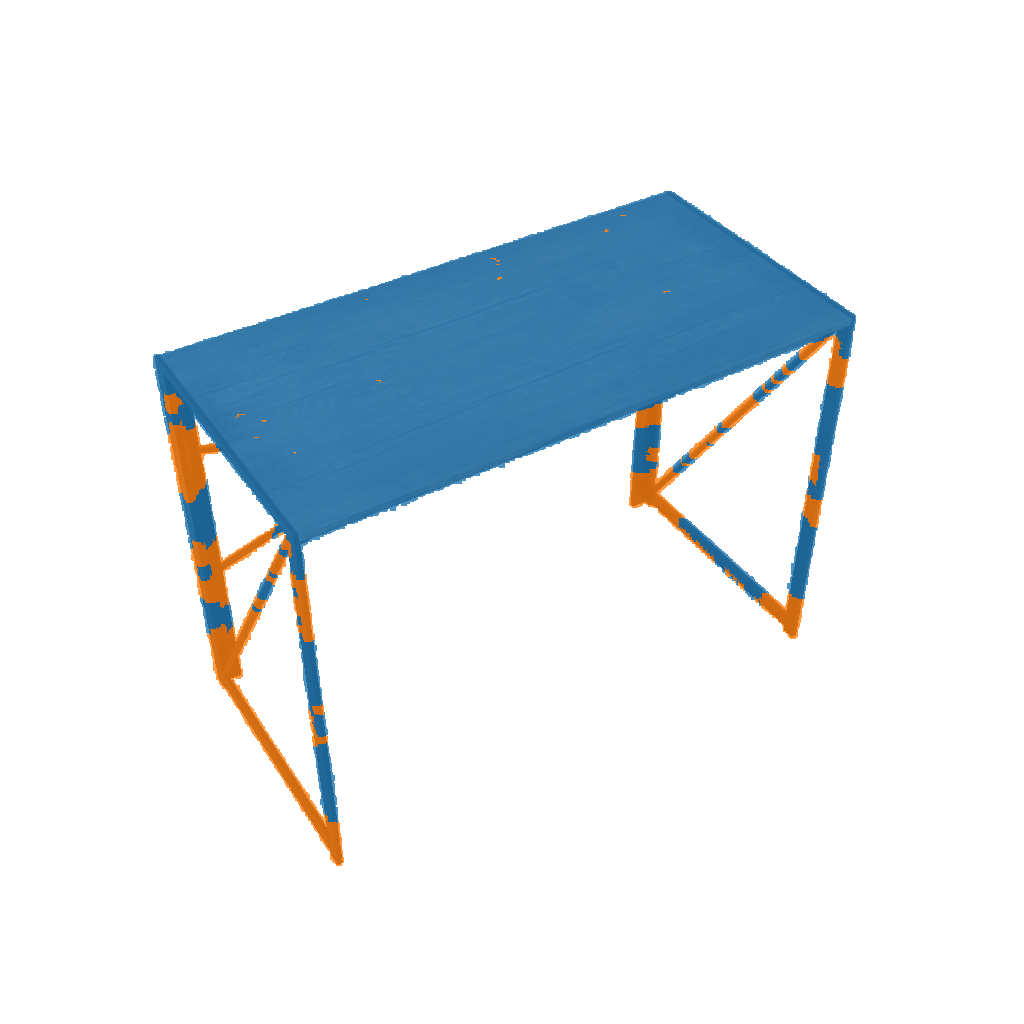}}
        }&
        \tikz{
        \node[draw=white, line width=0mm, inner sep=0pt] 
        {\includegraphics[width=.16\linewidth, trim={0cm 0cm 0cm 0cm}, clip]{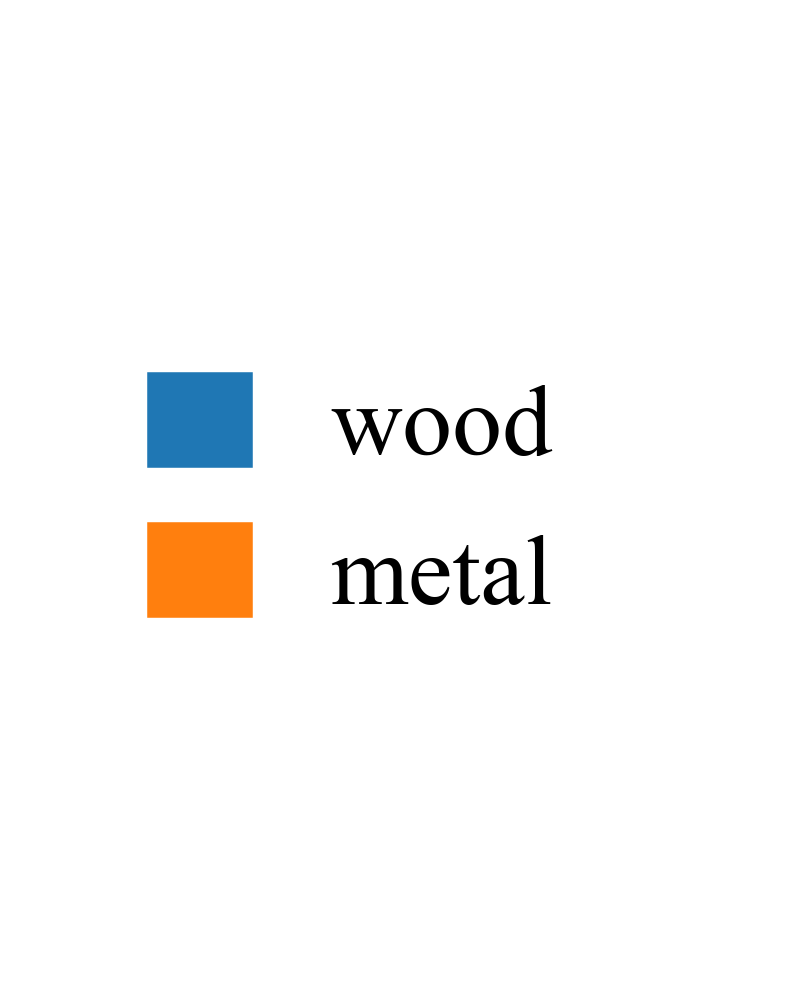}}
        }&
        \tikz{
        \node[draw=white, line width=0mm, inner sep=0pt] 
        { \includegraphics[width=.21\linewidth, trim={4cm 3.3cm 4cm 4.7cm}, clip]{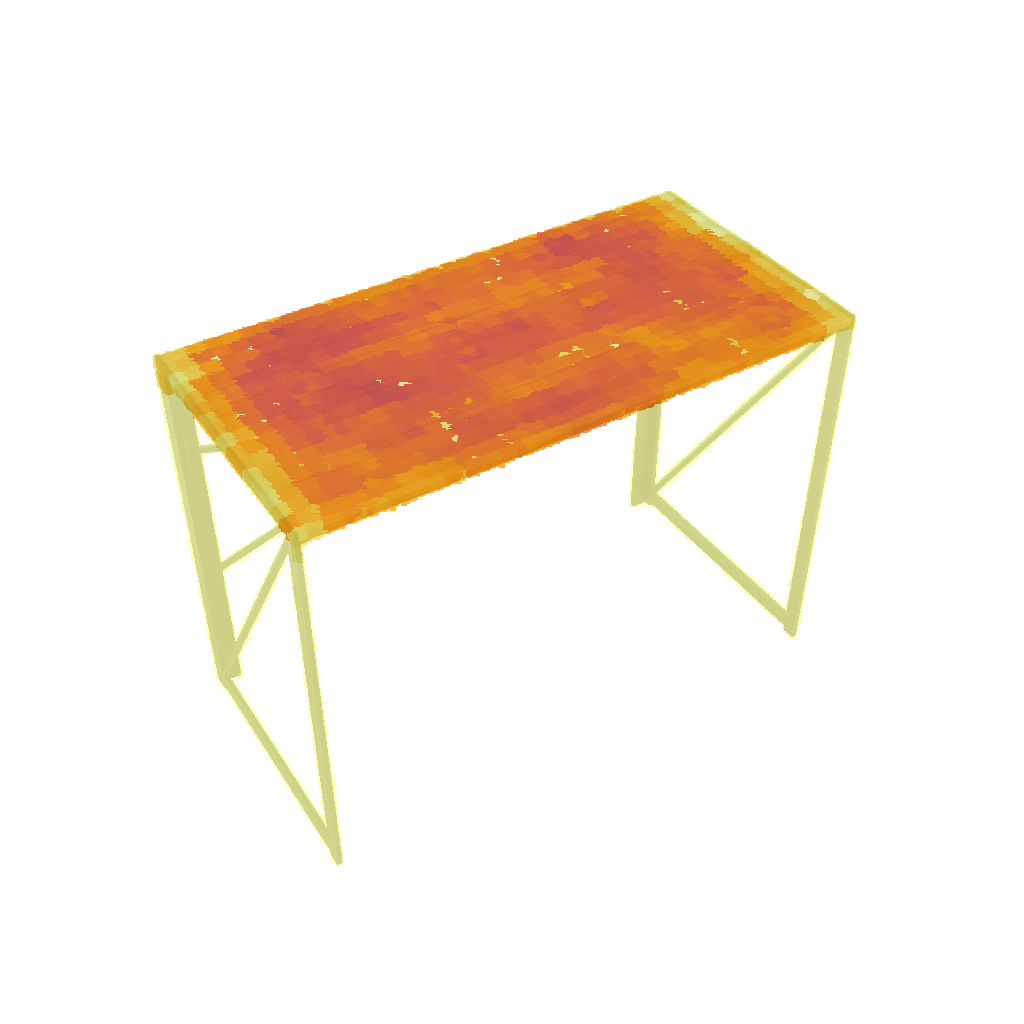}}
        }
        \vspace{-12pt}
        \\

        \tikz{
        \node[draw=white, line width=0mm, inner sep=0pt] 
        {\includegraphics[width=.21\linewidth, trim={7cm 7cm 7cm 7cm}, clip]{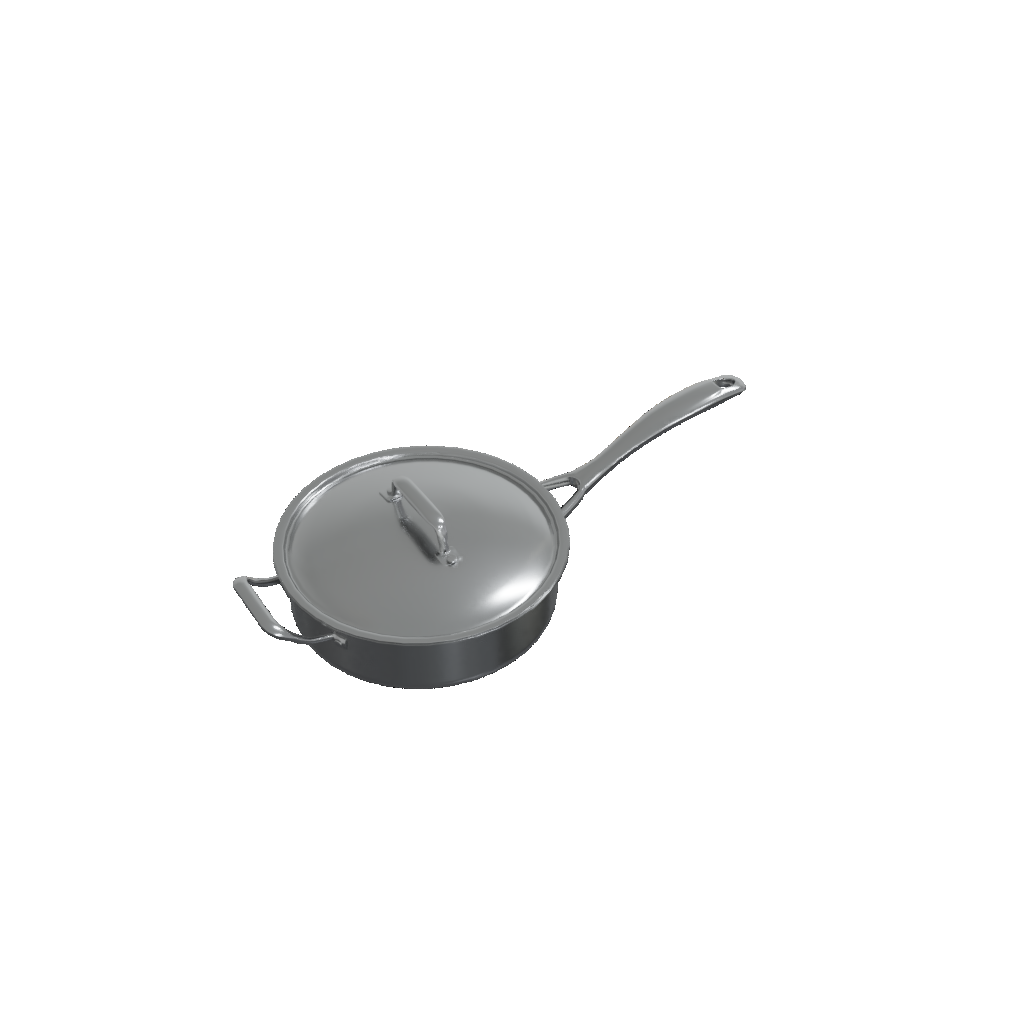}}
        }&
        \tikz{
        \node[draw=white, line width=0mm, inner sep=0pt] 
        {\includegraphics[width=.21\linewidth, trim={7cm 7cm 7cm 7cm}, clip]{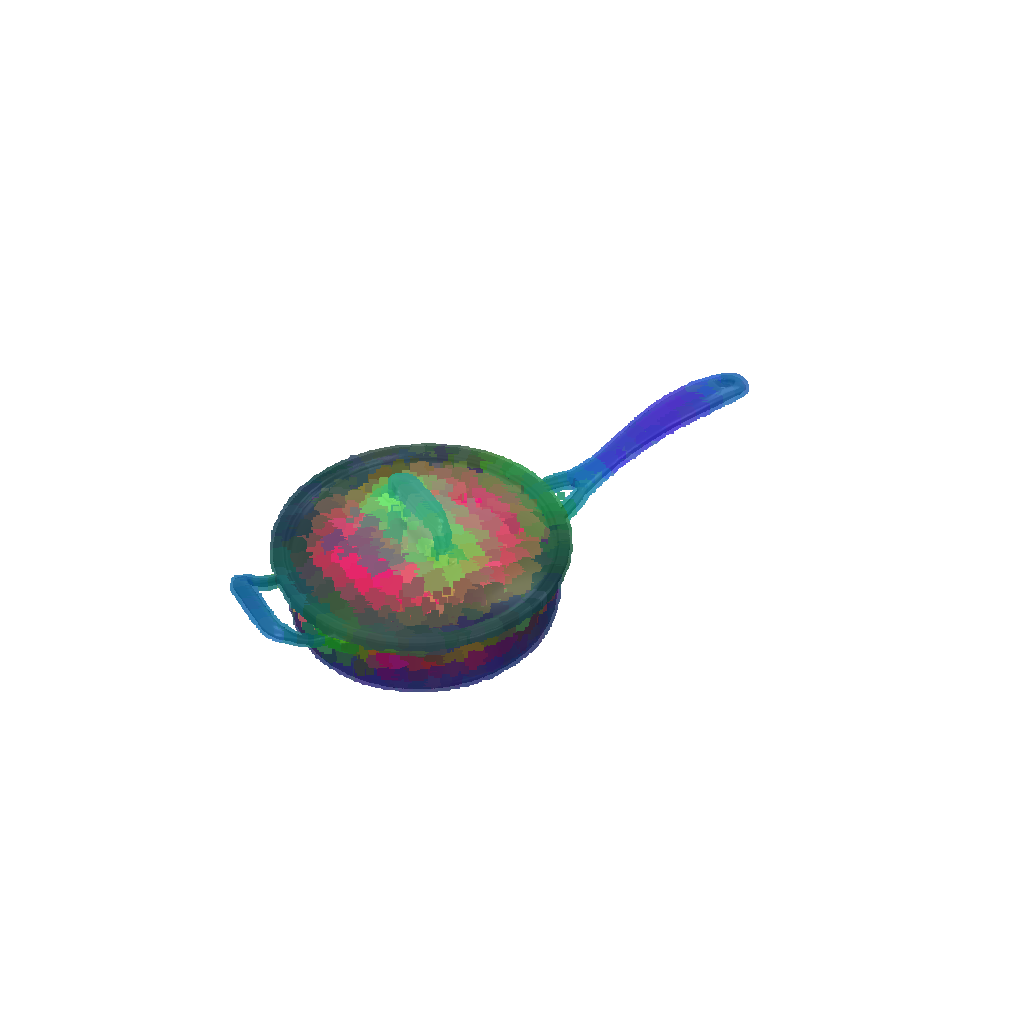} }
        }&
         \tikz{
        \node[draw=white, line width=0mm, inner sep=0pt] 
        {\includegraphics[width=.21\linewidth, trim={7cm 7cm 7cm 7cm}, clip]{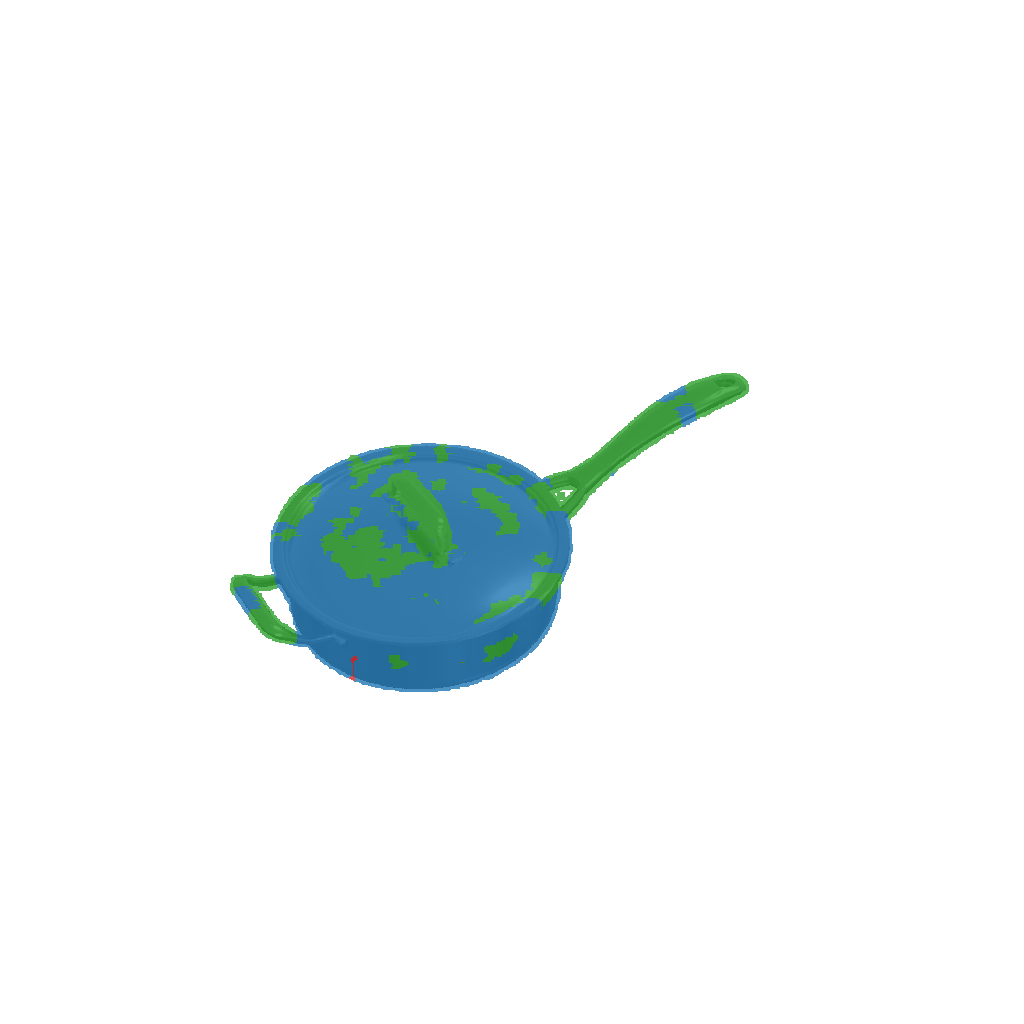}}
        }&
        \tikz{
        \node[draw=white, line width=0mm, inner sep=0pt] 
        {\includegraphics[width=.16\linewidth, trim={0cm 0cm 0cm 0cm}, clip]{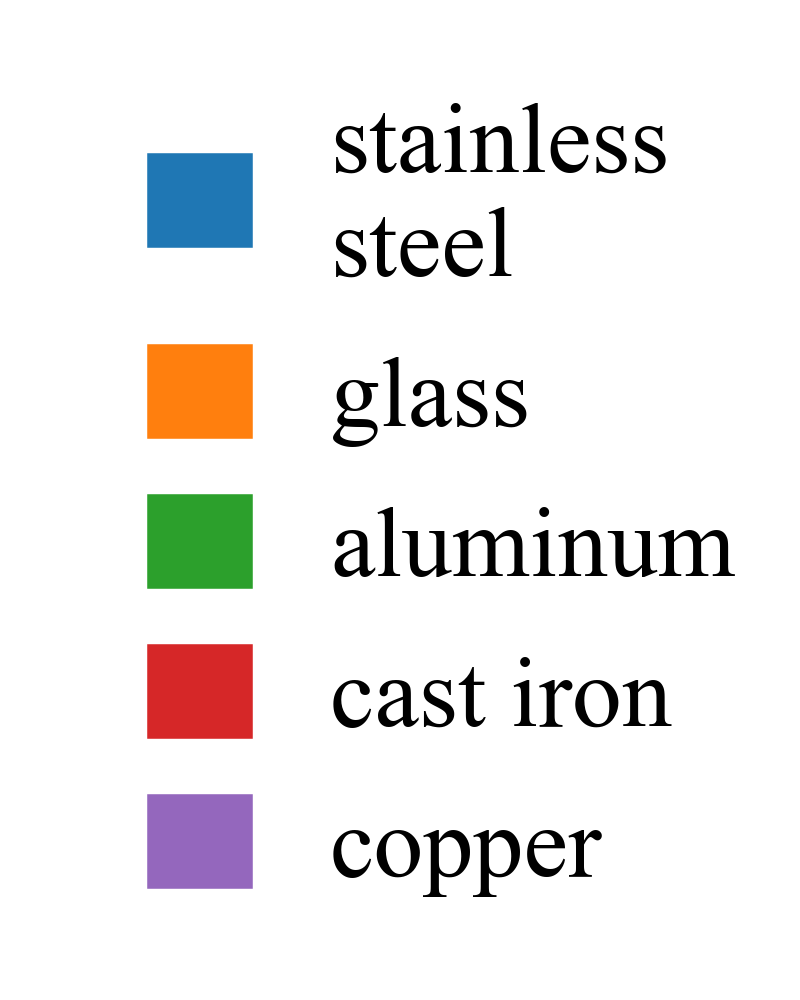}}
        }&
        \tikz{
        \node[draw=white, line width=0mm, inner sep=0pt] 
        { \includegraphics[width=.21\linewidth, trim={7cm 7cm 7cm 7cm}, clip]{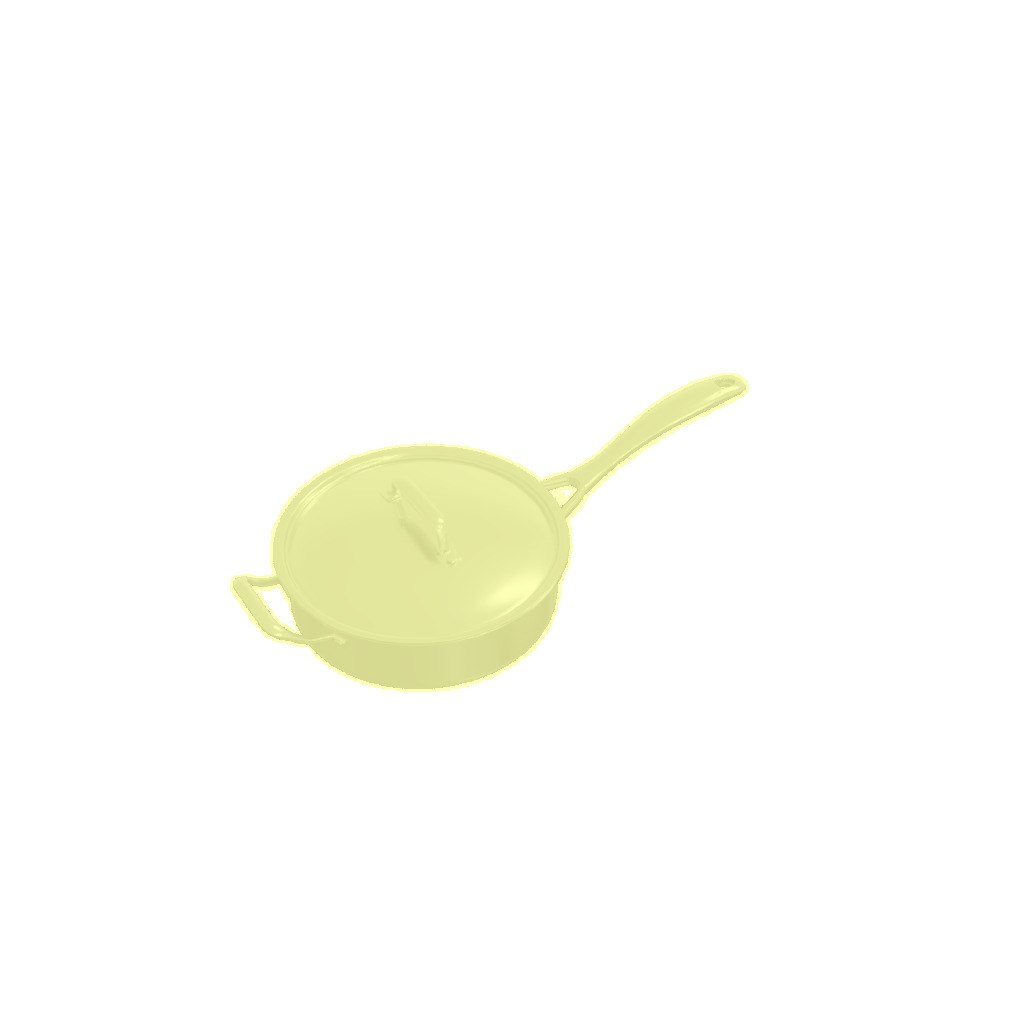}}
        }
        \vspace{-4pt}
        \\

        \tikz{
        \node[draw=white, line width=0mm, inner sep=0pt] 
        {\includegraphics[width=.21\linewidth, trim={4cm 4cm 4cm 4cm}, clip]{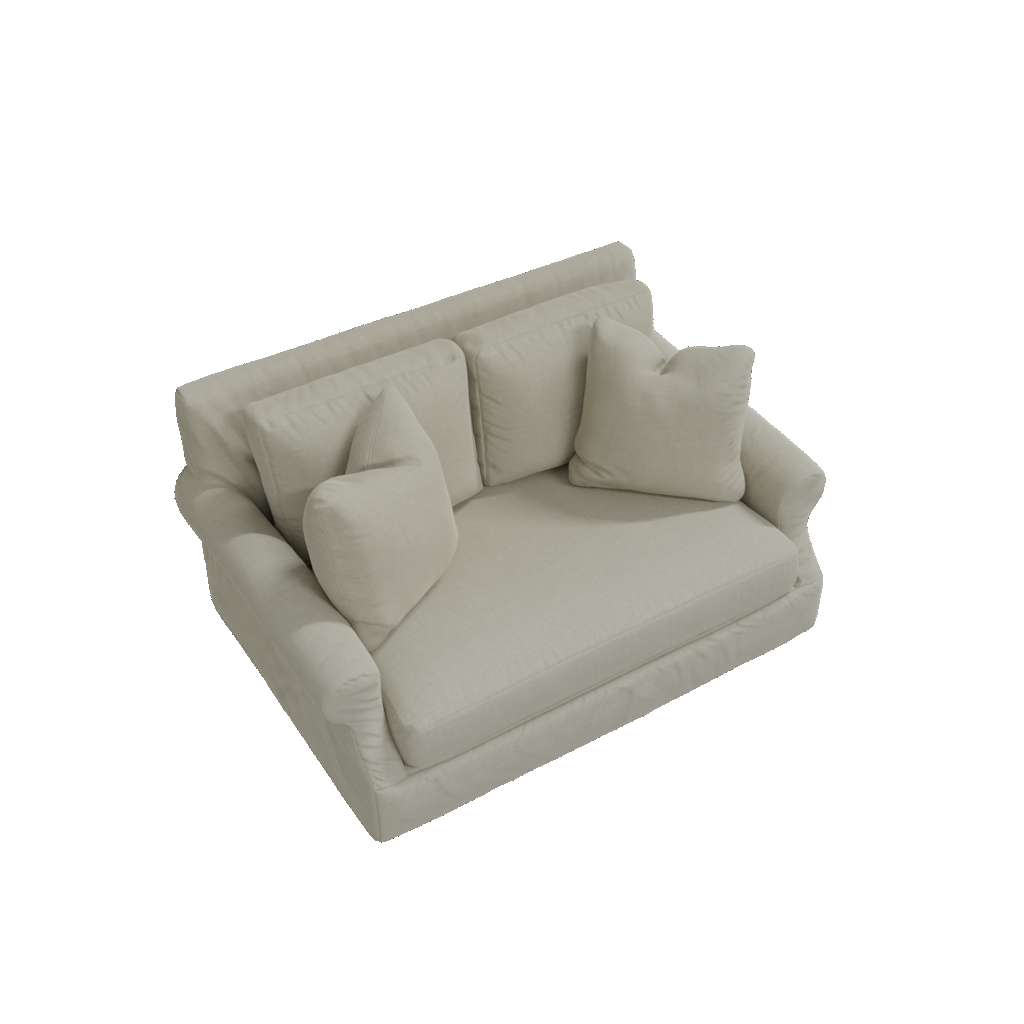}}
        }&
        \tikz{
        \node[draw=white, line width=0mm, inner sep=0pt] 
        {\includegraphics[width=.21\linewidth, trim={4cm 4cm 4cm 4cm}, clip]{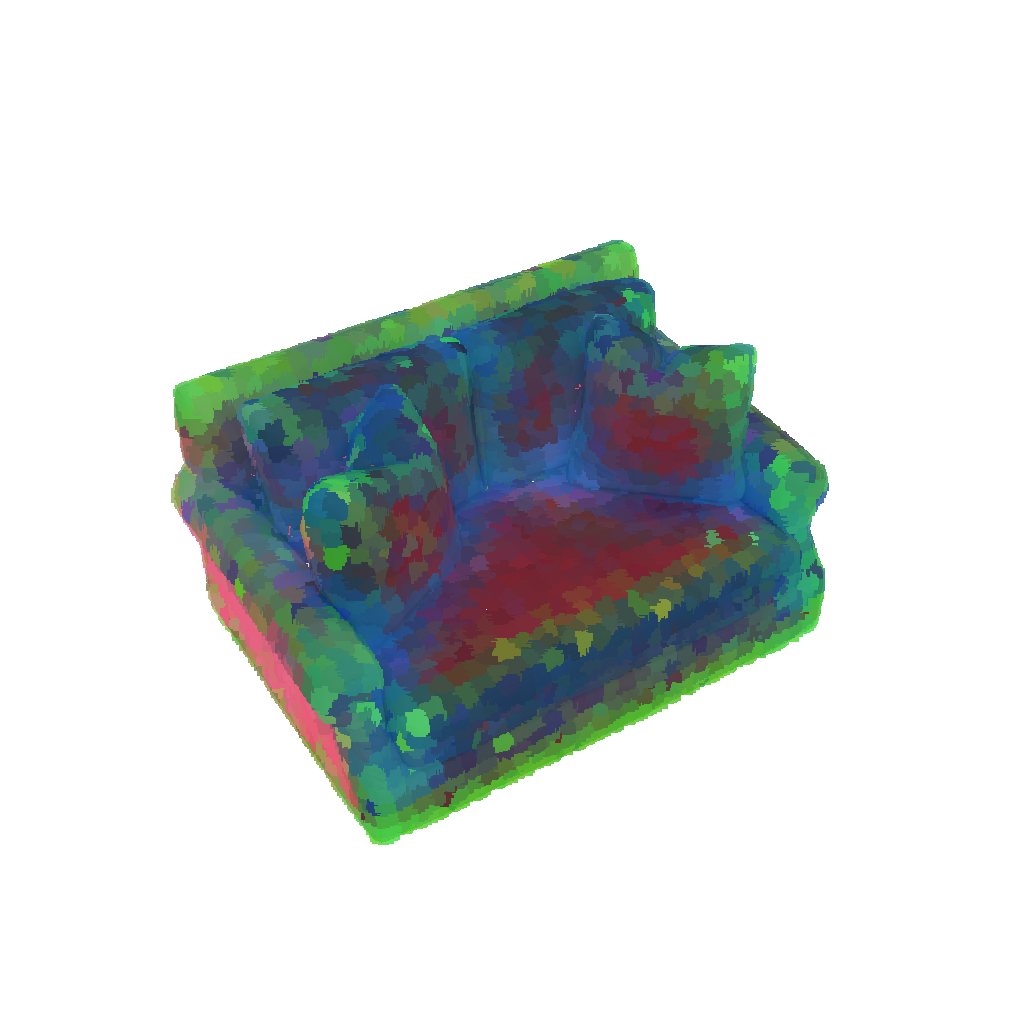} }
        }&
         \tikz{
        \node[draw=white, line width=0mm, inner sep=0pt] 
        {\includegraphics[width=.21\linewidth, trim={4cm 4cm 4cm 4cm}, clip]{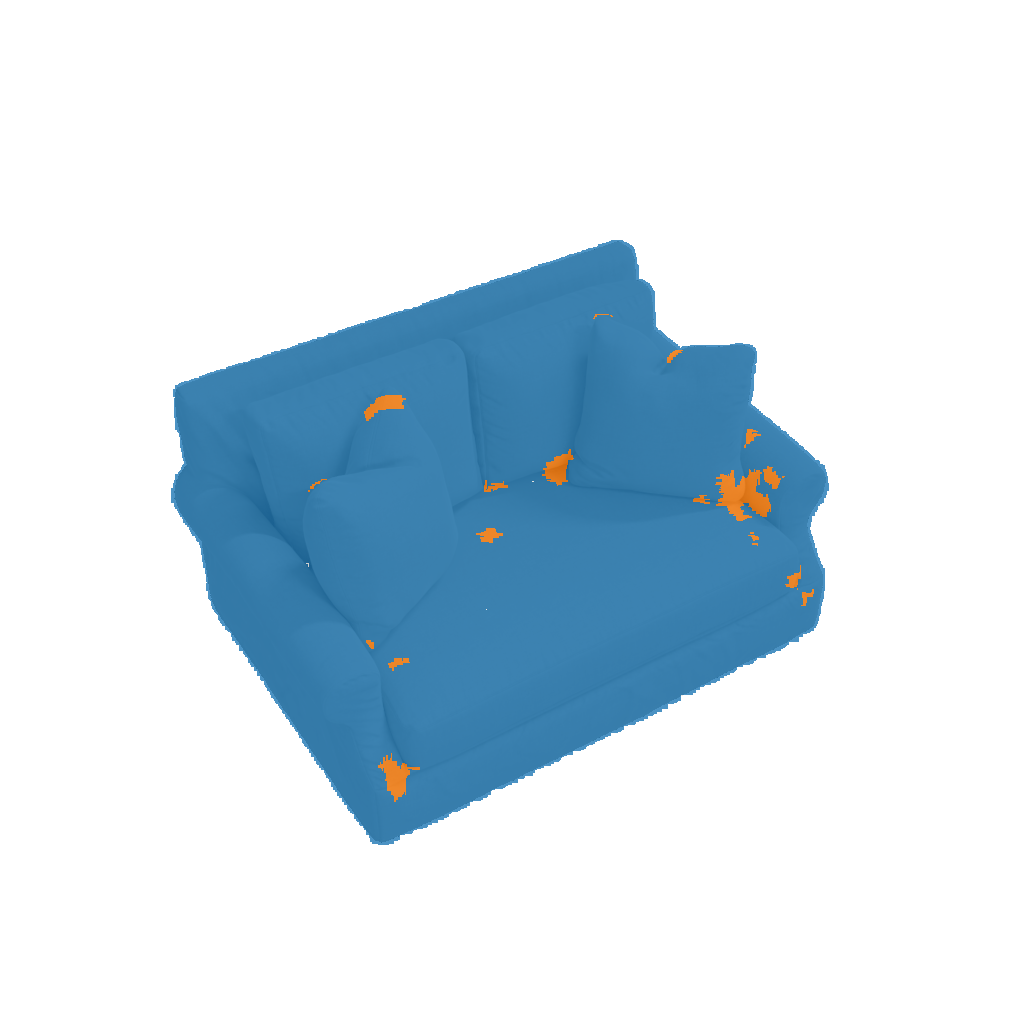}}
        }&
        \tikz{
        \node[draw=white, line width=0mm, inner sep=0pt] 
        {\includegraphics[width=.16\linewidth, trim={0cm 0cm 0cm 0cm}, clip]{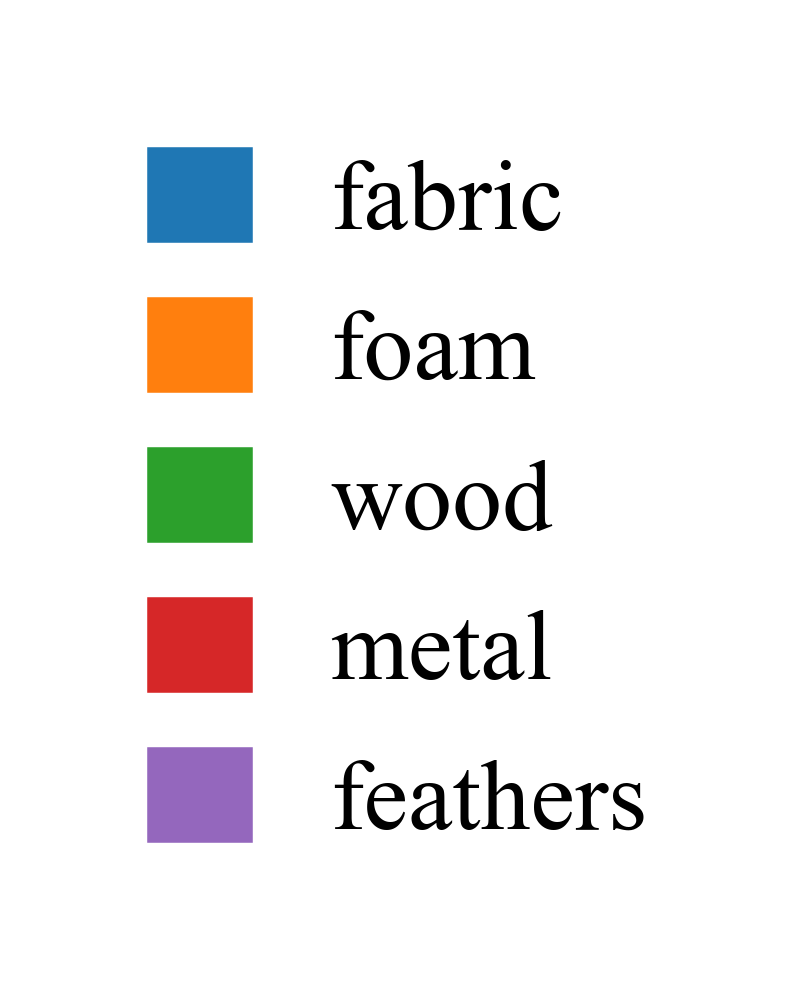}}
        }&
        \tikz{
        \node[draw=white, line width=0mm, inner sep=0pt] 
        { \includegraphics[width=.21\linewidth, trim={4cm 4cm 4cm 4cm}, clip]{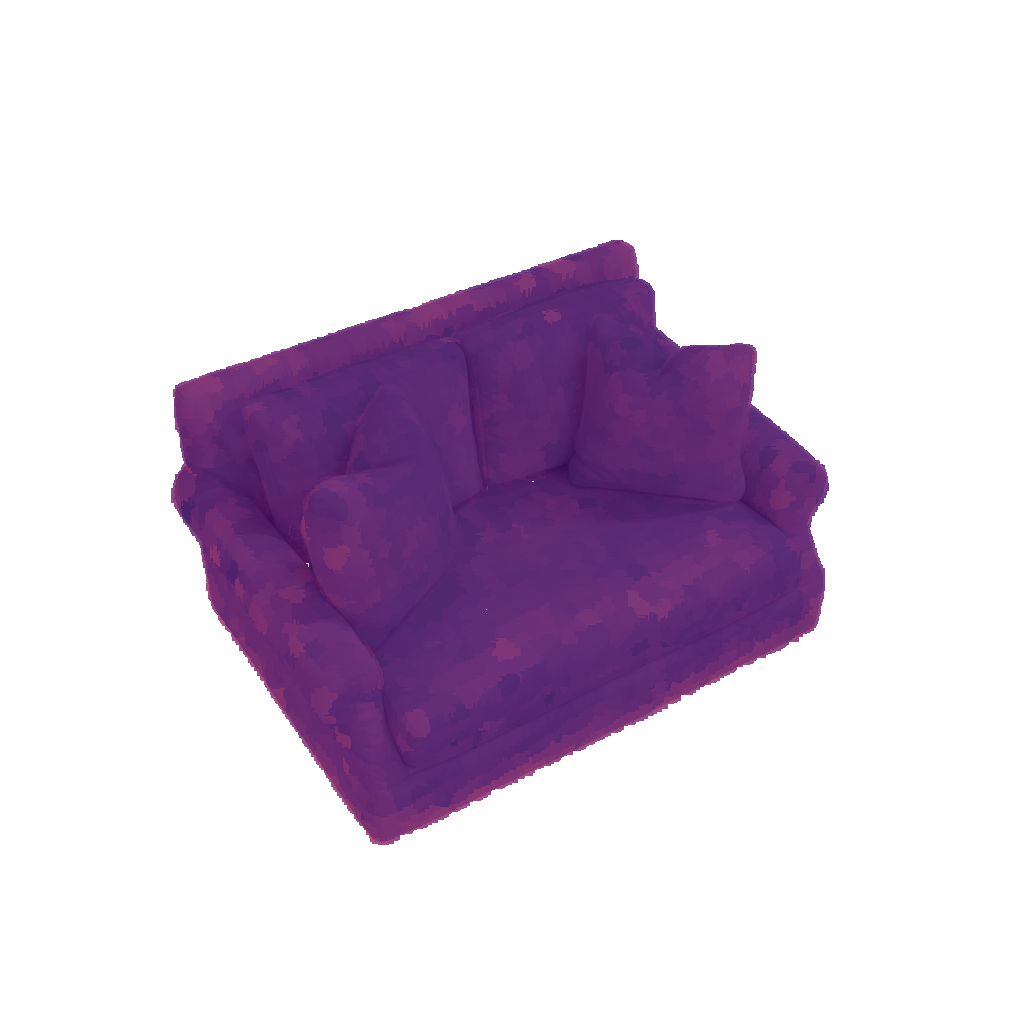}}
        }
        \vspace{-4pt}
        \\

        \tikz{
        \node[draw=white, line width=0mm, inner sep=0pt] 
        {\includegraphics[width=.21\linewidth, trim={8cm 8cm 6cm 6cm}, clip]{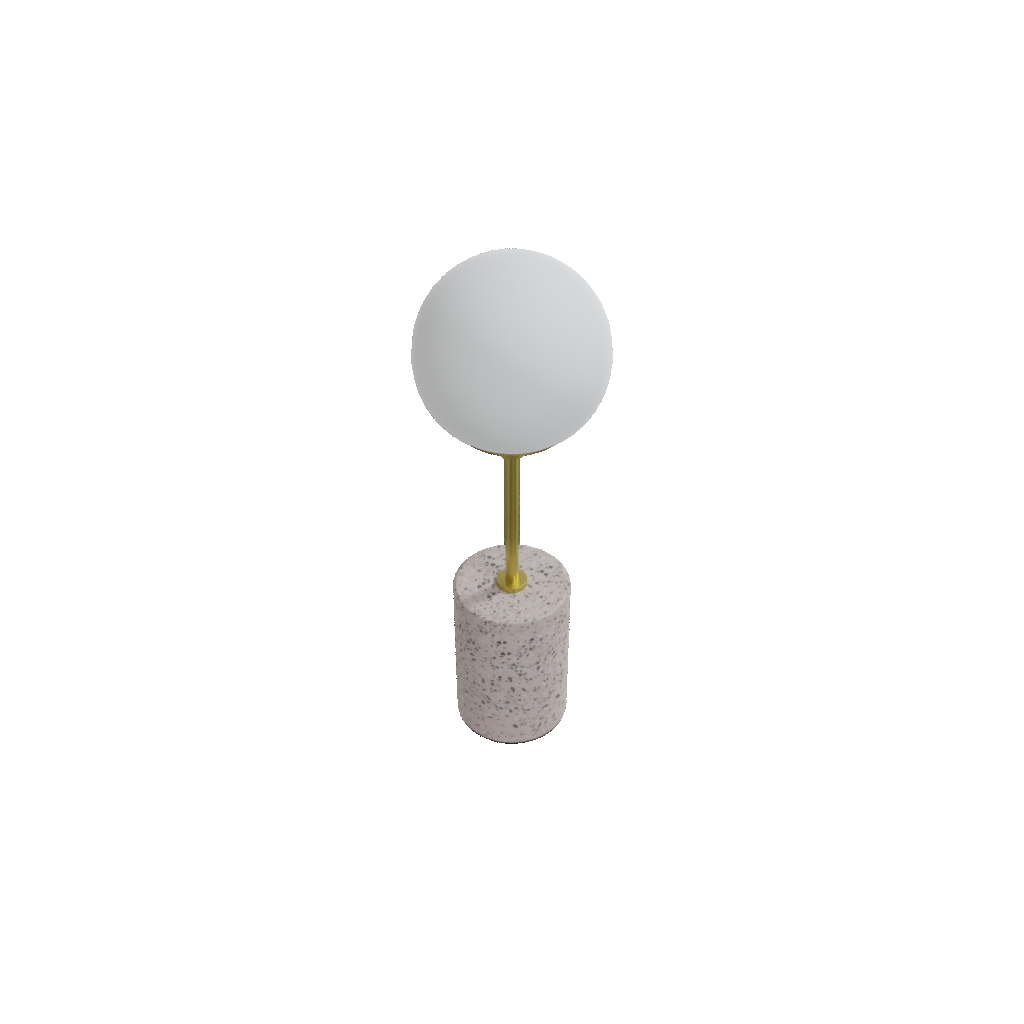}}
        }&
        \tikz{
        \node[draw=white, line width=0mm, inner sep=0pt] 
        {\includegraphics[width=.21\linewidth, trim={8cm 8cm 6cm 6cm}, clip]{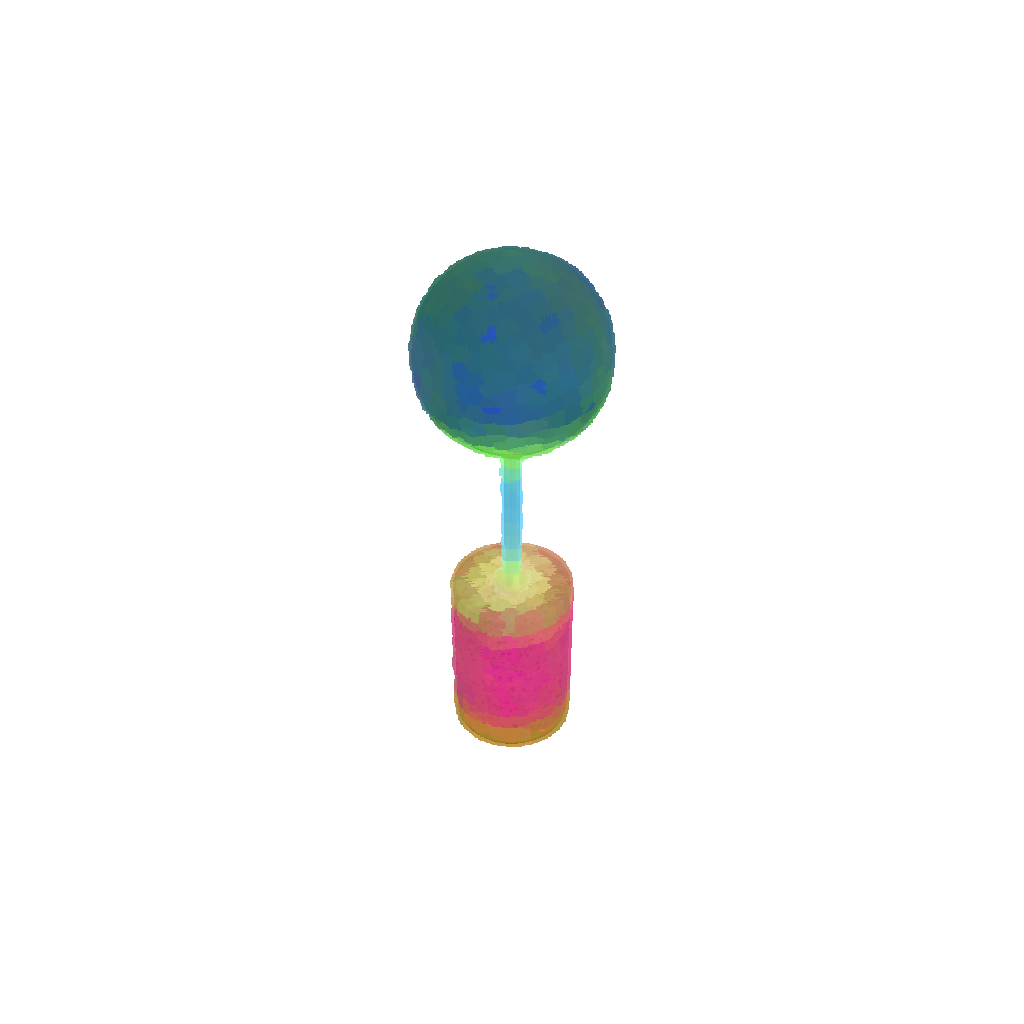} }
        }&
         \tikz{
        \node[draw=white, line width=0mm, inner sep=0pt] 
        {\includegraphics[width=.21\linewidth, trim={8cm 8cm 6cm 6cm}, clip]{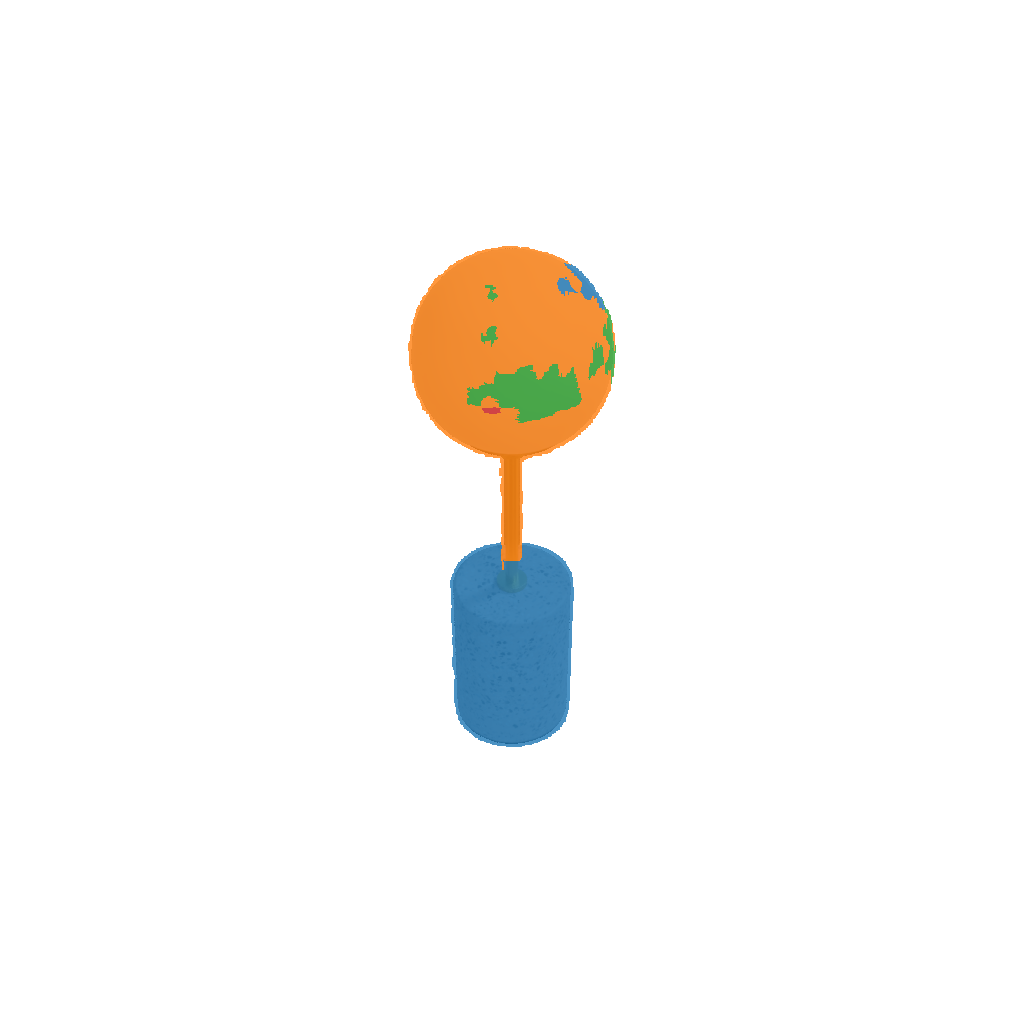}}
        }&
        \tikz{
        \node[draw=white, line width=0mm, inner sep=0pt] 
        {\includegraphics[width=.16\linewidth, trim={0cm 0cm 0cm 0cm}, clip]{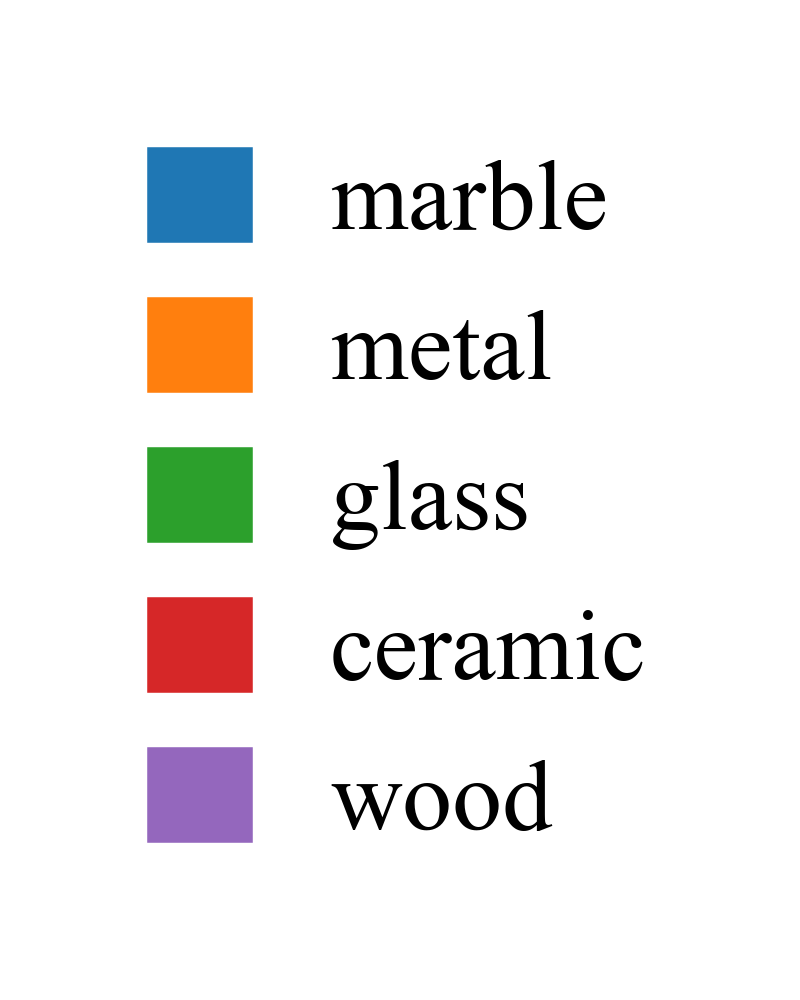}}
        }&
        \tikz{
        \node[draw=white, line width=0mm, inner sep=0pt] 
        { \includegraphics[width=.21\linewidth, trim={8cm 8cm 6cm 6cm}, clip]{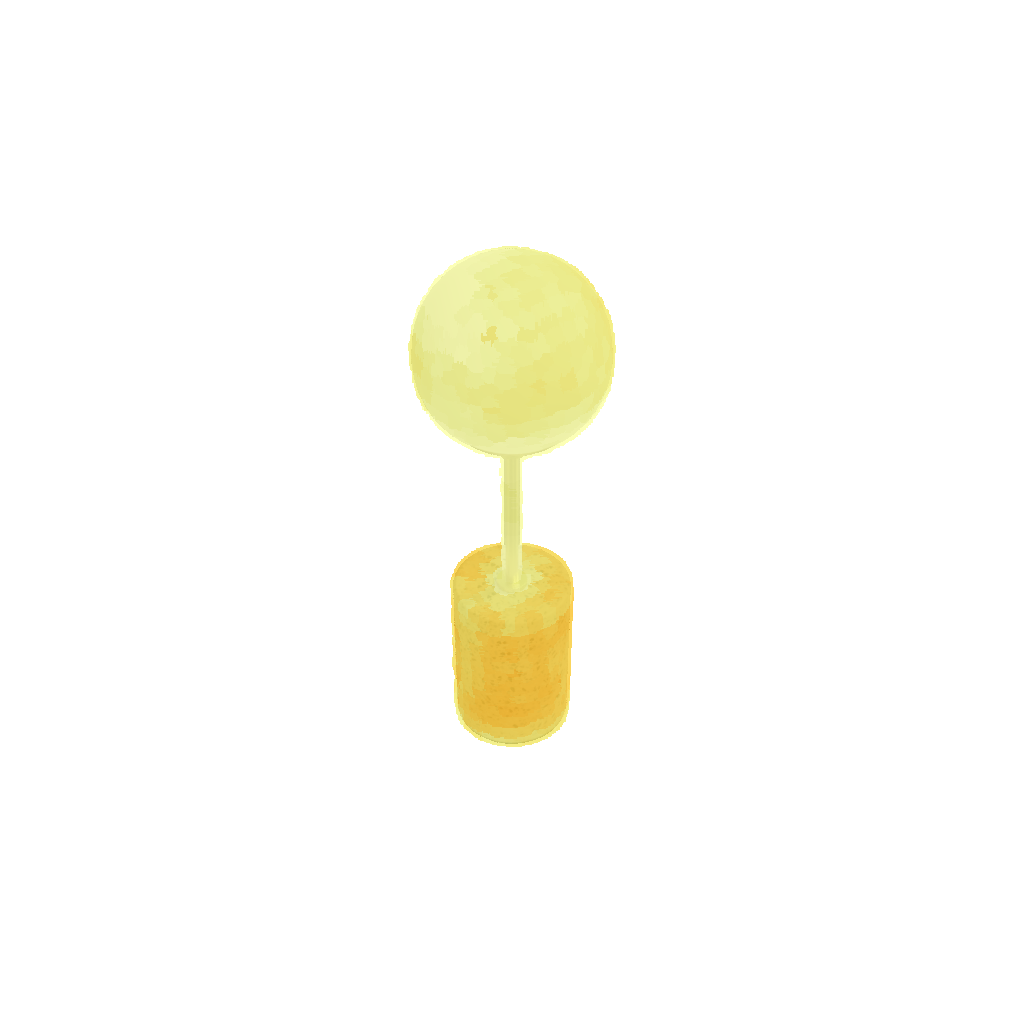}}
        }
        \vspace{-4pt}
        \\

        \tikz{
        \node[draw=white, line width=0mm, inner sep=0pt] 
        {\includegraphics[width=.21\linewidth, trim={4cm 5cm 4cm 3cm}, clip]{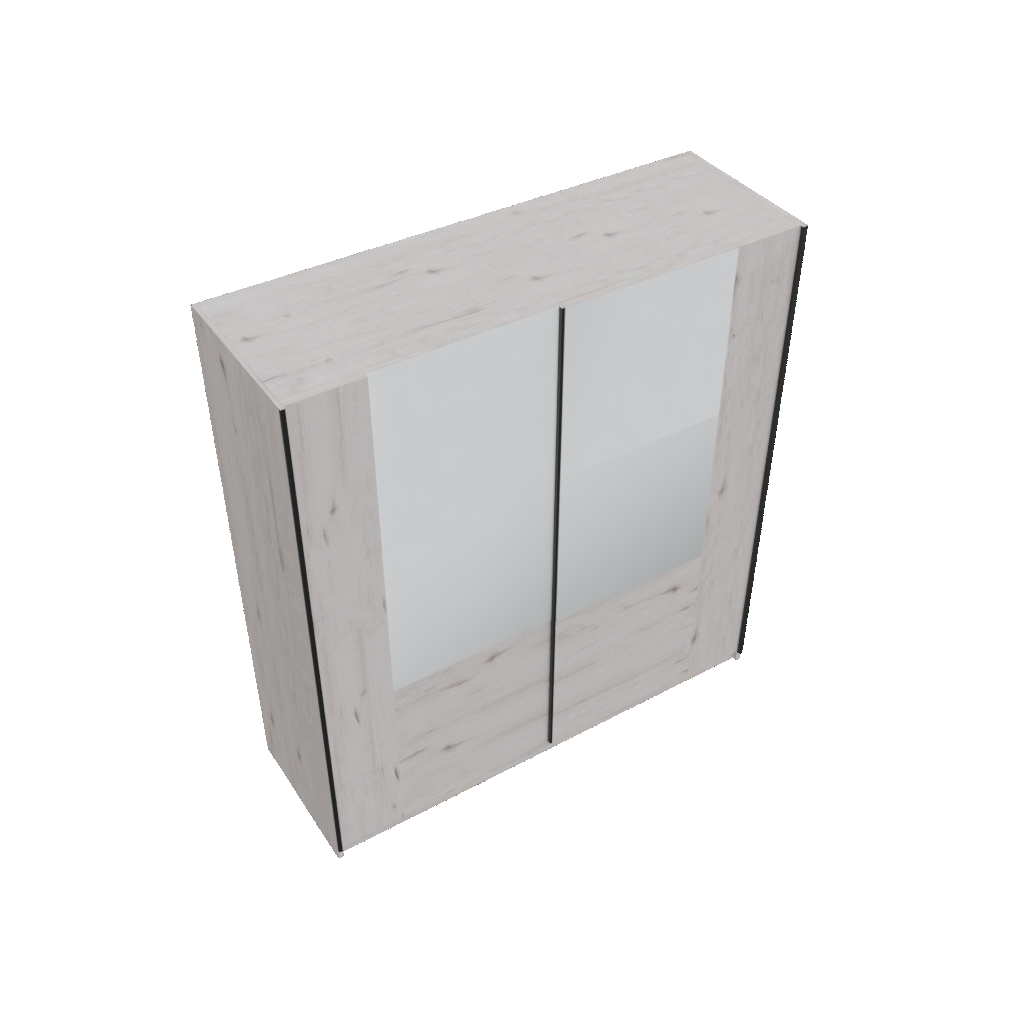}}
        }&
        \tikz{
        \node[draw=white, line width=0mm, inner sep=0pt] 
        {\includegraphics[width=.21\linewidth, trim={4cm 5cm 4cm 3cm}, clip]{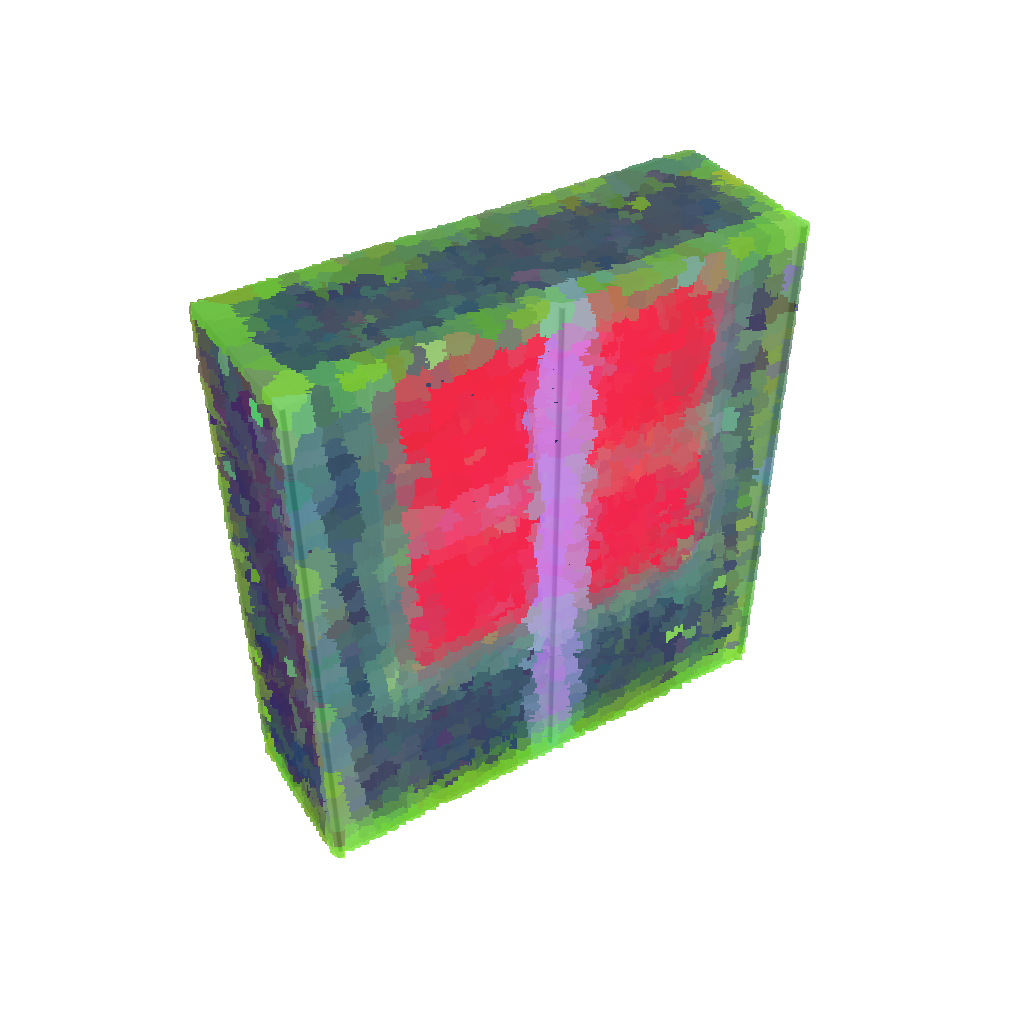} }
        }&
         \tikz{
        \node[draw=white, line width=0mm, inner sep=0pt] 
        {\includegraphics[width=.21\linewidth, trim={4cm 5cm 4cm 3cm}, clip]{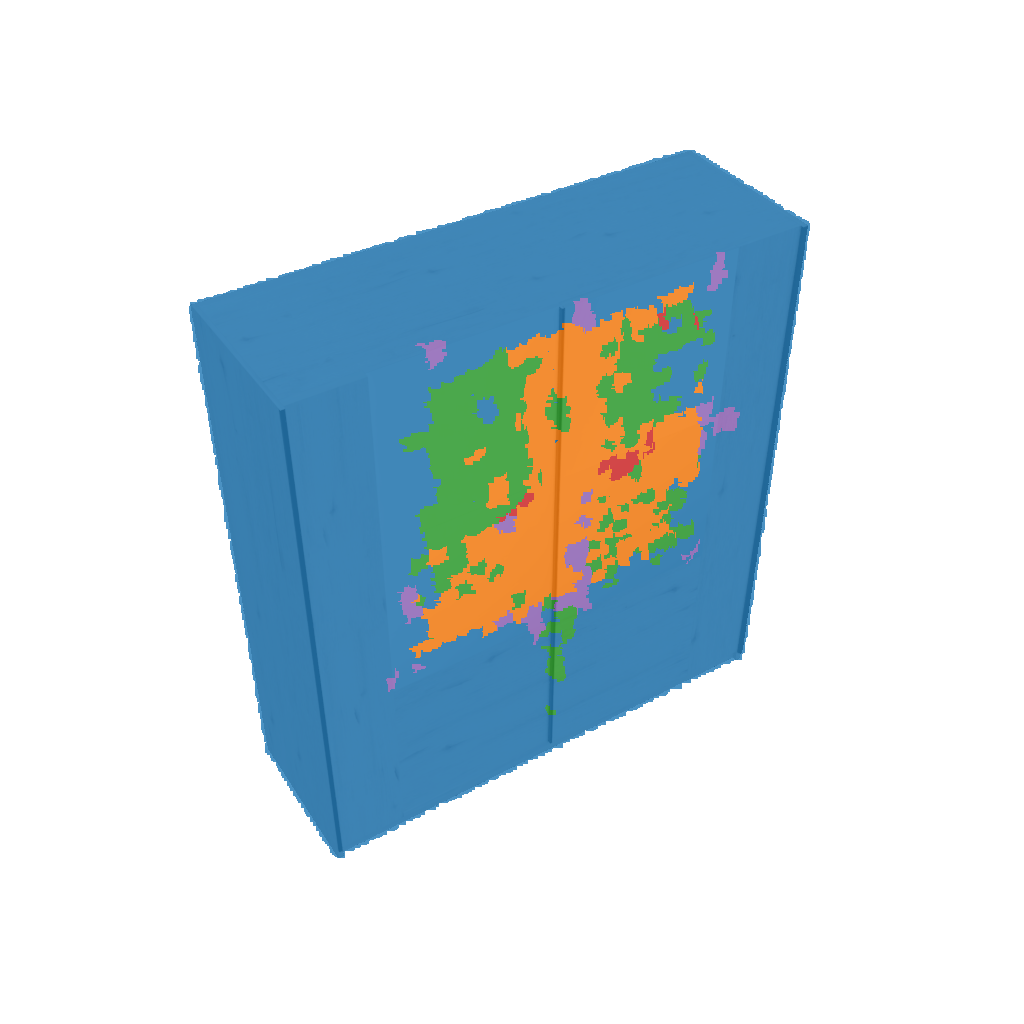}}
        }&
        \tikz{
        \node[draw=white, line width=0mm, inner sep=0pt] 
        {\includegraphics[width=.16\linewidth, trim={0cm 0cm 0cm 0cm}, clip]{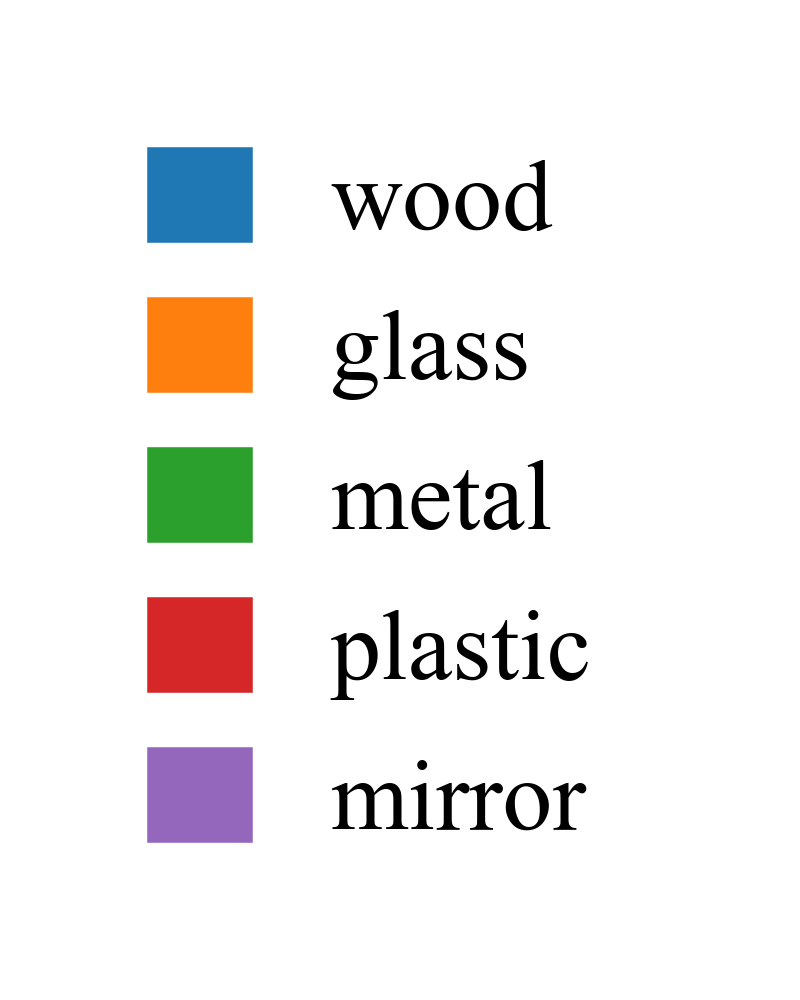}}
        }&
        \tikz{
        \node[draw=white, line width=0mm, inner sep=0pt] 
        { \includegraphics[width=.21\linewidth, trim={4cm 5cm 4cm 3cm}, clip]{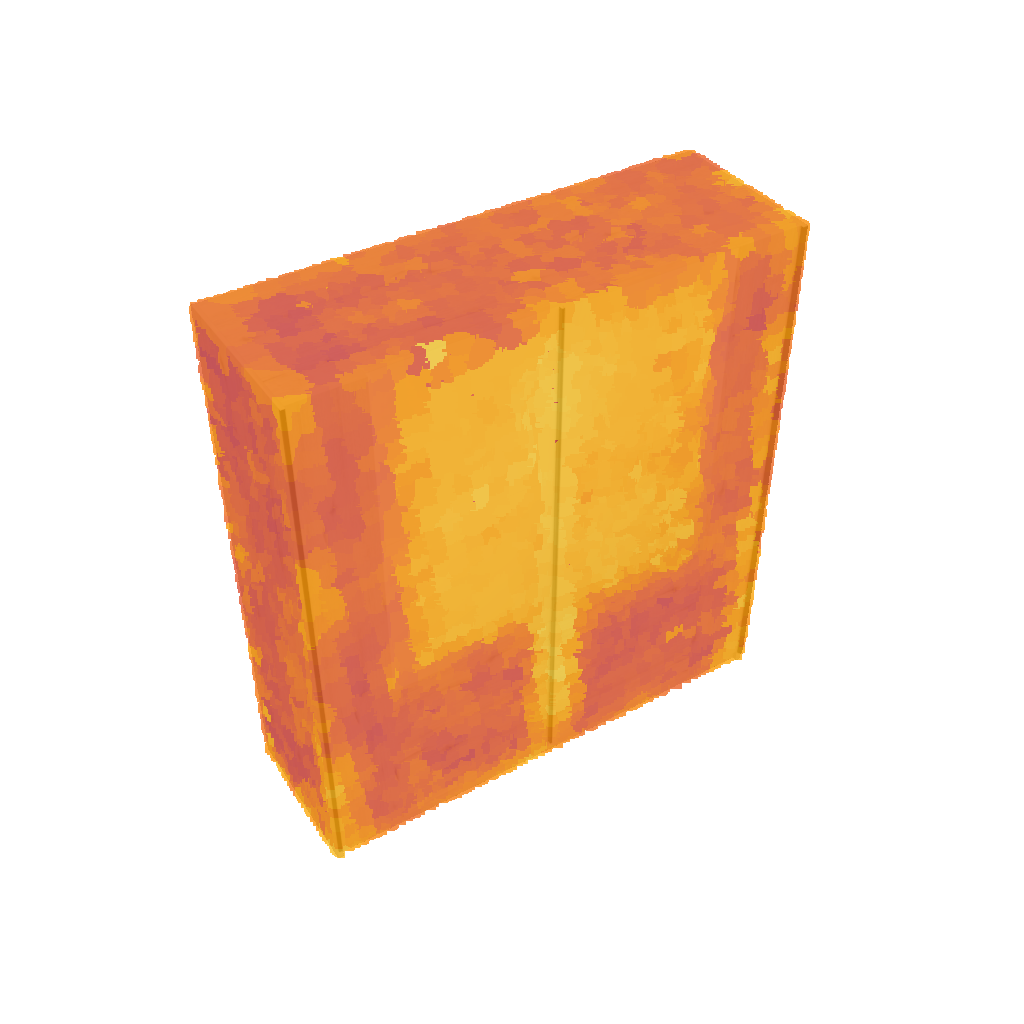}}
        }
        \vspace{-4pt}
        \\

        &
        &
        &
        &
        \includegraphics[width=.21\linewidth, trim={0 0 0 9cm}, clip]{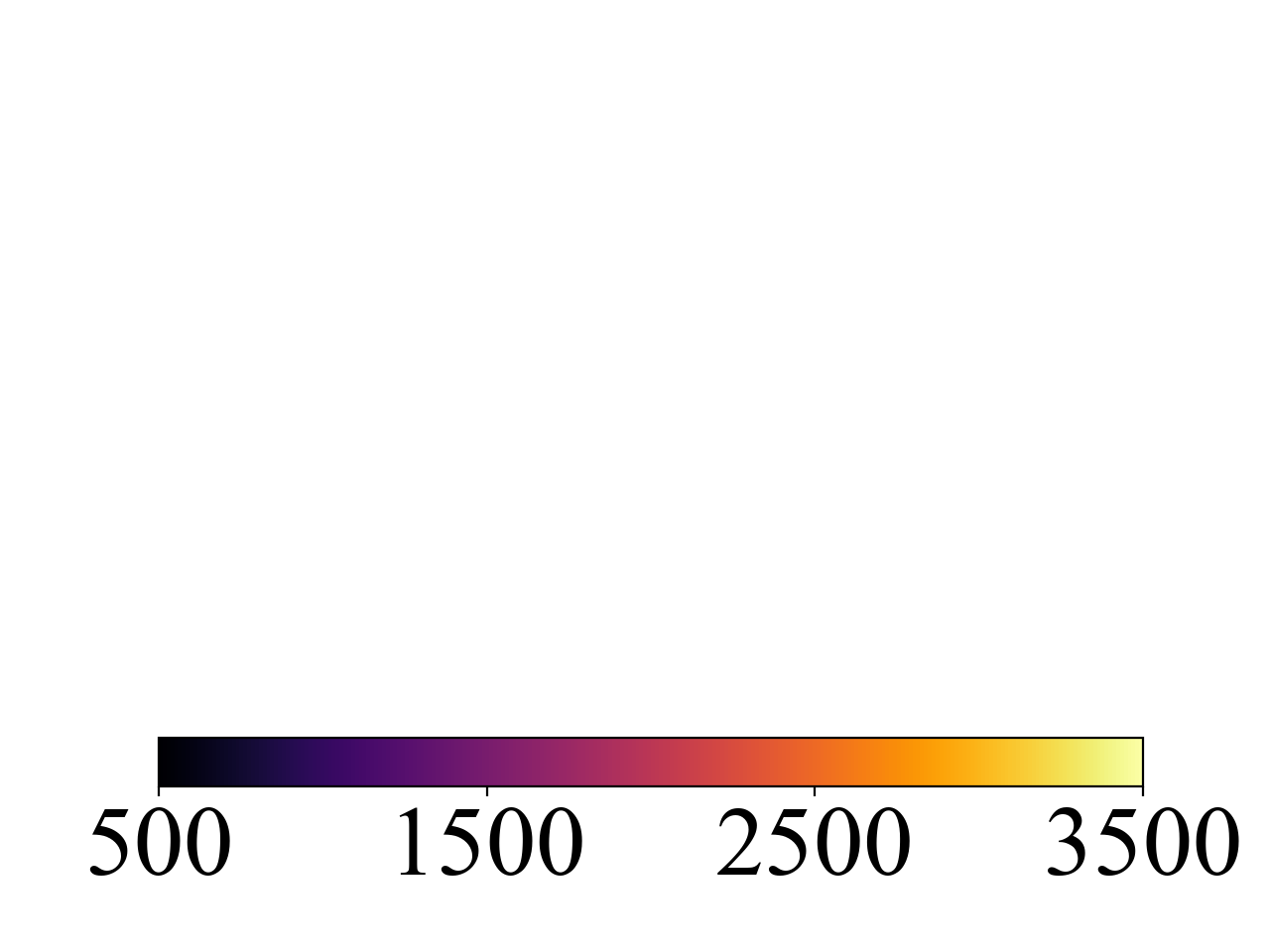}
    
    \end{tabular}
    }
    
     \vspace{-14pt}
\captionof{figure}{\textbf{Example visualizations.}  We visualize input images from ABO-500 along with our model's CLIP feature PCA components, zero-shot material segmentation, and predicted mass density. Our model makes reasonable predictions of materials across different parts of objects in 3D, allowing for grounded predictions of physical properties. }
\label{fig:example_results}
\end{table*}

\begin{table*}[!t]
    \centering
    \resizebox{\linewidth}{!}{
\setlength{\tabcolsep}{0.2em} %
\renewcommand{\arraystretch}{1.}
    \begin{tabular}{ccccc}
    
    Input RGB &
    Shore Hardness&
     ~~~~~ &
    Input RGB &
    Friction Coefficient
    \\
         \tikz{
        \node[draw=white, line width=0mm, inner sep=0pt] 
        {\includegraphics[width=.23\linewidth, trim={0 0 0 0}, clip]{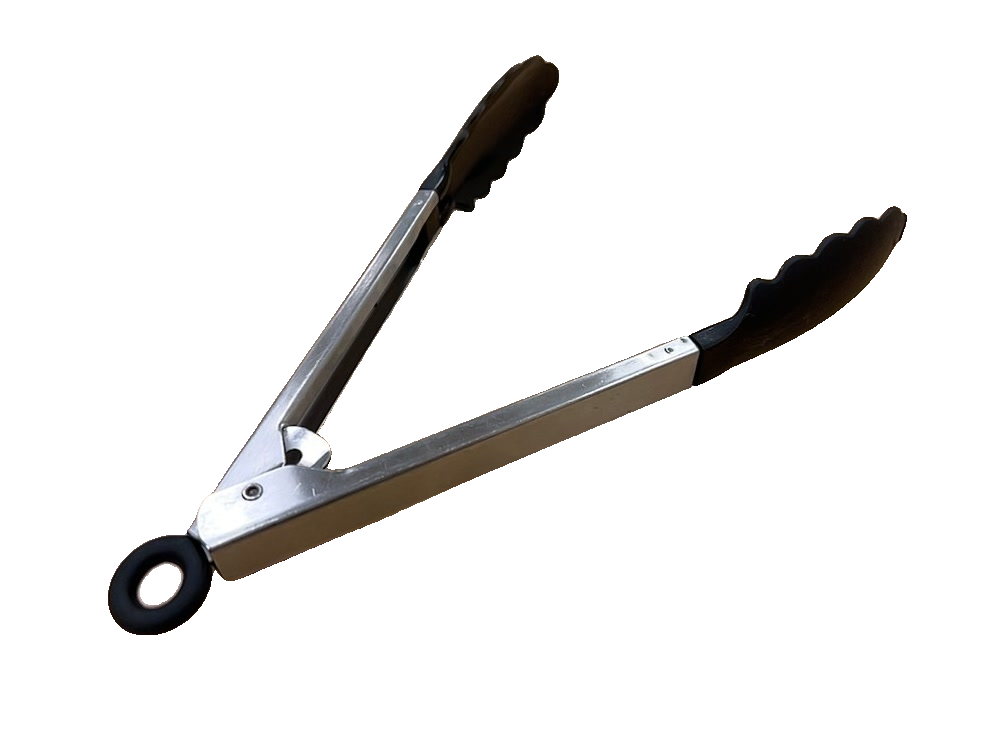}}
        }&
        \tikz{
        \node[draw=white, line width=0mm, inner sep=0pt] 
        {\includegraphics[width=.23\linewidth, trim={0 0 0 0}, clip]{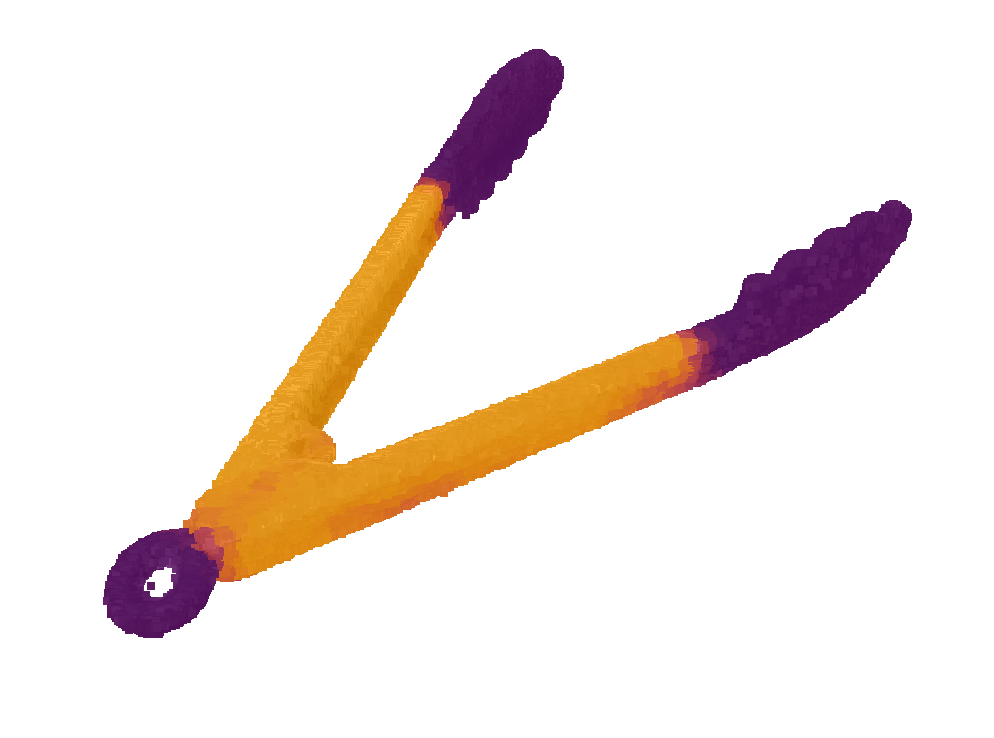}}
        }&
        
        &
        
        \tikz{
        \node[draw=white, line width=0mm, inner sep=0pt] 
        {\includegraphics[width=.23\linewidth, trim={0 0 0 0}, clip]{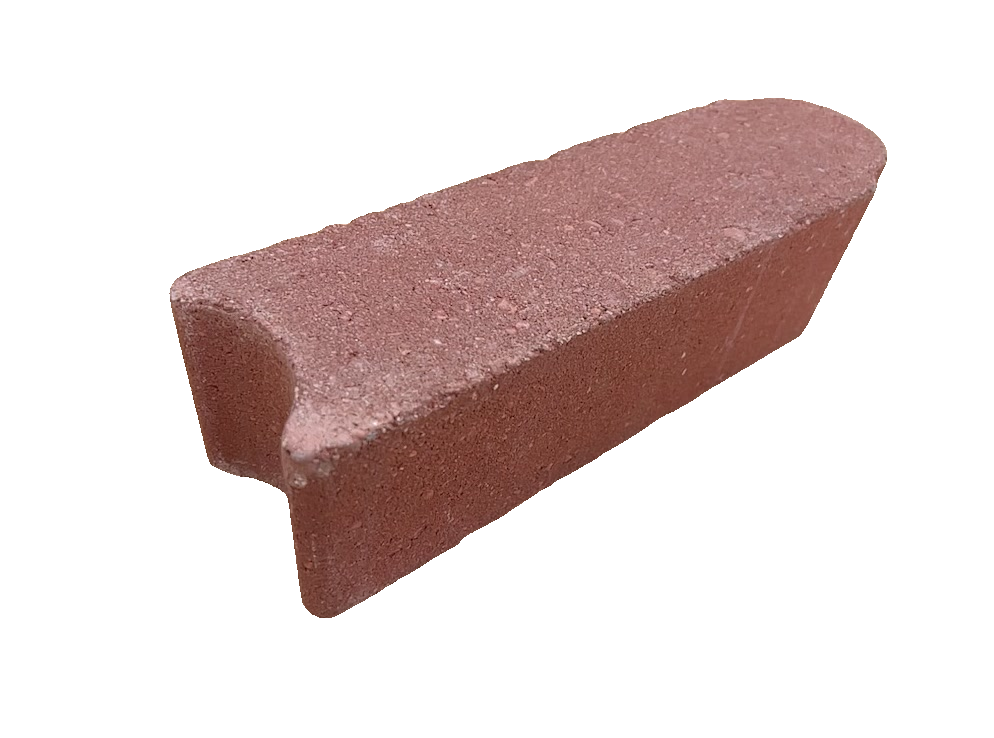}}
        }&
        \tikz{
        \node[draw=white, line width=0mm, inner sep=0pt] 
        {\includegraphics[width=.23\linewidth, trim={0 0 0 0}, clip]{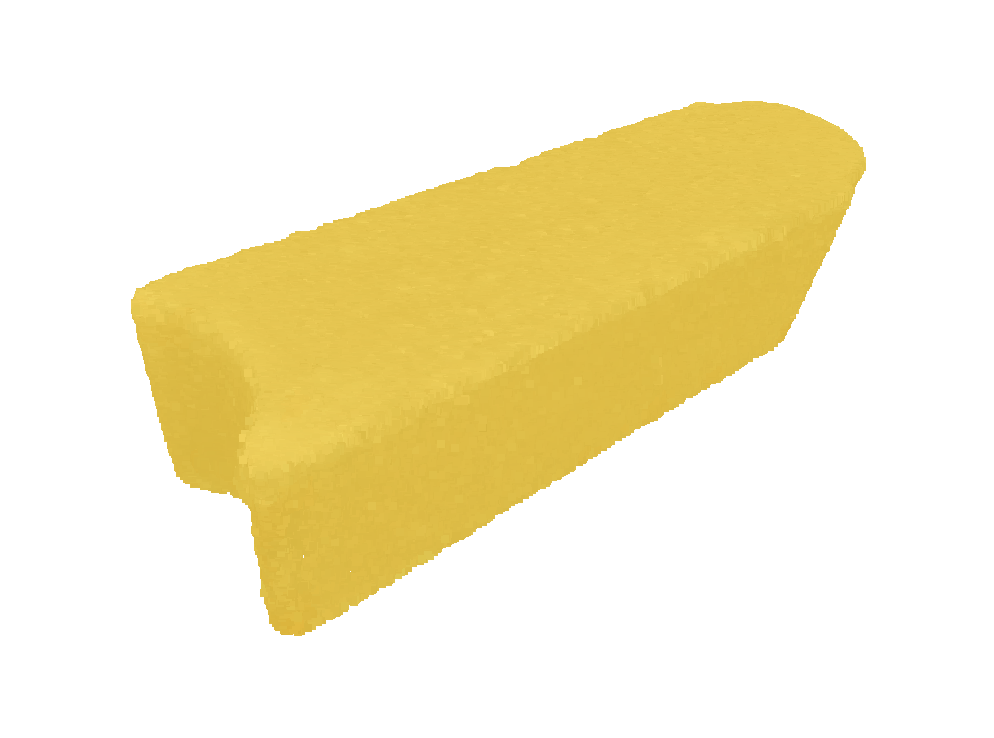}}
        }
        \vspace{-8pt}
        \\
        
         \tikz{
        \node[draw=white, line width=0mm, inner sep=0pt] 
        {\includegraphics[width=.23\linewidth, trim={0 0 0 0}, clip]{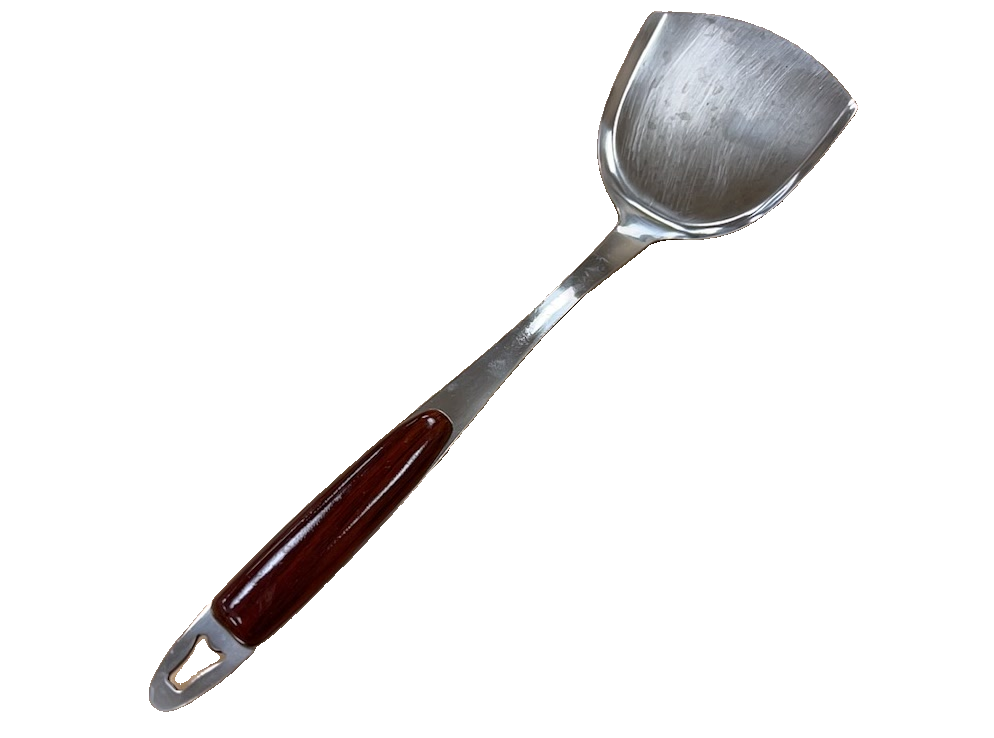}}
        }&
        \tikz{
        \node[draw=white, line width=0mm, inner sep=0pt] 
        {\includegraphics[width=.23\linewidth, trim={0 0 0 0}, clip]{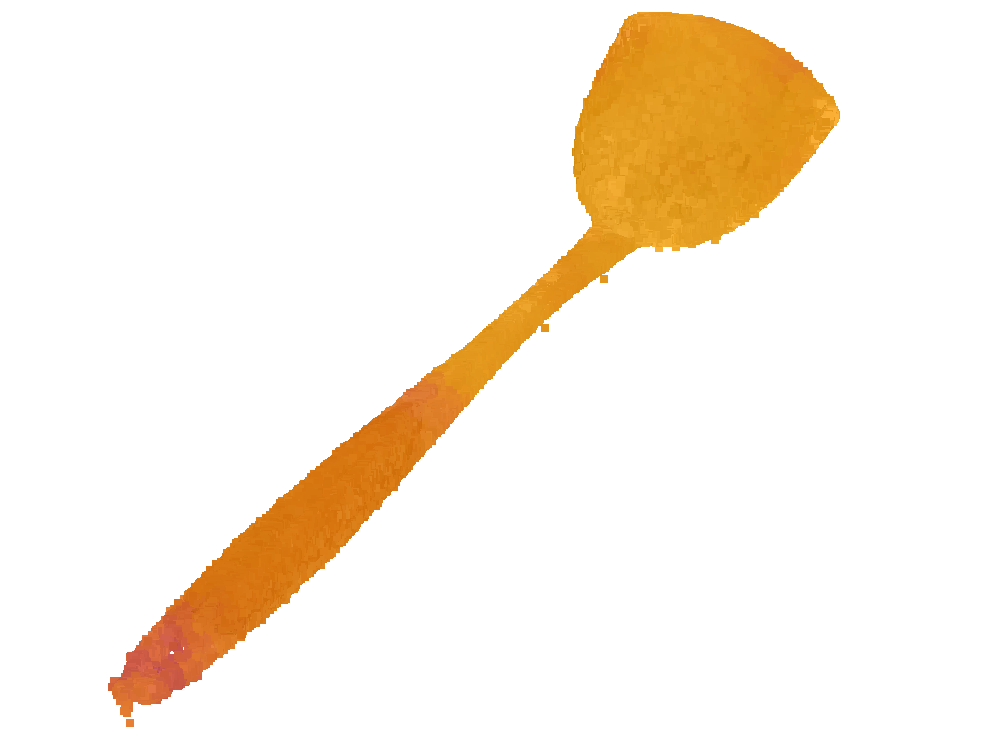}}
        }&
        
        &
        
        \tikz{
        \node[draw=white, line width=0mm, inner sep=0pt] 
        {\includegraphics[width=.23\linewidth, trim={0 0 0 0}, clip]{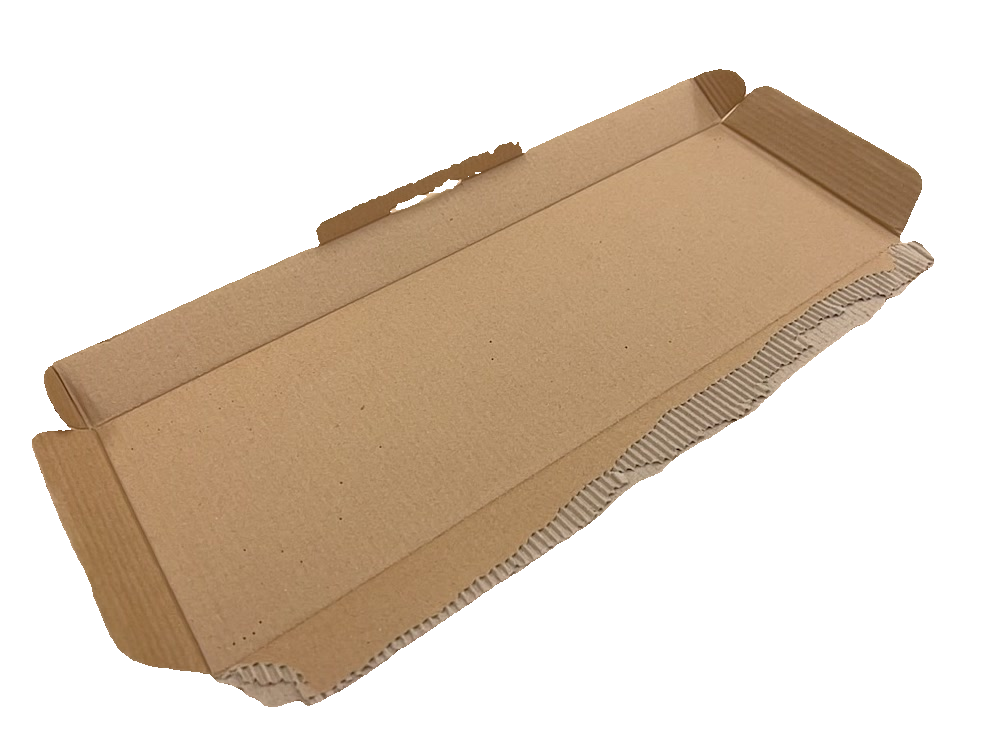}}
        }&
        \tikz{
        \node[draw=white, line width=0mm, inner sep=0pt] 
        {\includegraphics[width=.23\linewidth, trim={0 0 0 0}, clip]{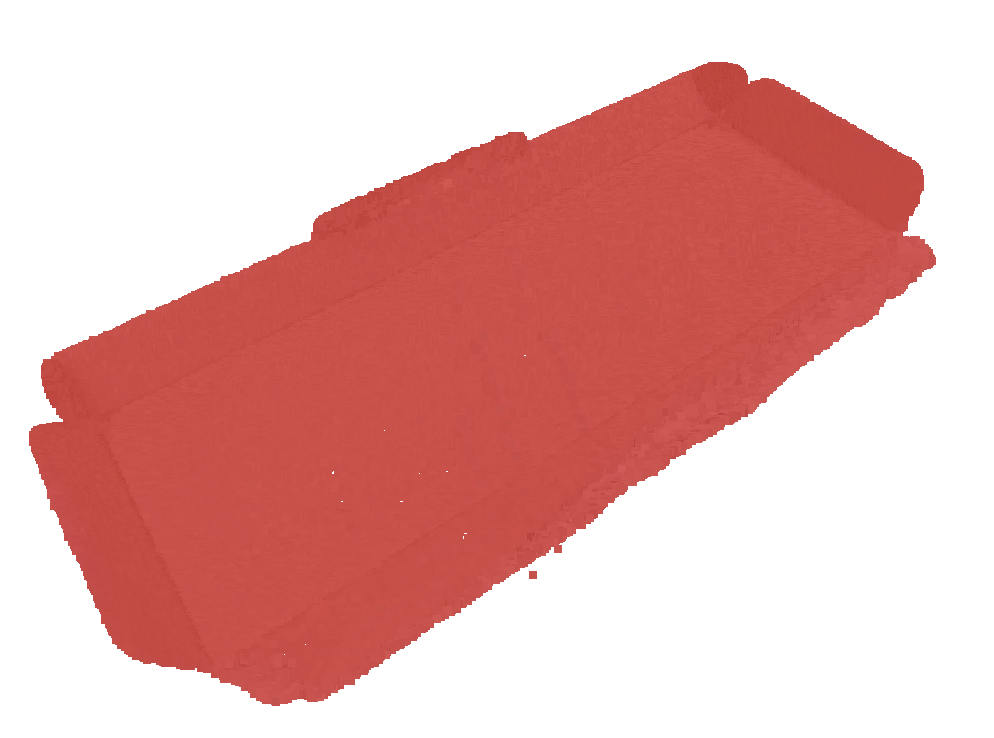}}
        }
        \\
        
         \tikz{
        \node[draw=white, line width=0mm, inner sep=0pt] 
        {\includegraphics[width=.23\linewidth, trim={0 0 0 0}, clip]{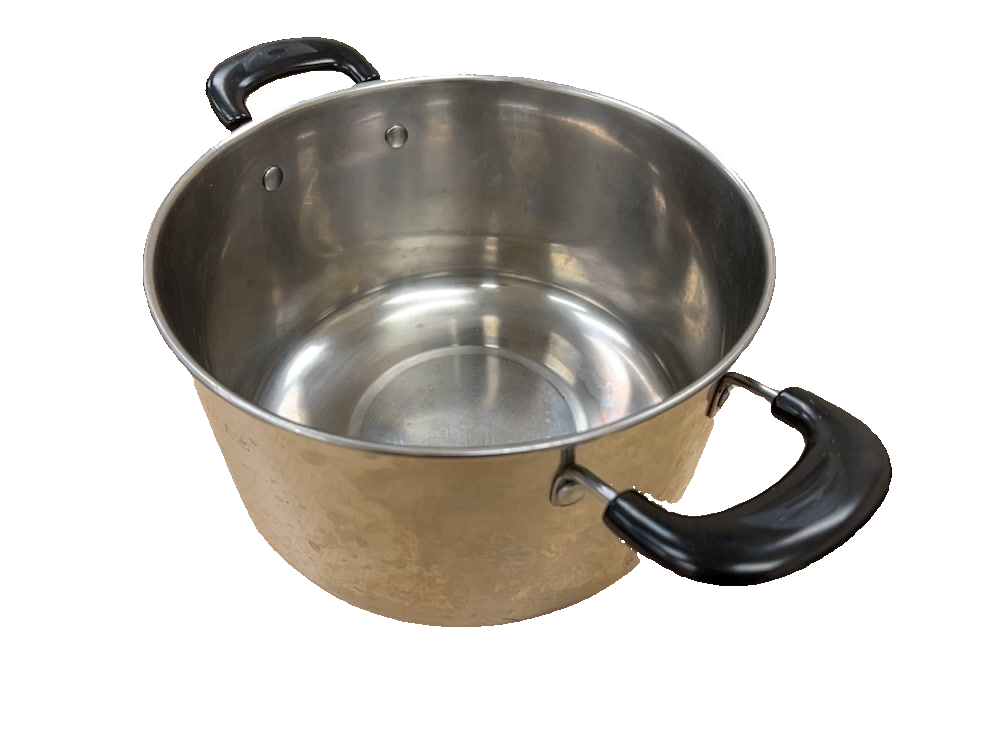}}
        }&
        \tikz{
        \node[draw=white, line width=0mm, inner sep=0pt] 
        {\includegraphics[width=.23\linewidth, trim={0 0 0 0}, clip]{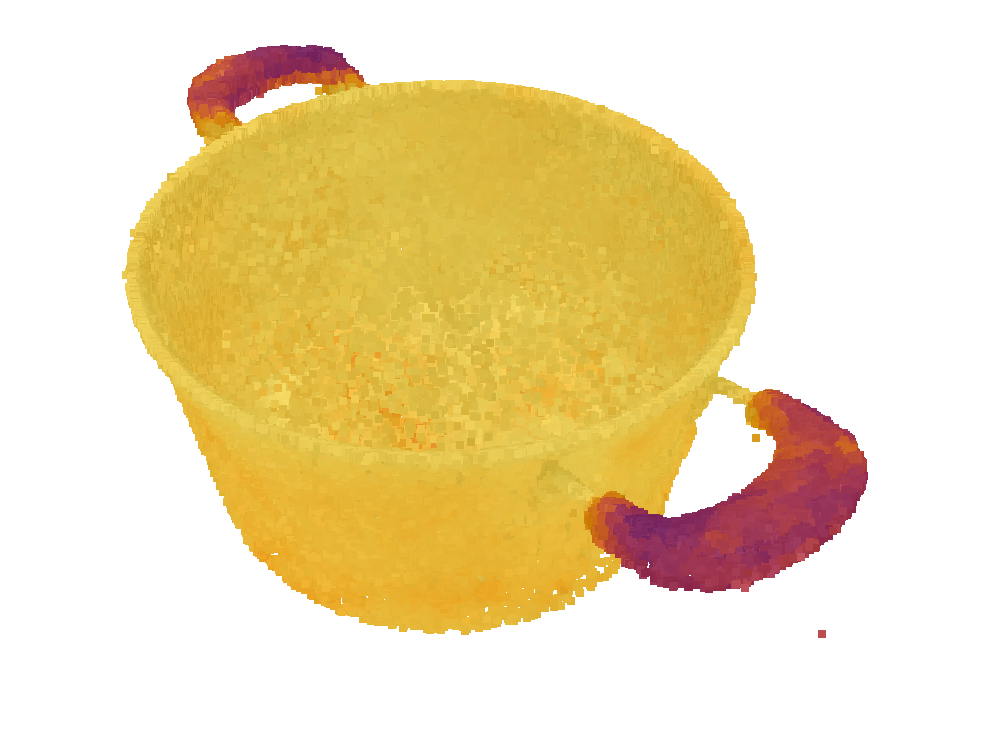}}
        }&
        
        &
        
        \tikz{
        \node[draw=white, line width=0mm, inner sep=0pt] 
        {\includegraphics[width=.23\linewidth, trim={0 0 0 0}, clip]{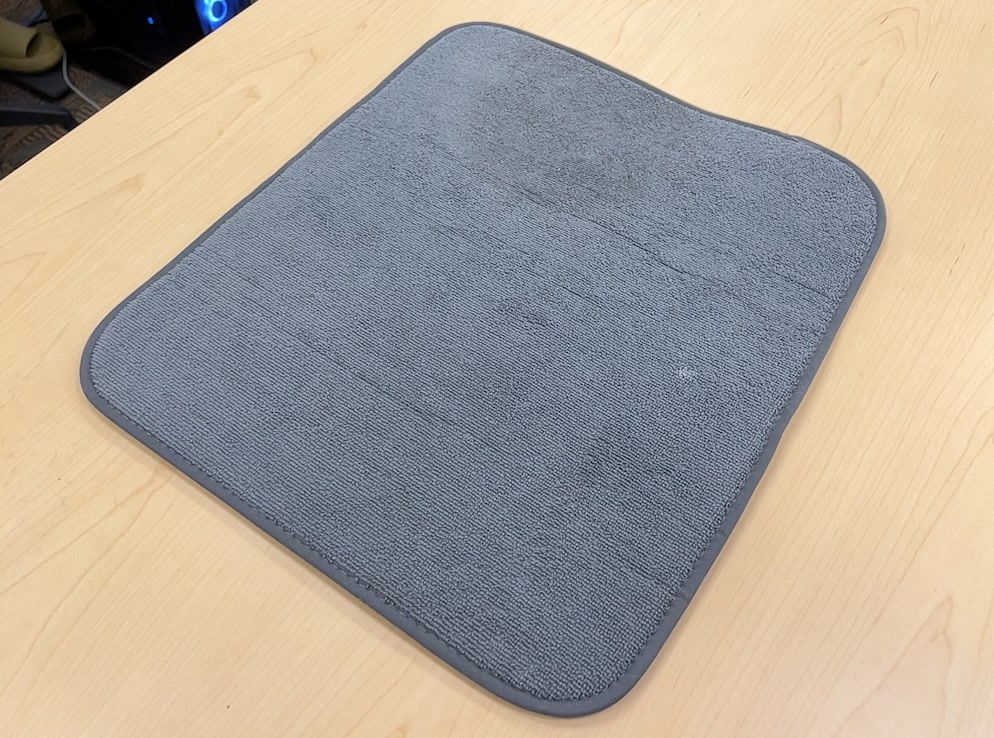}}
        }&
        \tikz{
        \node[draw=white, line width=0mm, inner sep=0pt] 
        {\includegraphics[width=.23\linewidth, trim={0 0 0 0}, clip]{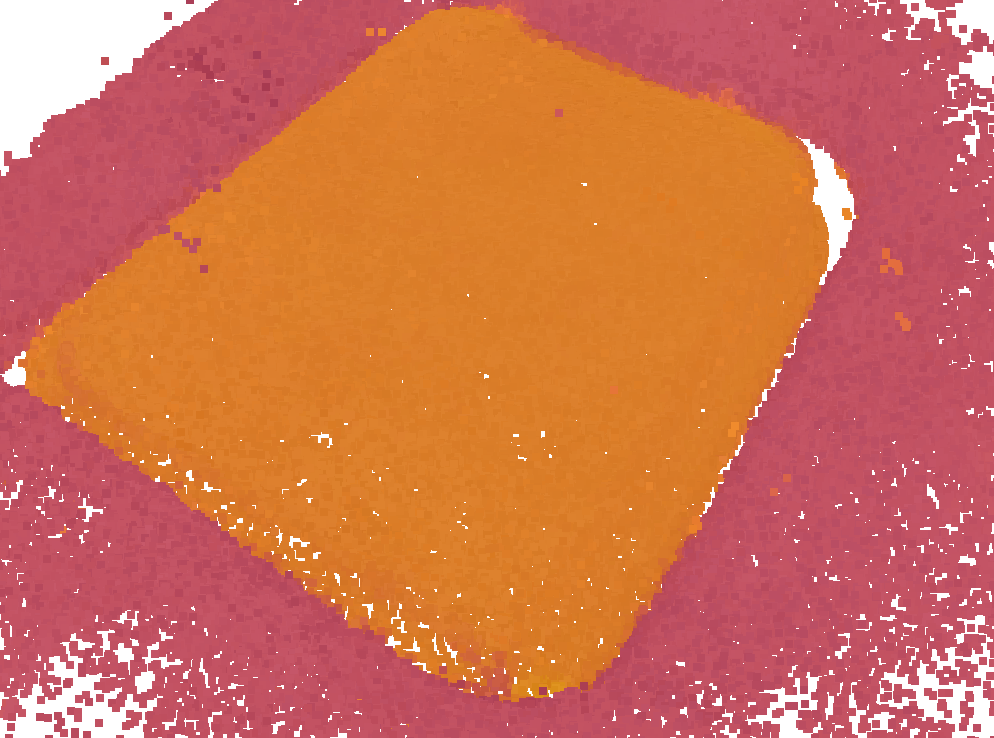}}
        }

         \\
        
       &
       \includegraphics[width=.23\linewidth, trim={0 0 0 9cm}, clip]{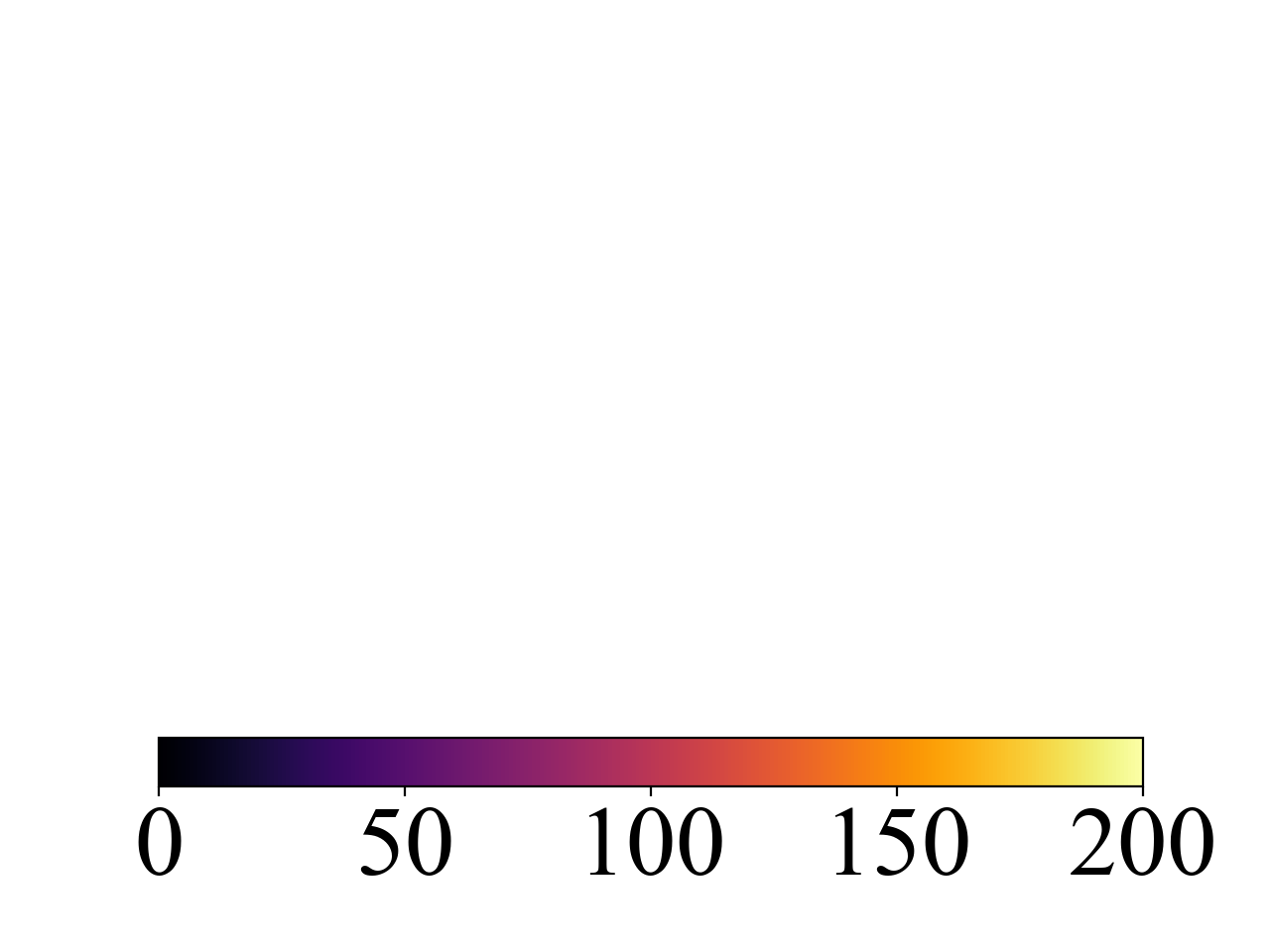}&
        
        &
        
        &
       \includegraphics[width=.23\linewidth, trim={0 0 0 9cm}, clip]{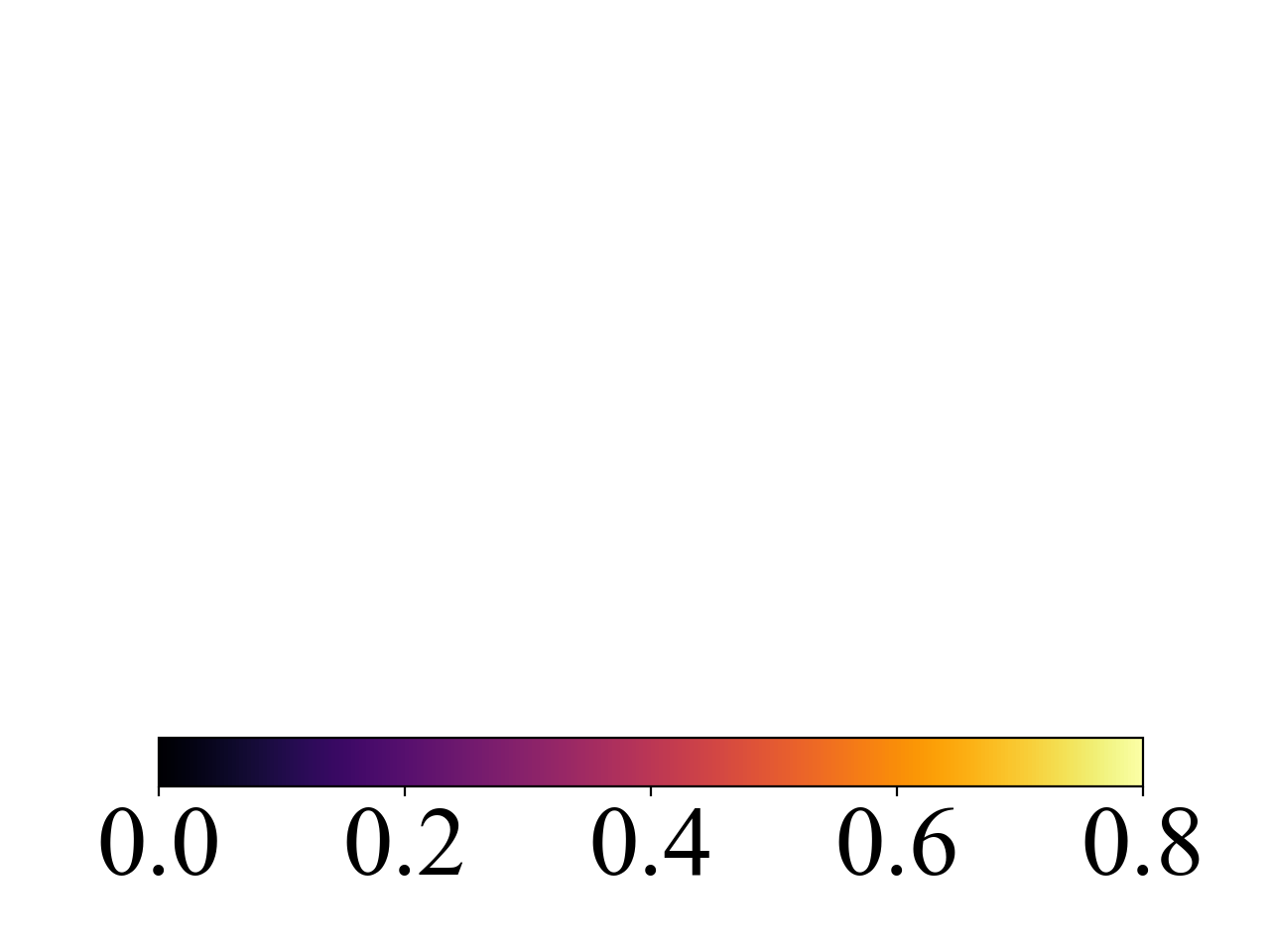}

    \end{tabular}
    }
    
     \vspace{-12pt}
\captionof{figure}{\textbf{Example predictions of different physical properties.}  We visualize predictions of hardness and friction on objects from our own collected dataset. For evaluation purposes, Shore A and Shore D hardness was combined into the same scale. The friction coefficient represents the coefficient of kinetic friction against a fabric surface. We quantitatively evaluate these predictions using a set of sparse per-point measurements (see Sec.~\ref{sec:hf_exp}).}
\vspace{-12pt}
\label{fig:hf_results}
\end{table*}

\subsection{Implementation Details}
Our NeRF directly uses the Nerfacto method from Nerfstudio~\cite{tancik2023nerfstudio} with default settings except with the near-plane for sampling set to 0.4, the far-plane set to 6.0, and the background color set to random. For our own dataset, we set the scene scale to 2.0. The camera poses are scaled per scene to fit in a $\pm 1$ box. We train each scene for 20K iterations, which takes around 8 minutes on an NVIDIA A40 GPU. 

For source point extraction, we sample $N= \num{100000}$ rays, voxel-downsample with a grid size of 0.01 (0.02 for our own dataset), and remove outliers (see supplementary for details). For language feature fusion, we use the OpenCLIP~\cite{openclip} ViT B-16 model trained on DataComp-1B and set the patch size to $P = 56$ and the occlusion threshold to 0.01. For captioning, we use BLIP-2-Flan-T5-XL~\cite{li2023blip}. For mass density (and thickness), we use GPT-3.5 Turbo~\cite{ye2023comprehensive} and set $K = 5, T = 0.1$. For friction and hardness, we use GPT-4~\cite{bubeck2023sparks} and set $K = 3, T = 0.01$. The exact prompts we used can be found in the supplementary. For mass integration, we voxel-downsample with a grid size of 0.005, carve with a grid size of 0.002, and scale the final mass by $c = 0.6$. Since our model usually returns a range of values, we take the center of the range as the final prediction.

\subsection{Mass Estimation}

\paragraph{Dataset.} 
The ABO dataset~\cite{collins2022abo} contains thousands of products sold on Amazon together with multi-view posed images, segmentation masks, mass measurements, and other product metadata. In order to create a diverse evaluation set, we created a stratified sample of 500 objects in which each ``product\_type'' (e.g. ``chair'', ``lamp'') appeared no more than 10 times. Each object/scene has 30 views facing the object with camera centers randomly distributed over a hemisphere around the object. We call this dataset ABO-500 and split the scenes randomly into 300 train / 100 val / 100 test.

\paragraph{Metrics.} 
We follow pioneering work on visual mass estimation~\cite{pmlr-v78-standley17a} and report the following metrics, where $m$ is the ground-truth mass and $\hat{m}$ is the estimated mass: 
\begin{itemize}[leftmargin=*]
    \item Absolute difference error (ADE)  $:|m - \hat{m}|$,
    \item Absolute log difference error (ALDE) $:|\ln m - \ln \hat{m} |$, 
    \item Absolute percentage error (APE) $:|\frac{m - \hat{m}}{m}|$, and
    \item Min ratio error (MnRE) $:\min(\frac{m}{\hat{m}}, \frac{\hat{m}}{m})$. 
\end{itemize}
We agree with the authors of \cite{pmlr-v78-standley17a} that MnRE is the preferred metric, because it is not biased towards models that systematically over- or under-estimate and also does not over-emphasize performance on heavier instances.

\paragraph{Baselines.} 
We compare NeRF2Physics with the following baselines on the ABO-500 dataset:
\begin{itemize}[leftmargin=*]
    \item \textbf{Image2mass}~\cite{pmlr-v78-standley17a} uses a CNN to predict mass directly from a single image and 3D bounding box dimensions. We evaluate the official model pretrained on Amazon products and use bounding boxes extracted from our source points.
    \item \textbf{2D CNN} takes a frozen ResNet50~\cite{he2016deep} pretrained on ImageNet~\cite{deng2009imagenet} and trains three addtional layers to predict mass on our dataset. We apply a negative LogSigmoid layer to ensure that the predictions are positive.
    \item \textbf{LLaVA}~\cite{liu2023llava} is a large vision-language model that is designed to follow arbitrary instructions given an image. We prompt LLaVA to estimate the mass of the object in the image (see supplementary for exact prompt).
    \footnote{We also tried to apply GPT-4V for this task but had difficulty preventing it from producing complaints about not having enough information.}
\end{itemize}
For all of the baselines, we provide the same canonical view as in our method, in which the background is set to white.

\paragraph{Qualitative Results.}
We show example visualizations of our language-embedded point cloud and material predictions in Fig.~\ref{fig:example_results}. CLIP features were converted to RGB values according to the top 3 PCA components per object. The PCA visualization suggests that the CLIP features give enough information to perform material segmentation. The material visualization shows that our method can propose reasonable candidate materials and use the CLIP features to identify the primary material in different parts of an object, such as metal in the legs of a table and wood on the tabletop. However, the boundaries of each part are not localized perfectly, and the model will often mix similar materials together (e.g.``stainless steel'' and ``aluminum''). The last column of visualizations show that sensible mass density estimates follow from the material predictions.

\begin{table}[!t]
\caption{Mass estimation on ABO-500 test set (100 objects). ADE is measured in kilograms. {\bf Bold}: best model. }
\small
\centering
\def\arraystretch{1.1}
\scalebox{0.9}{
\begin{tabular}{lcccc}
\toprule
Method  & ADE ($\downarrow$) & ALDE ($\downarrow$) & APE ($\downarrow$) & MnRE ($\uparrow$)  \\
\midrule
Image2mass \cite{pmlr-v78-standley17a} & 12.496 & 1.792 & \textbf{0.976} & 0.341 \\ 
2D CNN  & 15.431 & 1.609 & 14.459 & 0.362 \\ 
LLaVA \cite{liu2023llava} & 17.328 & 1.893  & 1.837 & 0.306 \\
\midrule

Ours & \textbf{8.730} & \textbf{0.771} & 1.061 & \textbf{0.552} \\

\bottomrule
\label{tab:diverse_results}
\end{tabular}
}
\vspace{-16pt}
\end{table}

\paragraph{Quantitative Results.}
We report test-set mass estimation metrics on ABO-500 in Tab.~\ref{tab:diverse_results}. Mass predictions for all models were clipped to be between 0.01 and 100 kilograms. The image2mass~\cite{pmlr-v78-standley17a} pretrained model performs poorly since it does not generalize well to objects larger than those found in its training data. The 2D CNN baseline also did not perform well -- it failed to learn meaningful patterns and tended to predict towards the mean of the dataset. The LLaVA~\cite{liu2023llava} model usually gives answers that are not metrically precise (e.g. 1 kg or 10 kg), despite extensive prompt engineering. Our zero-shot method's predictions outperform these baselines by a large margin in all metrics except APE. We note that APE is heavily biased towards models that underestimate, as it is dominated by overestimates on small objects.

\begin{table}[!t]
\caption{Ablation study for mass estimation on ABO-500 val set (100 objects). {\bf Bold}: best model.}
\small
\centering
\def\arraystretch{1.1}
\scalebox{0.9}{
\begin{tabular}{lcccc}
\toprule
Method  & ADE ($\downarrow$) & ALDE ($\downarrow$) & APE ($\downarrow$) & MnRE ($\uparrow$)  \\
\midrule
No thickness & 18.587 & 0.749 & 1.364 & 0.552 \\
Retrieval ($T \rightarrow 0$) & 12.266 & 0.780 & \textbf{0.801} & 0.536 \\
Uniform CLIP & 10.396 & 0.637 & 1.102 & 0.597 \\
\midrule

Ours & \textbf{9.786} & \textbf{0.610} & 0.931 & \textbf{0.609} \\

\bottomrule
\label{tab:ablation}
\end{tabular}
}
\vspace{-16pt}
\end{table}

\paragraph{Ablation Studies.} 
We perform ablations on various aspects of our method in Tab.~\ref{tab:ablation}. First, we remove the thickness estimation step and integrate over occupied voxel produced by depth-based carving. This results in consistent overestimation since it ignores the fact that many objects have empty space in their interior. Next, we examine the effect of performing kernel regression instead of just retrieving the most likely material per point (effectively setting the temperature $T$ to zero). Here, the retrieval performs worse because there is inherent uncertainty in predicting materials based on visual appearance. Lastly, we evaluate the use of a single global CLIP embedding of the canonical view instead of fused patch embeddings, which gives a uniform prediction across the whole object. We find that the performance is only slightly worse, suggesting that the total mass for most objects is dominated by a single material.

\subsection{Friction and Hardness Estimation} \label{sec:hf_exp}
\paragraph{Dataset.} The task of mass estimation does not directly evaluate the ability of our model to perform dense prediction of different physical property values within the same object. To the best of our knowledge, there does not exist a realistic dataset with images and paired measurements suitable for this purpose. Thus, we collect our own dataset containing 15 household objects across 13 scenes with real-world multi-view images and paired measurements of per-point kinetic friction coefficient and Shore hardness. The images and poses were captured using Polycam on an iPhone 13 Pro, with a median of 82 per scene. Coefficient of kinetic friction was collected on 6 surfaces using an iOLab with fabric pads attached, averaging over 10 trials. Shore hardness was collected at 31 points using Gain Express A/D durometers, averaging over 3 trials with Shore D being used when the Shore A reading was above 90. Each point's location is annotated as pixel coordinates in an image. Grounding SAM~\cite{kirillov2023segment, liu2023grounding} was used to obtain object masks.

Note that Shore A and Shore D durometers use different indenters and thus do not measure exactly the same physical property~\cite{mohamed2003uncertainty}. However, for evaluation purposes, we combine the measurements into a single scale from 0-200, where the Shore A measurements lie in the 0-100 range and Shore D measurements lie in the 100-200 range.

\paragraph{Metrics.} We report the same metrics as before, along with an additional metric of Pairwise Relationship Accuracy (PRA), defined as the classification accuracy of the predicted relationships (greater than or less than) between every pair of points. This metric focuses on relative comparisons and is thus more robust to measurement noise, which is especially significant for the hardness measurements due to local deformations in the object surface around each point.

\paragraph{Baselines.} There are no existing methods for predictions of arbitrary physical properties from images, so we design the following baselines for comparison:
\begin{itemize}[leftmargin=*]
    \item \textbf{GPT-4V}~\cite{yang2023dawn} is a large vision-language model that can accept masks in its prompt. For each point, we provide GPT-4V with the associated image and a mask highlighting its pixel location, and ask it to estimate the physical property at that point.
    \item \textbf{CLIP} refers to using a global CLIP embedding of the canonical view instead of fused patch features in our method. This was also considered in our ablations above. 
\end{itemize}
We instruct each LLM to choose Shore A/D hardness based on which is more appropriate for the material in question.

\begin{table}[!t]
\caption{Estimation of per-point Shore hardness on the real-world in-house collected dataset (31 points, 11 objects). {\bf Bold}: best model. }
\small
\centering
\def\arraystretch{1.1}
\scalebox{0.88}{
\begin{tabular}{lccccc}
\toprule
Method  & ADE ($\downarrow$) & ALDE ($\downarrow$) & APE ($\downarrow$) & MnRE ($\uparrow$) & PRA ($\uparrow$)  \\
\midrule
GPT-4V  & \textbf{32.752} & 0.330 & 0.304 & 0.758 & 0.609\\
CLIP &  32.857 &  \textbf{0.294} & \textbf{0.266 }& \textbf{0.774} & 0.647 \\
Ours & 34.295 & 0.315 &  0.276 & 0.765 & \textbf{0.710}\\
\bottomrule
\label{tab:hardness_results}
\end{tabular}
}
\vspace{-16pt}
\end{table}

\begin{table}[!t]
\caption{Estimation of per-point kinetic friction coefficient on the in-house collected dataset (6 points, 6 objects). {\bf Bold}: best model. }
\small
\centering
\def\arraystretch{1.1}
\scalebox{0.88}{
\begin{tabular}{lccccc}
\toprule
Method  & ADE ($\downarrow$) & ALDE ($\downarrow$) & APE ($\downarrow$) & MnRE ($\uparrow$) & PRA ($\uparrow$) \\
\midrule
GPT-4V & 0.209 & 0.430 & 0.549 & 0.692  & 0.467 \\
CLIP & 0.222 &  0.455 & 0.602 & 0.654 & 0.533 \\
Ours & \textbf{0.155} & \textbf{0.321} & \textbf{0.360} & \textbf{0.736} & \textbf{0.800} \\
\bottomrule
\label{tab:friction_results}
\end{tabular}
}
\vspace{-16pt}
\end{table}

\paragraph{Qualitative Results.} 
We show example predictions of hardness and friction from our model in Fig.~\ref{fig:hf_results}. Again, the model is able to distinguish different materials and derive reasonable physical property estimates from them, even for unusual objects such as the ripped piece of cardboard. In addition, the model is fairly robust to errors in the geometry from NeRF, thanks to our feature fusion strategy. The example with the bath mat demonstrates that our method can be applied with or without object segmentation masks.

\paragraph{Quantitative Results.} 
We report quantitative evaluation metrics for hardness prediction in Tab.~\ref{tab:hardness_results} and for friction prediction in Tab.~\ref{tab:friction_results}. For hardness, we observe that all three models perform similarly across most of the metrics, but ours achieves the highest PRA, indicating that it localizes different materials more precisely than the other models.  For friction, we find that our model outperforms the others by a wide margin in all of the metrics. GPT-4V performs similarly with the uniform CLIP model, suggesting that it has trouble distinguishing between different surfaces within the same scene. Also note that GPT-4V must run on each individual query point, which is extremely computationally expensive. In contrast, once the feature field and candidate materials for our model have been prepared, thousands of points can be queried with little computational cost.

\begin{table}[t]
    \centering
\setlength{\tabcolsep}{0.1em} %

\includegraphics[width=\linewidth,   clip]{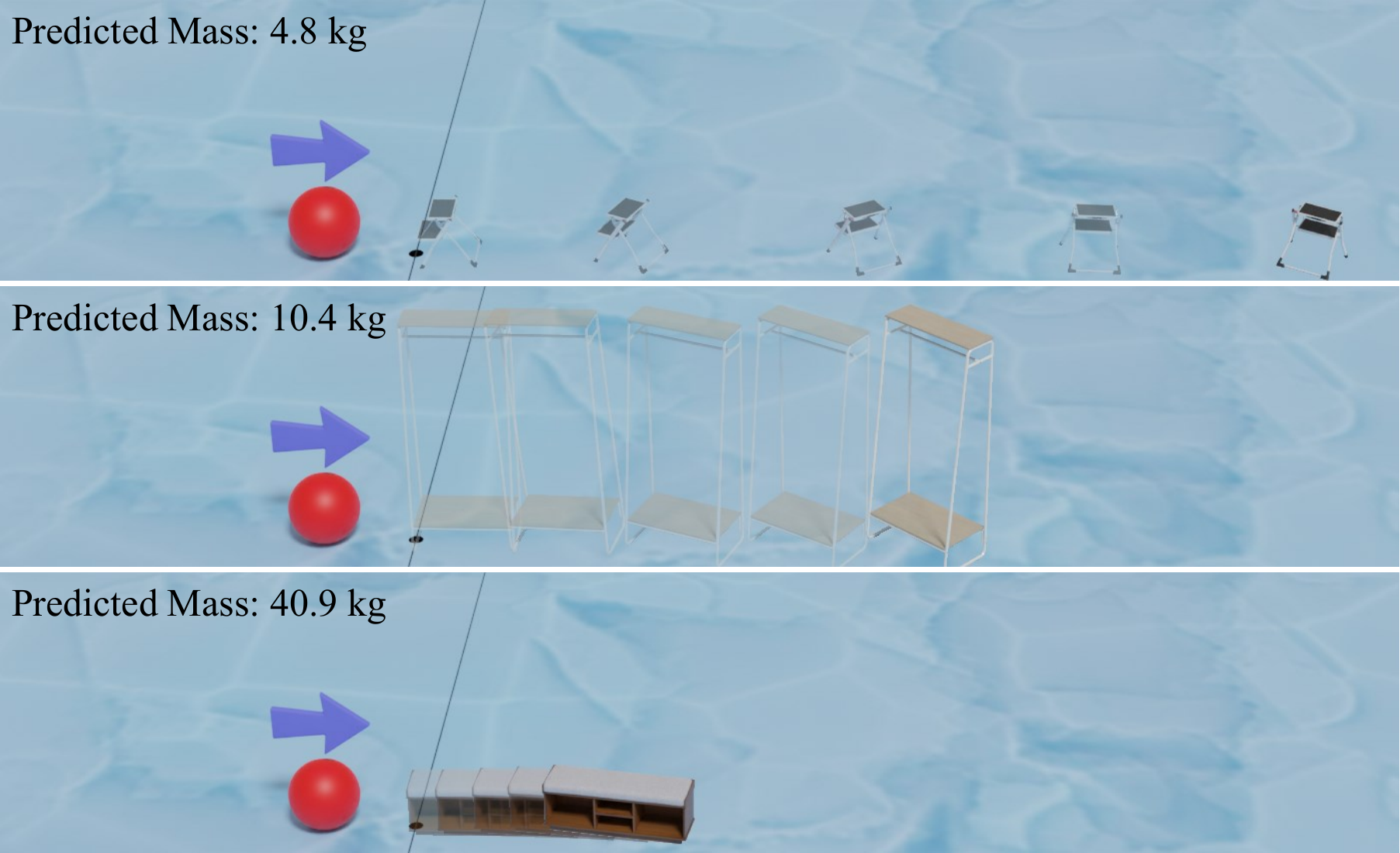}
    \vspace{-16pt}
\captionof{figure}{\textbf{Digital twins with realistic physical properties.} We show that realistic physical interactions can be simulated using mass-aware digital twins created by NeRF2Physics. In each example trajectory visualization here, the ball hits the object with the same initial momentum, and friction is zero.}
\vspace{-10pt}
\label{fig:simulation}
\end{table}

\subsection{Applications}

NeRF2Physics can be applied to create physically realistic digital twins for immersive computing and content creation (Fig.~\ref{fig:simulation}). Improved physical property understanding is also crucial for advancing embodied AI and robot simulation. Another application is estimating crop biomass~\cite{cover_crop}, which is important for agriculture but labor-intensive and destructive to measure manually.   
\section{Conclusion}

We presented NeRF2Physics, a novel method for dense prediction of physical properties from a collection of images. Our method fuses vision-language embeddings into a 3D point cloud and leverages LLMs to provide material information, enabling zero-shot estimation of any physical property for any object in the open world. Experimental results demonstrate that our method outperforms baselines on estimation of mass, hardness, and friction coefficients across a variety of objects. In the future, our approach may be improved by incorporating prior knowledge to reason about materials in internal object parts that cannot be seen.

\paragraph{Acknowledgments. } This project is supported by the Intel AI SRS gift, the IBM IIDAI Grant, the Insper-Illinois Innovation Grant, the NCSA Faculty Fellowship, the Agroecosystem Sustainability Center at UIUC, and NSF Awards \#2331878, \#2340254, and \#2312102. We greatly appreciate NCSA for providing computing resources. 

{
    \small
    \bibliographystyle{ieeenat_fullname}
    \bibliography{main}

\begin{thebibliography}{44}
\providecommand{\natexlab}[1]{#1}
\providecommand{\url}[1]{\texttt{#1}}
\expandafter\ifx\csname urlstyle\endcsname\relax
  \providecommand{\doi}[1]{doi: #1}\else
  \providecommand{\doi}{doi: \begingroup \urlstyle{rm}\Url}\fi

\bibitem[cov()]{cover_crop}
Estimating cover crop biomass.
\newblock
  \url{https://www.nrcs.usda.gov/sites/default/files/2022-09/EstBiomassCoverCrops_Sept2018.pdf}.
\newblock Accessed: 2023-11-17.

\bibitem[Adelson(2001)]{adelson2001seeing}
Edward~H Adelson.
\newblock On seeing stuff: the perception of materials by humans and machines.
\newblock In \emph{Human vision and electronic imaging VI}. SPIE, 2001.

\bibitem[Bell et~al.(2014)Bell, Bala, and Snavely]{bell14intrinsic}
Sean Bell, Kavita Bala, and Noah Snavely.
\newblock Intrinsic images in the wild.
\newblock \emph{SIGGRAPH}, 2014.

\bibitem[Bell et~al.(2015)Bell, Upchurch, Snavely, and Bala]{bell2015material}
Sean Bell, Paul Upchurch, Noah Snavely, and Kavita Bala.
\newblock Material recognition in the wild with the materials in context
  database.
\newblock In \emph{CVPR}, 2015.

\bibitem[Bubeck et~al.(2023)Bubeck, Chandrasekaran, Eldan, Gehrke, Horvitz,
  Kamar, Lee, Lee, Li, Lundberg, et~al.]{bubeck2023sparks}
S{\'e}bastien Bubeck, Varun Chandrasekaran, Ronen Eldan, Johannes Gehrke, Eric
  Horvitz, Ece Kamar, Peter Lee, Yin~Tat Lee, Yuanzhi Li, Scott Lundberg,
  et~al.
\newblock Sparks of artificial general intelligence: Early experiments with
  gpt-4.
\newblock \emph{arXiv preprint arXiv:2303.12712}, 2023.

\bibitem[Chen et~al.(2023)Chen, Xia, Ichter, Rao, Gopalakrishnan, Ryoo, Stone,
  and Kappler]{chen2023open}
Boyuan Chen, Fei Xia, Brian Ichter, Kanishka Rao, Keerthana Gopalakrishnan,
  Michael~S Ryoo, Austin Stone, and Daniel Kappler.
\newblock Open-vocabulary queryable scene representations for real world
  planning.
\newblock In \emph{ICRA}, 2023.

\bibitem[Collins et~al.(2022)Collins, Goel, Deng, Luthra, Xu, Gundogdu, Zhang,
  Yago~Vicente, Dideriksen, Arora, Guillaumin, and Malik]{collins2022abo}
Jasmine Collins, Shubham Goel, Kenan Deng, Achleshwar Luthra, Leon Xu, Erhan
  Gundogdu, Xi Zhang, Tomas~F Yago~Vicente, Thomas Dideriksen, Himanshu Arora,
  Matthieu Guillaumin, and Jitendra Malik.
\newblock Abo: Dataset and benchmarks for real-world 3d object understanding.
\newblock In \emph{CVPR}, 2022.

\bibitem[Deng et~al.(2009)Deng, Dong, Socher, Li, Li, and
  Fei-Fei]{deng2009imagenet}
Jia Deng, Wei Dong, Richard Socher, Li-Jia Li, Kai Li, and Li Fei-Fei.
\newblock Imagenet: A large-scale hierarchical image database.
\newblock In \emph{CVPR}, 2009.

\bibitem[Fleming(2014)]{FLEMING201462}
Roland~W. Fleming.
\newblock Visual perception of materials and their properties.
\newblock \emph{Vision Research}, 2014.

\bibitem[Fleming et~al.(2013)Fleming, Wiebel, and
  Gegenfurtner]{fleming2013perceptual}
Roland~W Fleming, Christiane Wiebel, and Karl Gegenfurtner.
\newblock Perceptual qualities and material classes.
\newblock \emph{Journal of vision}, 2013.

\bibitem[Fragkiadaki et~al.(2016)Fragkiadaki, Agrawal, Levine, and
  Malik]{fragkiadaki2015learning}
Katerina Fragkiadaki, Pulkit Agrawal, Sergey Levine, and Jitendra Malik.
\newblock Learning visual predictive models of physics for playing billiards.
\newblock \emph{ICLR}, 2016.

\bibitem[Gadre et~al.(2022)Gadre, Wortsman, Ilharco, Schmidt, and
  Song]{gadre2022clip}
Samir~Yitzhak Gadre, Mitchell Wortsman, Gabriel Ilharco, Ludwig Schmidt, and
  Shuran Song.
\newblock Clip on wheels: Zero-shot object navigation as object localization
  and exploration.
\newblock \emph{arXiv preprint arXiv:2203.10421}, 2022.

\bibitem[He et~al.(2016)He, Zhang, Ren, and Sun]{he2016deep}
Kaiming He, Xiangyu Zhang, Shaoqing Ren, and Jian Sun.
\newblock Deep residual learning for image recognition.
\newblock In \emph{CVPR}, 2016.

\bibitem[Hong et~al.(2023)Hong, Lin, Du, Chen, Tenenbaum, and Gan]{hong20233d}
Yining Hong, Chunru Lin, Yilun Du, Zhenfang Chen, Joshua~B Tenenbaum, and
  Chuang Gan.
\newblock 3d concept learning and reasoning from multi-view images.
\newblock In \emph{CVPR}, 2023.

\bibitem[Ilharco et~al.(2021)Ilharco, Wortsman, Wightman, Gordon, Carlini,
  Taori, Dave, Shankar, Namkoong, Miller, Hajishirzi, Farhadi, and
  Schmidt]{openclip}
Gabriel Ilharco, Mitchell Wortsman, Ross Wightman, Cade Gordon, Nicholas
  Carlini, Rohan Taori, Achal Dave, Vaishaal Shankar, Hongseok Namkoong, John
  Miller, Hannaneh Hajishirzi, Ali Farhadi, and Ludwig Schmidt.
\newblock Open{CLIP}, 2021.
\newblock https://github.com/mlfoundations/open\_clip.

\bibitem[Jatavallabhula et~al.(2023)Jatavallabhula, Kuwajerwala, Gu, Omama,
  Chen, Li, Iyer, Saryazdi, Keetha, Tewari,
  et~al.]{jatavallabhula2023conceptfusion}
Krishna~Murthy Jatavallabhula, Alihusein Kuwajerwala, Qiao Gu, Mohd Omama, Tao
  Chen, Shuang Li, Ganesh Iyer, Soroush Saryazdi, Nikhil Keetha, Ayush Tewari,
  et~al.
\newblock Conceptfusion: Open-set multimodal 3d mapping.
\newblock \emph{arXiv preprint arXiv:2302.07241}, 2023.

\bibitem[Kerr et~al.(2023)Kerr, Kim, Goldberg, Kanazawa, and Tancik]{lerf2023}
Justin Kerr, Chung~Min Kim, Ken Goldberg, Angjoo Kanazawa, and Matthew Tancik.
\newblock {LERF}: Language embedded radiance fields.
\newblock In \emph{ICCV}, 2023.

\bibitem[Kirillov et~al.(2023)Kirillov, Mintun, Ravi, Mao, Rolland, Gustafson,
  Xiao, Whitehead, Berg, Lo, et~al.]{kirillov2023segment}
Alexander Kirillov, Eric Mintun, Nikhila Ravi, Hanzi Mao, Chloe Rolland, Laura
  Gustafson, Tete Xiao, Spencer Whitehead, Alexander~C Berg, Wan-Yen Lo, et~al.
\newblock Segment anything.
\newblock \emph{arXiv preprint arXiv:2304.02643}, 2023.

\bibitem[Li et~al.(2022{\natexlab{a}})Li, Weinberger, Belongie, Koltun, and
  Ranftl]{li2022languagedriven}
Boyi Li, Kilian~Q Weinberger, Serge Belongie, Vladlen Koltun, and Rene Ranftl.
\newblock Language-driven semantic segmentation.
\newblock In \emph{ICLR}, 2022{\natexlab{a}}.

\bibitem[Li et~al.(2023)Li, Li, Savarese, and Hoi]{li2023blip}
Junnan Li, Dongxu Li, Silvio Savarese, and Steven Hoi.
\newblock Blip-2: Bootstrapping language-image pre-training with frozen image
  encoders and large language models.
\newblock \emph{arXiv preprint arXiv:2301.12597}, 2023.

\bibitem[Li et~al.(2022{\natexlab{b}})Li, Qiao, Chen, Jatavallabhula, Lin,
  Jiang, and Gan]{li2022pac}
Xuan Li, Yi-Ling Qiao, Peter~Yichen Chen, Krishna~Murthy Jatavallabhula, Ming
  Lin, Chenfanfu Jiang, and Chuang Gan.
\newblock {PAC-NeRF}: Physics augmented continuum neural radiance fields for
  geometry-agnostic system identification.
\newblock In \emph{ICLR}, 2022{\natexlab{b}}.

\bibitem[Liu et~al.(2023{\natexlab{a}})Liu, Li, Wu, and Lee]{liu2023llava}
Haotian Liu, Chunyuan Li, Qingyang Wu, and Yong~Jae Lee.
\newblock Visual instruction tuning.
\newblock In \emph{NeurIPS}, 2023{\natexlab{a}}.

\bibitem[Liu et~al.(2023{\natexlab{b}})Liu, Zeng, Ren, Li, Zhang, Yang, Li,
  Yang, Su, Zhu, et~al.]{liu2023grounding}
Shilong Liu, Zhaoyang Zeng, Tianhe Ren, Feng Li, Hao Zhang, Jie Yang, Chunyuan
  Li, Jianwei Yang, Hang Su, Jun Zhu, et~al.
\newblock Grounding dino: Marrying dino with grounded pre-training for open-set
  object detection.
\newblock \emph{arXiv preprint arXiv:2303.05499}, 2023{\natexlab{b}}.

\bibitem[Michel et~al.(2022)Michel, Bar-On, Liu, Benaim, and
  Hanocka]{michel2022text2mesh}
Oscar Michel, Roi Bar-On, Richard Liu, Sagie Benaim, and Rana Hanocka.
\newblock Text2mesh: Text-driven neural stylization for meshes.
\newblock In \emph{CVPR}, 2022.

\bibitem[Mildenhall et~al.(2020)Mildenhall, Srinivasan, Tancik, Barron,
  Ramamoorthi, and Ng]{mildenhall2020nerf}
Ben Mildenhall, Pratul~P Srinivasan, Matthew Tancik, Jonathan~T Barron, Ravi
  Ramamoorthi, and Ren Ng.
\newblock Nerf: Representing scenes as neural radiance fields for view
  synthesis.
\newblock In \emph{ECCV}, 2020.

\bibitem[Mohamed and Aggag(2003)]{mohamed2003uncertainty}
Magdy~I Mohamed and Gamal~A Aggag.
\newblock Uncertainty evaluation of shore hardness testers.
\newblock \emph{Measurement}, 33\penalty0 (3):\penalty0 251--257, 2003.

\bibitem[Novack et~al.(2023)Novack, McAuley, Lipton, and Garg]{novack2023chils}
Zachary Novack, Julian McAuley, Zachary Lipton, and Saurabh Garg.
\newblock Chils: Zero-shot image classification with hierarchical label sets.
\newblock In \emph{ICML}, 2023.

\bibitem[Peng et~al.(2022)Peng, Genova, Jiang, Tagliasacchi, Pollefeys,
  Funkhouser, et~al.]{peng2022openscene}
Songyou Peng, Kyle Genova, Chiyu Jiang, Andrea Tagliasacchi, Marc Pollefeys,
  Thomas Funkhouser, et~al.
\newblock Openscene: 3d scene understanding with open vocabularies.
\newblock \emph{arXiv preprint arXiv:2211.15654}, 2022.

\bibitem[Pinto et~al.(2016)Pinto, Gandhi, Han, Park, and
  Gupta]{pinto2016curious}
Lerrel Pinto, Dhiraj Gandhi, Yuanfeng Han, Yong-Lae Park, and Abhinav Gupta.
\newblock The curious robot: Learning visual representations via physical
  interactions.
\newblock In \emph{ECCV}, pages 3--18. Springer, 2016.

\bibitem[Radford et~al.(2021)Radford, Kim, Hallacy, Ramesh, Goh, Agarwal,
  Sastry, Askell, Mishkin, Clark, et~al.]{radford2021learning}
Alec Radford, Jong~Wook Kim, Chris Hallacy, Aditya Ramesh, Gabriel Goh,
  Sandhini Agarwal, Girish Sastry, Amanda Askell, Pamela Mishkin, Jack Clark,
  et~al.
\newblock Learning transferable visual models from natural language
  supervision.
\newblock In \emph{ICML}, 2021.

\bibitem[Rashid et~al.(2023)Rashid, Sharma, Kim, Kerr, Chen, Kanazawa, and
  Goldberg]{lerftogo2023}
Adam Rashid, Satvik Sharma, Chung~Min Kim, Justin Kerr, Lawrence~Yunliang Chen,
  Angjoo Kanazawa, and Ken Goldberg.
\newblock Language embedded radiance fields for zero-shot task-oriented
  grasping.
\newblock In \emph{CoRL}, 2023.

\bibitem[Shafiullah et~al.(2022)Shafiullah, Paxton, Pinto, Chintala, and
  Szlam]{shafiullah2022clip}
Nur Muhammad~Mahi Shafiullah, Chris Paxton, Lerrel Pinto, Soumith Chintala, and
  Arthur Szlam.
\newblock Clip-fields: Weakly supervised semantic fields for robotic memory.
\newblock \emph{arXiv preprint arXiv:2210.05663}, 2022.

\bibitem[Sharan et~al.(2009)Sharan, Rosenholtz, and
  Adelson]{sharan2009material}
Lavanya Sharan, Ruth Rosenholtz, and Edward Adelson.
\newblock Material perception: What can you see in a brief glance?
\newblock \emph{Journal of Vision}, 2009.

\bibitem[Sharan et~al.(2013)Sharan, Liu, Rosenholtz, and
  Adelson]{sharan2013recognizing}
Lavanya Sharan, Ce Liu, Ruth Rosenholtz, and Edward~H Adelson.
\newblock Recognizing materials using perceptually inspired features.
\newblock \emph{IJCV}, 2013.

\bibitem[Standley et~al.(2017)Standley, Sener, Chen, and
  Savarese]{pmlr-v78-standley17a}
Trevor Standley, Ozan Sener, Dawn Chen, and Silvio Savarese.
\newblock image2mass: Estimating the mass of an object from its image.
\newblock In \emph{CoRL}, 2017.

\bibitem[Tancik et~al.(2023)Tancik, Weber, Ng, Li, Yi, Wang, Kristoffersen,
  Austin, Salahi, Ahuja, et~al.]{tancik2023nerfstudio}
Matthew Tancik, Ethan Weber, Evonne Ng, Ruilong Li, Brent Yi, Terrance Wang,
  Alexander Kristoffersen, Jake Austin, Kamyar Salahi, Abhik Ahuja, et~al.
\newblock Nerfstudio: A modular framework for neural radiance field
  development.
\newblock In \emph{SIGGRAPH}, 2023.

\bibitem[Varma and Zisserman(2008)]{varma2008statistical}
Manik Varma and Andrew Zisserman.
\newblock A statistical approach to material classification using image patch
  exemplars.
\newblock \emph{TPAMI}, 2008.

\bibitem[Wu et~al.(2015)Wu, Yildirim, Lim, Freeman, and
  Tenenbaum]{NIPS2015_d09bf415}
Jiajun Wu, Ilker Yildirim, Joseph~J Lim, Bill Freeman, and Josh Tenenbaum.
\newblock Galileo: Perceiving physical object properties by integrating a
  physics engine with deep learning.
\newblock In \emph{NeurIPS}, 2015.

\bibitem[Wu et~al.(2016)Wu, Lim, Zhang, Tenenbaum, and Freeman]{wu2016physics}
Jiajun Wu, Joseph~J Lim, Hongyi Zhang, Joshua~B Tenenbaum, and William~T
  Freeman.
\newblock Physics 101: Learning physical object properties from unlabeled
  videos.
\newblock In \emph{BMVC}, 2016.

\bibitem[Wu et~al.(2022)Wu, Chen, Liu, Ren, and Wang]{wu2022casa}
Yuefan Wu, Zeyuan Chen, Shaowei Liu, Zhongzheng Ren, and Shenlong Wang.
\newblock Casa: Category-agnostic skeletal animal reconstruction.
\newblock \emph{NeurIPS}, 2022.

\bibitem[Yang et~al.(2023)Yang, Li, Lin, Wang, Lin, Liu, and
  Wang]{yang2023dawn}
Zhengyuan Yang, Linjie Li, Kevin Lin, Jianfeng Wang, Chung-Ching Lin, Zicheng
  Liu, and Lijuan Wang.
\newblock The dawn of lmms: Preliminary explorations with gpt-4v (ision).
\newblock \emph{arXiv preprint arXiv:2309.17421}, 9, 2023.

\bibitem[Yao and Hauser(2023)]{yaoestimating}
Shaoxiong Yao and Kris Hauser.
\newblock Estimating tactile models of heterogeneous deformable objects in real
  time.
\newblock In \emph{ICRA}, 2023.

\bibitem[Ye et~al.(2023)Ye, Chen, Xu, Zu, Shao, Liu, Cui, Zhou, Gong, Shen,
  et~al.]{ye2023comprehensive}
Junjie Ye, Xuanting Chen, Nuo Xu, Can Zu, Zekai Shao, Shichun Liu, Yuhan Cui,
  Zeyang Zhou, Chao Gong, Yang Shen, et~al.
\newblock A comprehensive capability analysis of gpt-3 and gpt-3.5 series
  models.
\newblock \emph{arXiv preprint arXiv:2303.10420}, 2023.

\bibitem[Yildirim et~al.()Yildirim, Wu, Du, and
  Tenenbaum]{yildiriminterpreting}
Ilker Yildirim, Jiajun Wu, Yilun Du, and Joshua~B Tenenbaum.
\newblock Interpreting dynamic scenes by a physics engine and bottom-up visual
  cues.

\end{thebibliography}
}

\clearpage
\setcounter{section}{0}
\renewcommand*{\thesection}{\Alph{section}}
\maketitlesupplementary
\begin{abstract}
In the following supplementary material, we provide additional experimental details (Sec.~\ref{sec:additional_details}) and additional results (Sec.~\ref{sec:additional_results}). We invite readers to watch the supplementary video (\texttt{supp.mp4}) for further visualization.
\end{abstract}

\section{Additional Experimental Details}
\label{sec:additional_details}

\subsection{Point Cloud Extraction Details}
We use the default point cloud exporter in Nerfstudio~\cite{tancik2023nerfstudio}, which allows one to define a bounding box for the points to keep. For ABO-500, we use a bounding box of size $1 \times 1 \times 1$ centered at $(0, 0, 0)$. For our in-house dataset, we use a bounding box of size $1.5 \times 1.5 \times 1.5$ centered at $(0, 0, -0.75)$. Points are filtered out if the average distance to their 20 nearest neighbors is over 10 standard deviations away from the mean.

\subsection{Prompting Details}
We provide the prompts used for captioning~(Fig.~\ref{fig:captioning_prompt}) and material proposal with mass density~(Fig.~\ref{fig:density_prompt}), friction coefficient~(Fig.~\ref{fig:friction_prompt}), and hardness~(Fig.~\ref{fig:hardness_prompt}) values. The prompt used for thickness estimation is provided in Fig.~\ref{fig:thickness_prompt}. We also provide the prompts used for LLaVA mass estimation~(Fig.~\ref{fig:llava_mass_prompt}), GPT-4V friction estimation~(Fig.~\ref{fig:gpt4v_friction_prompt}), and GPT-4V hardness estimation~(Fig.~\ref{fig:gpt4v_hardness_prompt}).

\begin{figure}[h]
\begin{tcolorbox}[colback=white]
\footnotesize
Question: Give a detailed description of the object. Answer:
\end{tcolorbox}

\caption{Prompt used for captioning with BLIP-2.}
\label{fig:captioning_prompt}
\end{figure}

\begin{figure}[h]
\begin{tcolorbox}[colback=white]
\footnotesize
\textbf{System:} You will be provided with captions that each describe an image of an object. The captions will be delimited with quotes ("). Based on the caption, give me 5 materials that the object might be made of, along with the mass densities (in kg/m$^3$) of each of those materials. You may provide a range of values for the mass density instead of a single value. Try to consider all the possible parts of the object. Do not include coatings like "paint" in your answer.\\

Format Requirement:\\
You must provide your answer as a list of 5 (material: mass density) pairs, each separated by a semi-colon (;). Do not include any other text in your answer, as it will be parsed by a code script later. Your answer must look like:\\
(material 1: low-high kg/m$^3$);(material 2: low-high kg/m$^3$);(material 3: low-high kg/m$^3$);(material 4: low-high kg/m$^3$);(material 5: low-high kg/m$^3$)
\end{tcolorbox}

\caption{Prompt used for proposing materials and providing their mass density values.}
\label{fig:density_prompt}
\end{figure}

\begin{figure}[h]
\begin{tcolorbox}[colback=white]
\footnotesize
\textbf{System:} You will be provided with captions that each describe an image. The captions will be delimited with quotes ("). Based on the caption, give me 3 materials that the surfaces in the image might be made of, along with the kinetic friction coefficient of each material when sliding against a fabric surface. You may provide a range of values for the friction coefficient instead of a single value. Try to consider all the possible surfaces.\\

Format Requirement:\\
You must provide your answer as a list of 3 (material: friction coefficient) pairs, each separated by a semi-colon (;). Do not include any other text in your answer, as it will be parsed by a code script later. Your answer must look like:\\
(material 1: low-high);(material 2: low-high);(material 3: low-high)\\
Try to provide as narrow of a range as possible for the friction coefficient.
\end{tcolorbox}

\caption{Prompt used for proposing materials and providing their friction coefficients.}
\label{fig:friction_prompt}
\end{figure}

\begin{figure}[h]
\begin{tcolorbox}[colback=white]
\footnotesize
\textbf{System:} You will be provided with captions that each describe an image of an object. The captions will be delimited with quotes ("). Based on the caption, give me 3 materials that the object might be made of, along with the hardness of each of those materials. Choose whether to use Shore A hardness or Shore D hardness depending on the material. You may provide a range of values for hardness instead of a single value. Try to consider all the possible parts of the object.\\

Format Requirement:\\
You must provide your answer as a list of 3 (material: hardness, Shore A/D) tuples, each separated by a semi-colon (;). Do not include any other text in your answer, as it will be parsed by a code script later. Your answer must look like:\\
(material 1: low-high, <Shore A or Shore D>);(material 2: low-high, <Shore A or Shore D>);(material 3: low-high, <Shore A or Shore D>)\\
Make sure to use Shore A or Shore D hardness, not Mohs hardness.
\end{tcolorbox}

\caption{Prompt used for proposing materials and providing their Shore hardness values.}
\label{fig:hardness_prompt}
\end{figure}

\begin{figure}[h]
\begin{tcolorbox}[colback=white]
\footnotesize
\textbf{System:} You will be provided with captions that each describe an image of an object, along with a set of possible materials used to make the object. For each material, estimate the thickness (in cm) of that material in the object. You may provide a range of values for the thickness instead of a single value.\\

Format Requirement:\\
You must provide your answer as a list of 5 (material: thickness) pairs, each separated by a semi-colon (;). Do not include any other text in your answer, as it will be parsed by a code script later. Your answer must look like:\\
(material 1: low-high cm);(material 2: low-high cm);(material 3: low-high cm);(material 4: low-high cm);(material 5: low-high cm)\\

\textbf{User:} Caption: "a lamp with a white shade" Materials: "fabric, plastic, metal, ceramic, glass"\\
\textbf{Assistant:} (fabric: 0.1-0.2 cm);(plastic: 0.3-1.0 cm);(metal: 0.1-0.2 cm);(ceramic: 0.2-0.5 cm);(glass: 0.3-0.8 cm)\\
\textbf{User:} Caption: "a grey ottoman" Materials: "wood, fabric, foam, metal, plastic"\\
\textbf{Assistant:} (wood: 2.0-4.0 cm);(fabric: 0.2-0.5 cm);(foam: 5.0-15.0 cm);(metal: 0.1-0.2 cm);(plastic: 0.5-1.0 cm)\\
\textbf{User:} Caption: "a white frame" Materials: "plastic, wood, aluminum, steel, glass"\\
\textbf{Assistant:} (plastic: 0.1-0.3 cm);(wood: 1.0-1.5 cm);(aluminum: 0.1-0.3 cm);(steel: 0.1-0.2 cm);(glass: 0.2-0.5 cm)\\
\textbf{User:} Caption: "a metal rack with three shelves" Materials: "steel, aluminum, wood, plastic, iron"\\
\textbf{Assistant:} (steel: 0.1-0.2 cm);(aluminum: 0.1-0.3 cm);(wood: 1.0-2.0 cm);(plastic: 0.5-1.0 cm);(iron: 0.5-1.0 cm)\\
\end{tcolorbox}

\caption{Prompt used for estimating the thickness of proposed materials. Since this is a somewhat confusing task, we provide a few in-context examples to help the LLM understand what we mean by thickness.}
\label{fig:thickness_prompt}
\end{figure}

\begin{figure}[h]
\begin{tcolorbox}[colback=white]
\footnotesize
\textbf{System:} Estimate the mass of the object in kilograms. Provide your answer as only a decimal number.
\end{tcolorbox}

\caption{Prompt used for estimating mass with LLaVA.}
\label{fig:llava_mass_prompt}
\end{figure}

\begin{figure}[h]
\begin{tcolorbox}[colback=white]
\footnotesize
\textbf{System:} You will be given an image, followed by a mask specifying a point on the image. Estimate the coefficient of kinetic friction between a fabric surface and the surface at that point in the image.\\

Format Requirement:\\
You must provide either a single number or a range (e.g. "0.6-0.8") as your answer. Give your best guess. Do not include any other text in your answer, as it will be parsed by a code script later.
\end{tcolorbox}

\caption{Prompt used for estimating friction with GPT-4V.}
\label{fig:gpt4v_friction_prompt}
\end{figure}

\begin{figure}[h]
\begin{tcolorbox}[colback=white]
\footnotesize
\textbf{System:} You will be given an image, followed by a mask specifying a point on the image. Estimate the hardness of the object in the image at the given point. Choose whether to use Shore A hardness or Shore D hardness depending on the material.\\

Format Requirement:\\
You must provide a pair of either a single number or a range (e.g. "0.6-0.8") and whether it is in Shore A or Shore D as your answer. Give your best estimate, and make sure to use Shore A or Shore D hardness, not Mohs hardness. Do not include any other text in your answer. Your answer must look like:\\
(number, <Shore A or Shore D>)
\end{tcolorbox}

\caption{Prompt used for estimating hardness with GPT-4V.}
\label{fig:gpt4v_hardness_prompt}
\end{figure}

\subsection{CNN Baseline Details}
Our CNN baseline takes one RGB image and outputs the predicted mass. We use an ImageNet-pretrained ResNet50 backbone with the last classification layer removed. A set of fully connected layers with ReLU activations follows the ResNet50. Within the fully connected layers, feature dimensions are as follows: 2048, 32, 16, and 1. We attach a LogSigmoid layer to ensure the output is non-negative. Our input image is normalized based on ImageNet statistics before feeding into our model. We train for 20 epochs using an Adam optimizer with a learning rate of $\alpha = 0.001$.

\section{Additional Results}
\label{sec:additional_results}

\subsection{Young's Modulus and Thermal Conductivity}
Our method can be used to predict physical properties in an open-vocabulary manner. We show that it can be used to predict Young's modulus and thermal conductivity on objects from ABO-500 (val) in Fig.~\ref{fig:supp_phys_props}. For visualization purposes, we force the candidate materials to be the same as those proposed for mass density estimation. The exact prompts used for these results can be found in Fig.~\ref{fig:ym_prompt} (Young's modulus) and Fig.~\ref{fig:tc_prompt} (thermal conductivity). We use GPT-3.5 Turbo and a kernel regression temperature of $T=0.01$. 

\begin{table*}[!t]
    \centering
    \resizebox{\linewidth}{!}{
\setlength{\tabcolsep}{0.2em} %
\renewcommand{\arraystretch}{1.}
    \begin{tabular}{cccccc}
    
    Input RGB &
    \multicolumn{2}{c}{Material Segmentation} &
    Mass Density (kg/m$^3$) &
    Young's Mod. ($\log$ GPa) &
    Thermal Cond. ($\log$ W/mK) 
    \\

    \tikz{
        \node[draw=white, line width=0mm, inner sep=0pt] 
        {\includegraphics[width=.17\linewidth, trim={4cm 5cm 4cm 3cm}, clip]{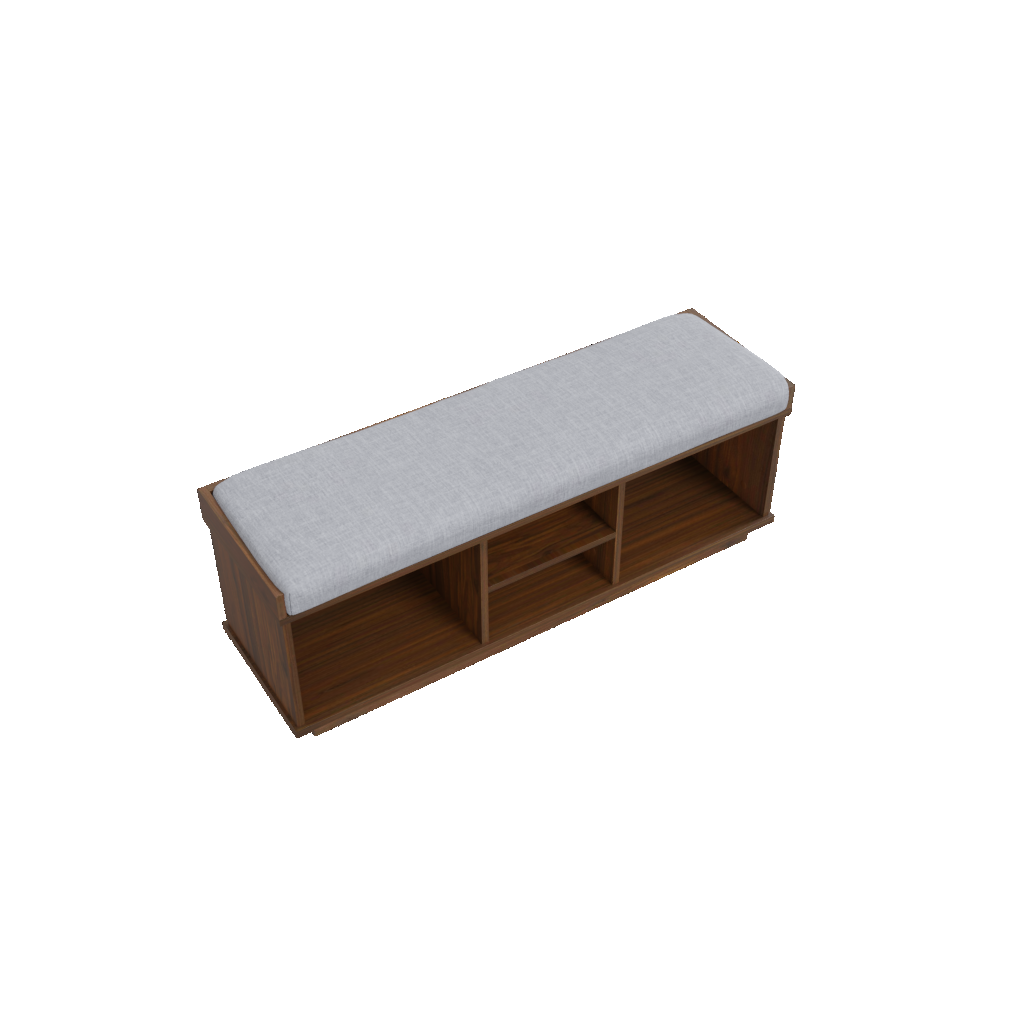}}
        }&
        \tikz{
        \node[draw=white, line width=0mm, inner sep=0pt] 
        {\includegraphics[width=.17\linewidth, trim={4cm 5cm 4cm 3cm}, clip]{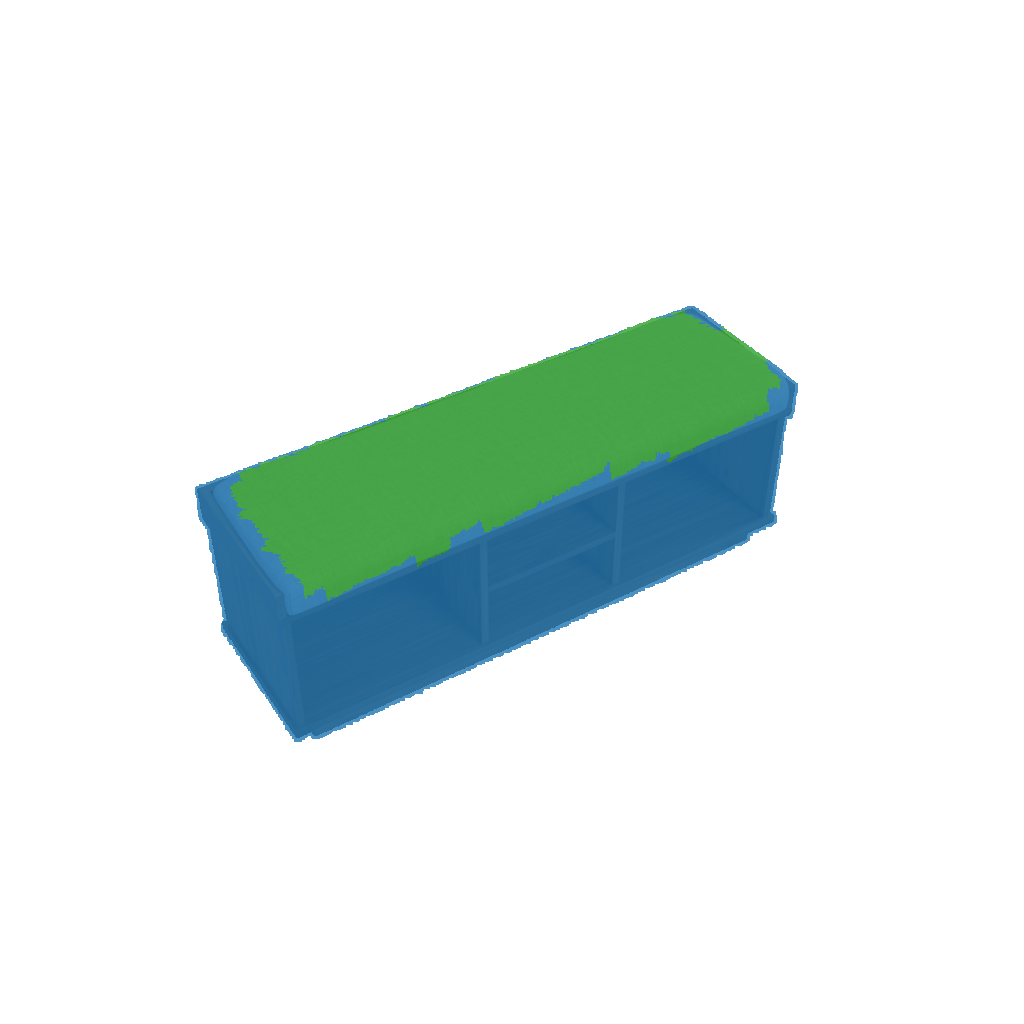} }
        }&
         \tikz{
        \node[draw=white, line width=0mm, inner sep=0pt] 
        {\includegraphics[width=.14\linewidth, clip]{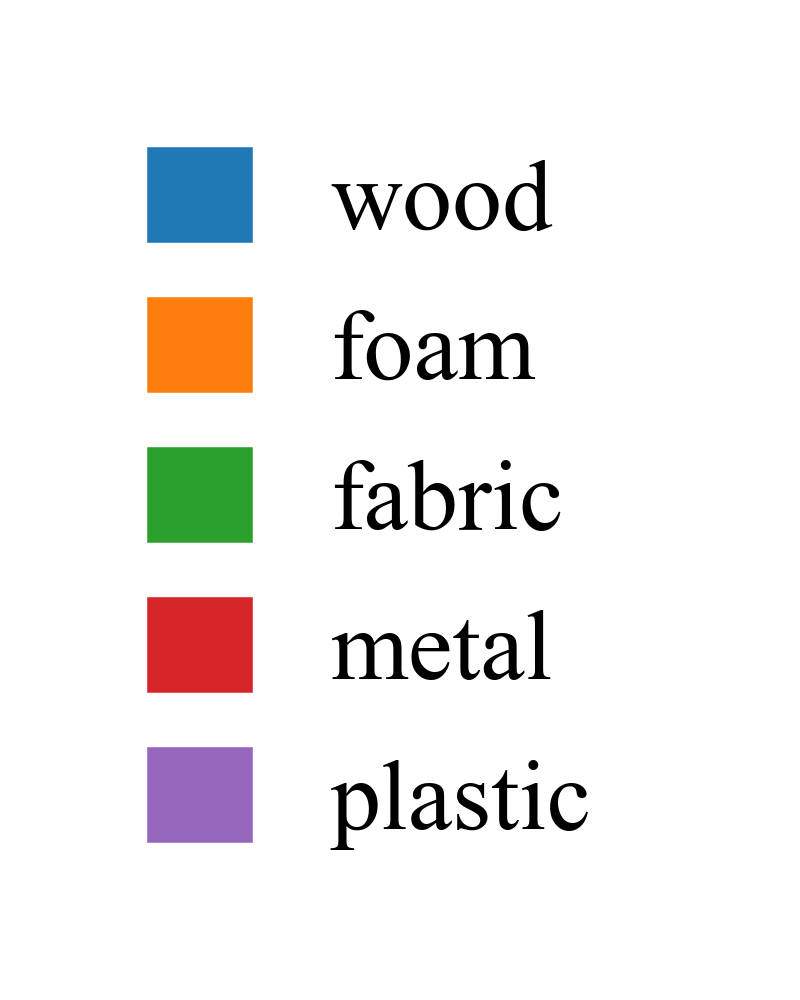}}
        }&
         \tikz{
        \node[draw=white, line width=0mm, inner sep=0pt] 
        {\includegraphics[width=.17\linewidth, trim={4cm 5cm 4cm 3cm}, clip]{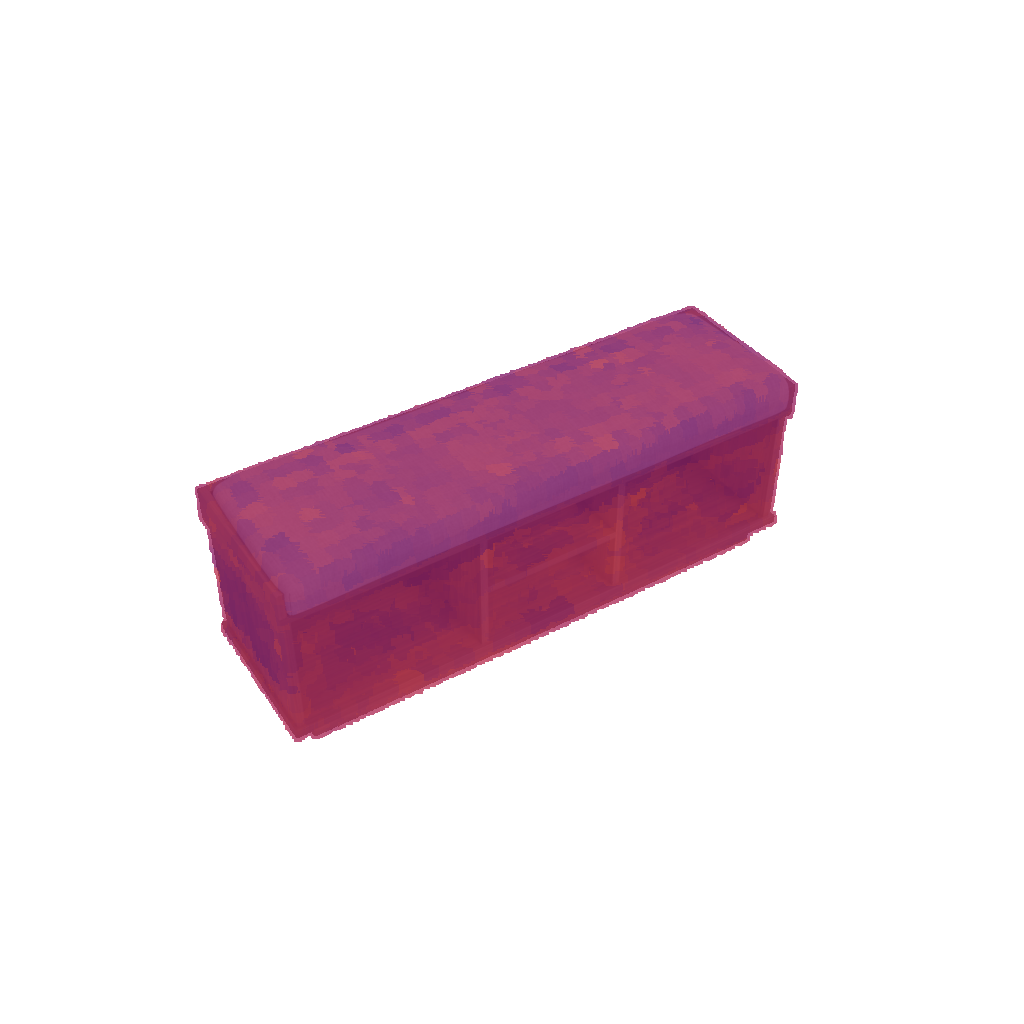}}
        }&
        \tikz{
        \node[draw=white, line width=0mm, inner sep=0pt] 
        { \includegraphics[width=.17\linewidth, trim={4cm 5cm 4cm 3cm}, clip]{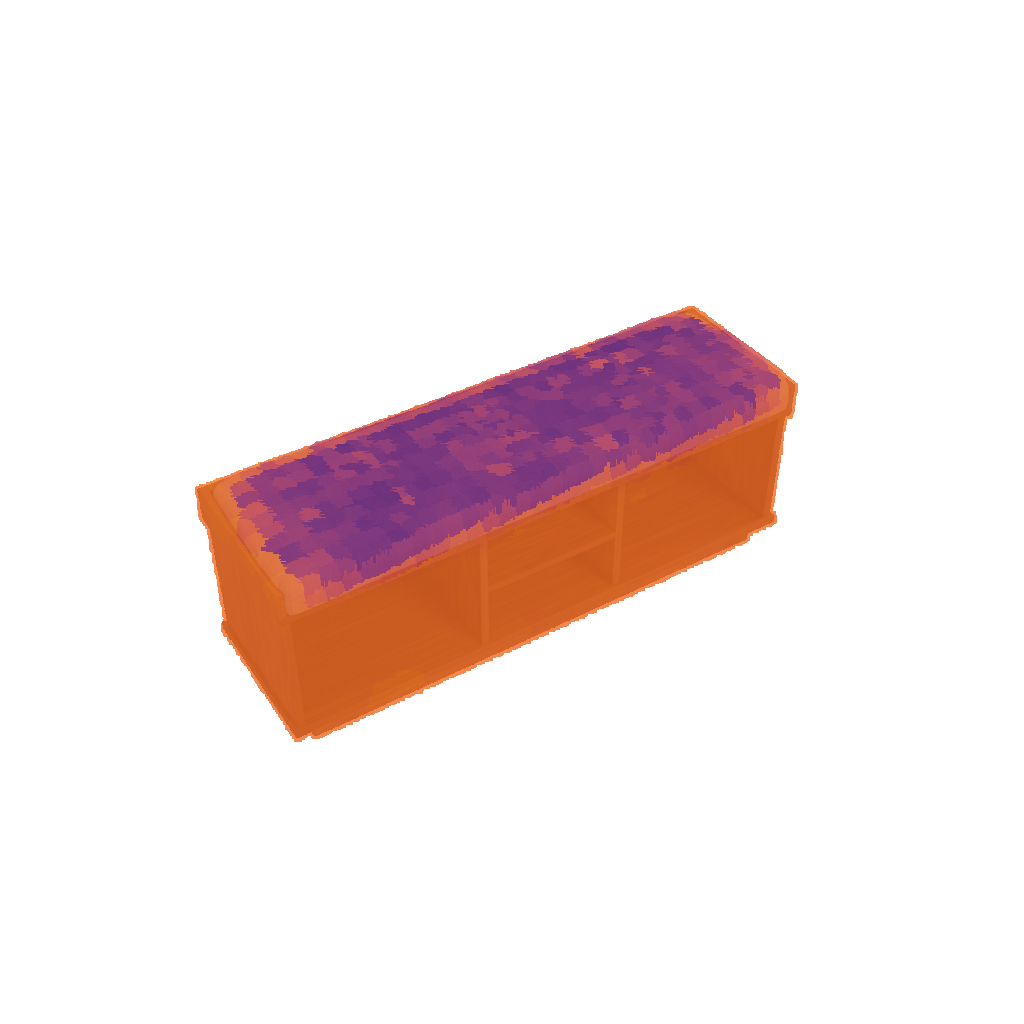}}
        }&
        \tikz{
        \node[draw=white, line width=0mm, inner sep=0pt] 
        { \includegraphics[width=.17\linewidth, trim={4cm 5cm 4cm 3cm}, clip]{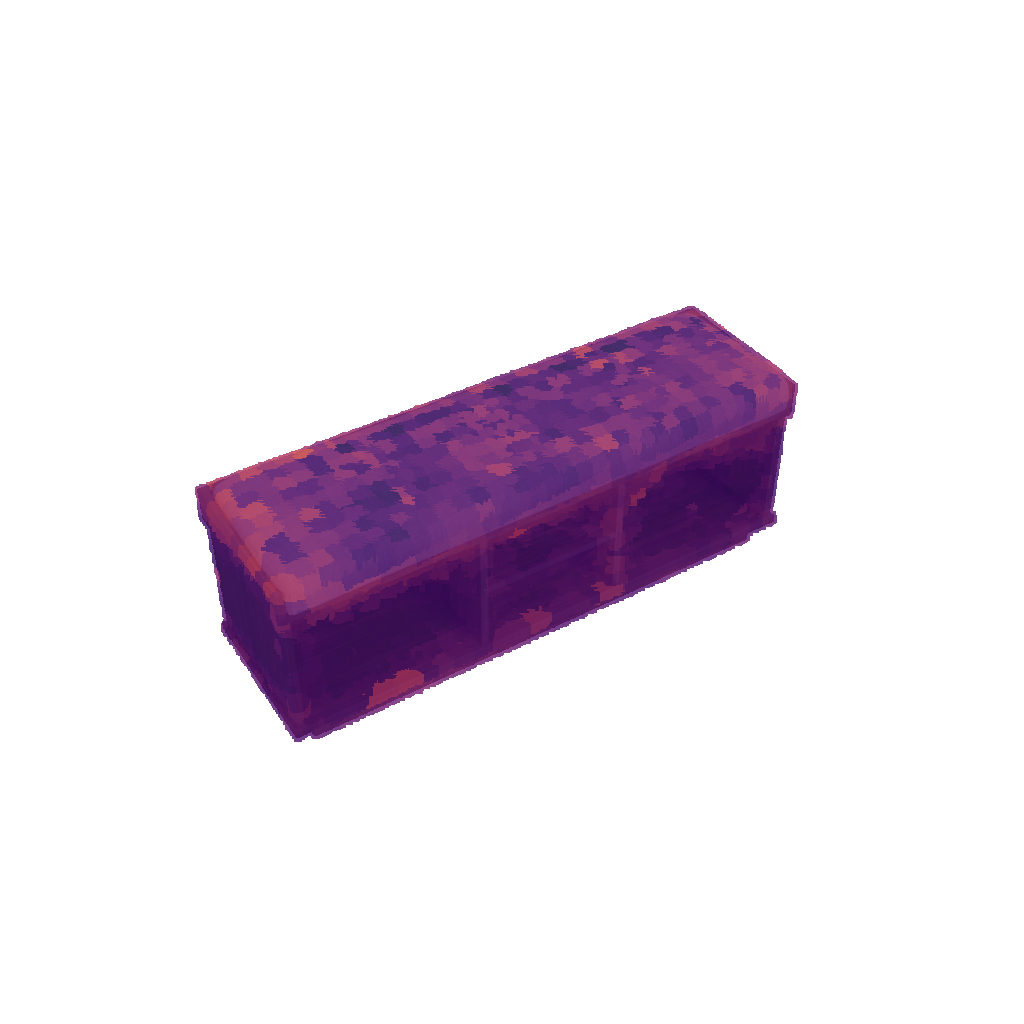}}
        }
        \\
        
         \tikz{
        \node[draw=white, line width=0mm, inner sep=0pt] 
        {\includegraphics[width=.17\linewidth, trim={4cm 5cm 4cm 3cm}, clip]{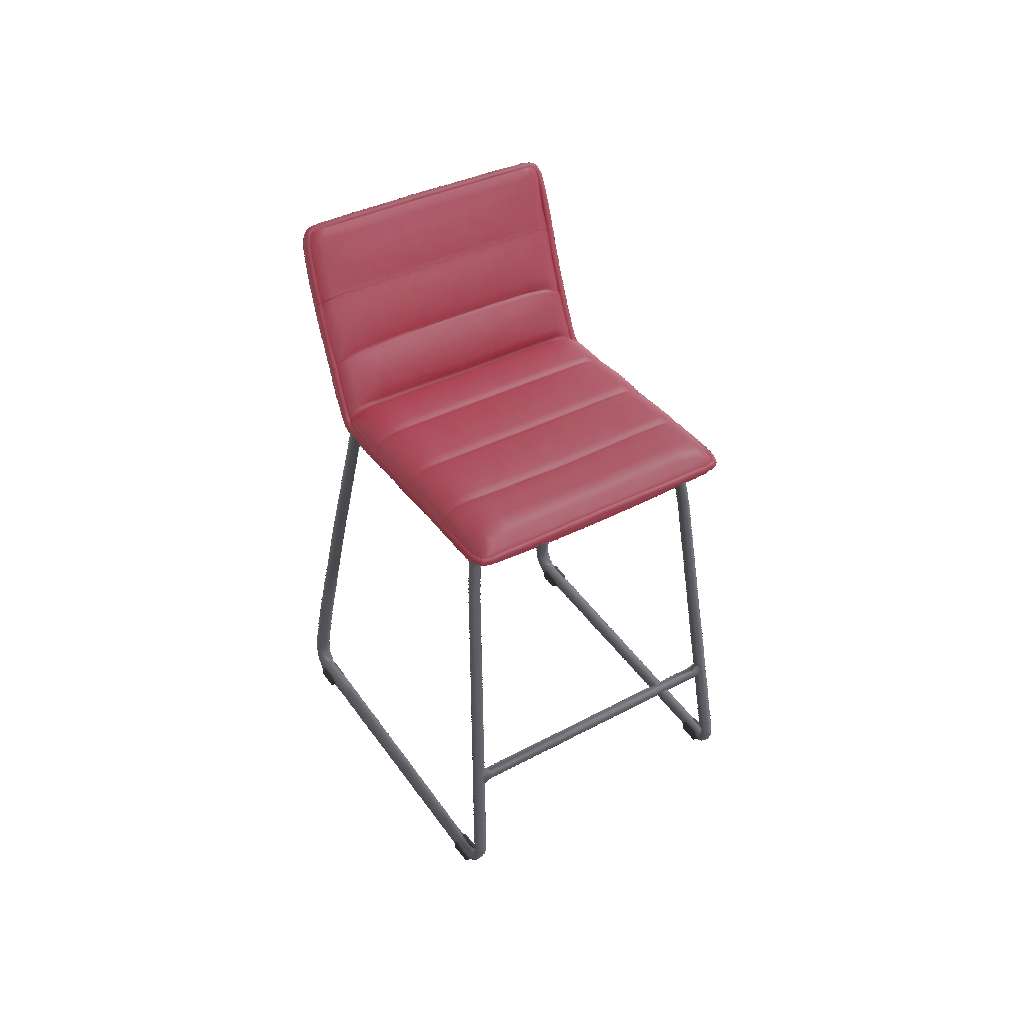}}
        }&
        \tikz{
        \node[draw=white, line width=0mm, inner sep=0pt] 
        {\includegraphics[width=.17\linewidth, trim={4cm 5cm 4cm 3cm}, clip]{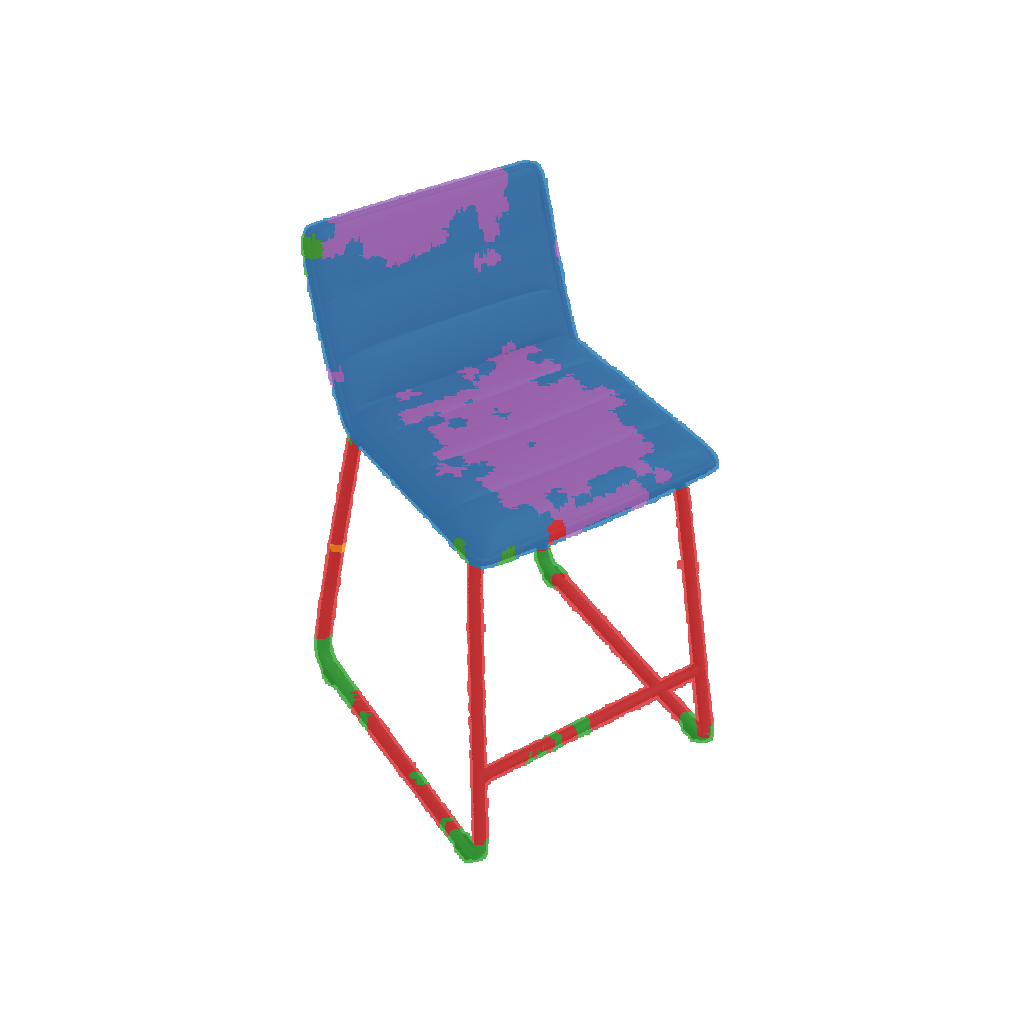} }
        }&
         \tikz{
        \node[draw=white, line width=0mm, inner sep=0pt] 
        {\includegraphics[width=.14\linewidth, clip]{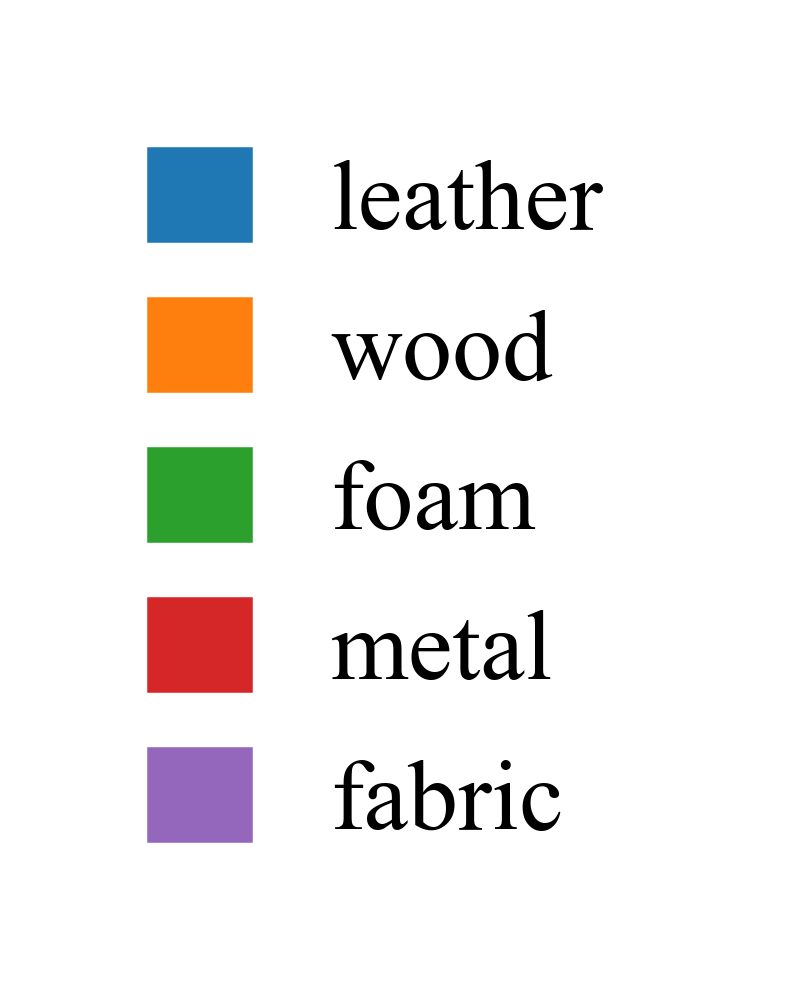}}
        }&
         \tikz{
        \node[draw=white, line width=0mm, inner sep=0pt] 
        {\includegraphics[width=.17\linewidth, trim={4cm 5cm 4cm 3cm}, clip]{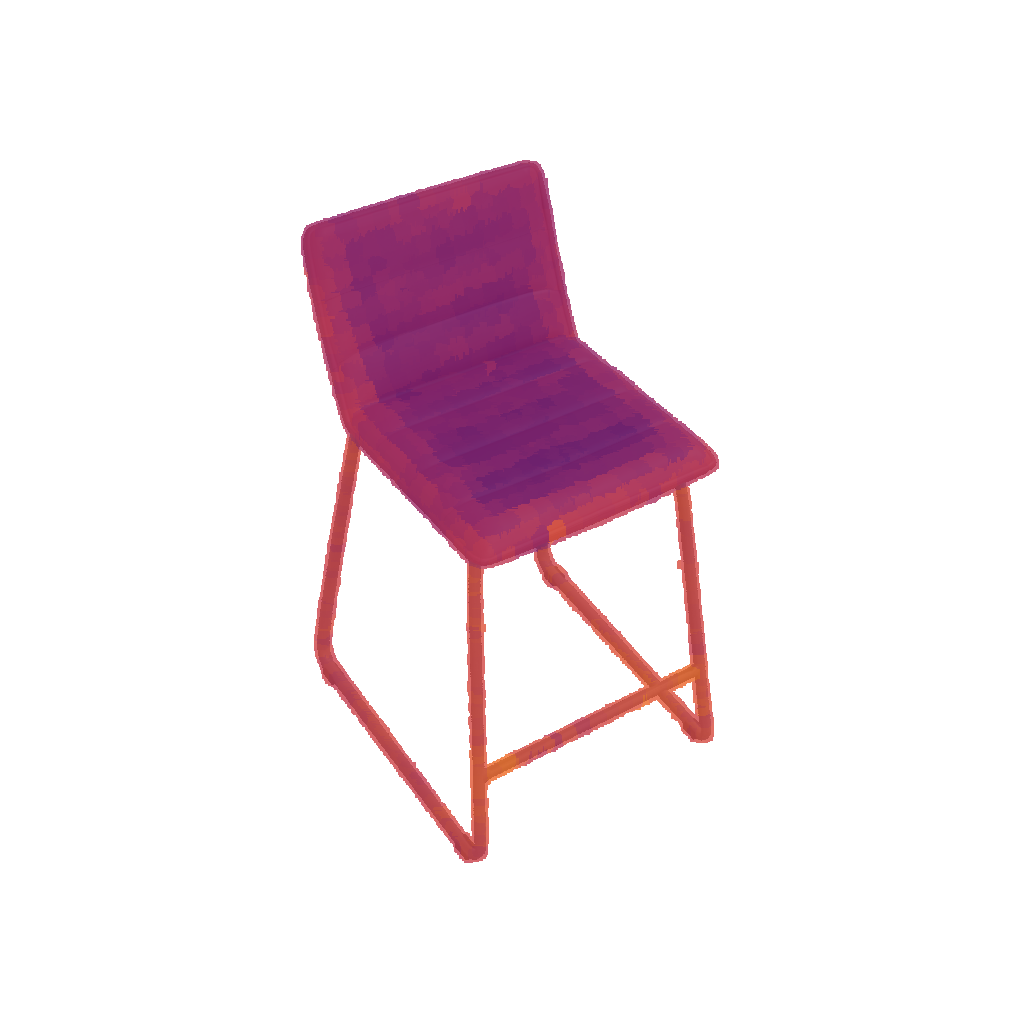}}
        }&
        \tikz{
        \node[draw=white, line width=0mm, inner sep=0pt] 
        { \includegraphics[width=.17\linewidth, trim={4cm 5cm 4cm 3cm}, clip]{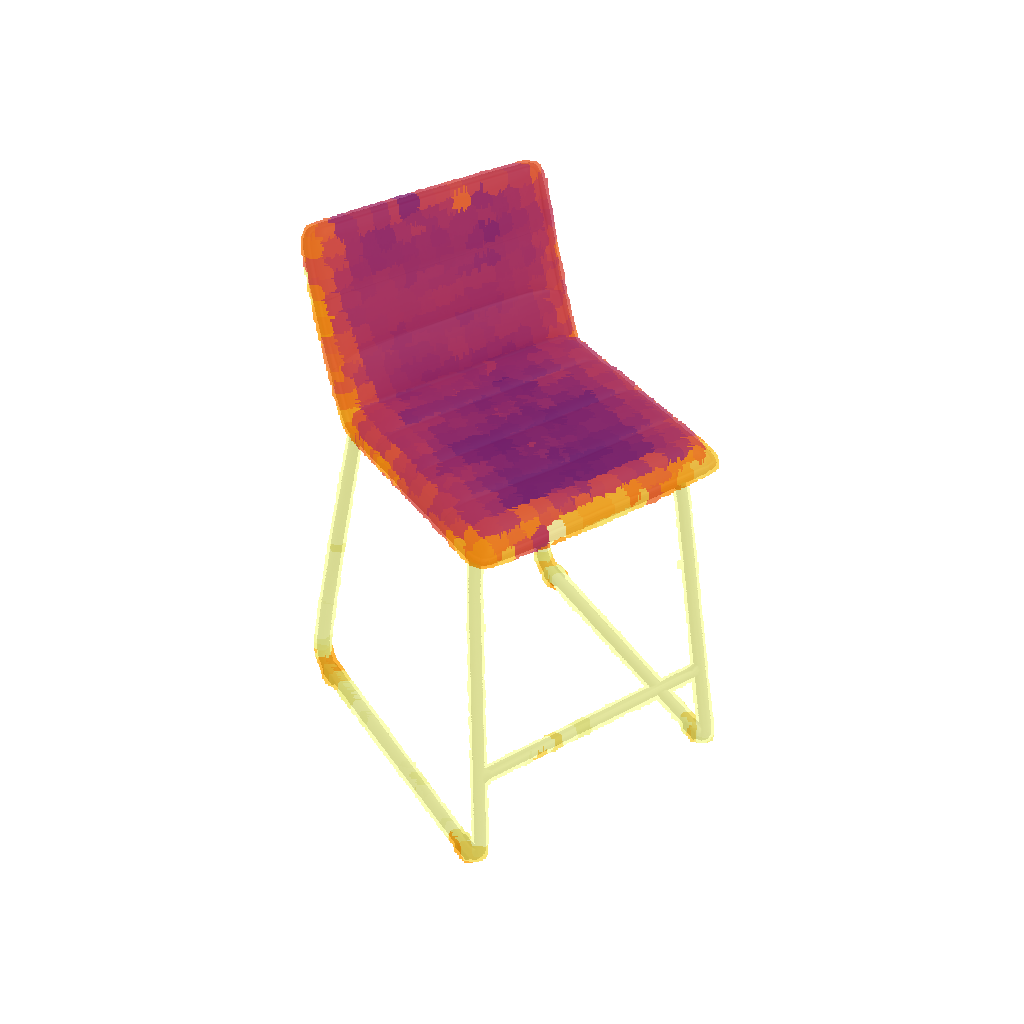}}
        }&
        \tikz{
        \node[draw=white, line width=0mm, inner sep=0pt] 
        { \includegraphics[width=.17\linewidth, trim={4cm 5cm 4cm 3cm}, clip]{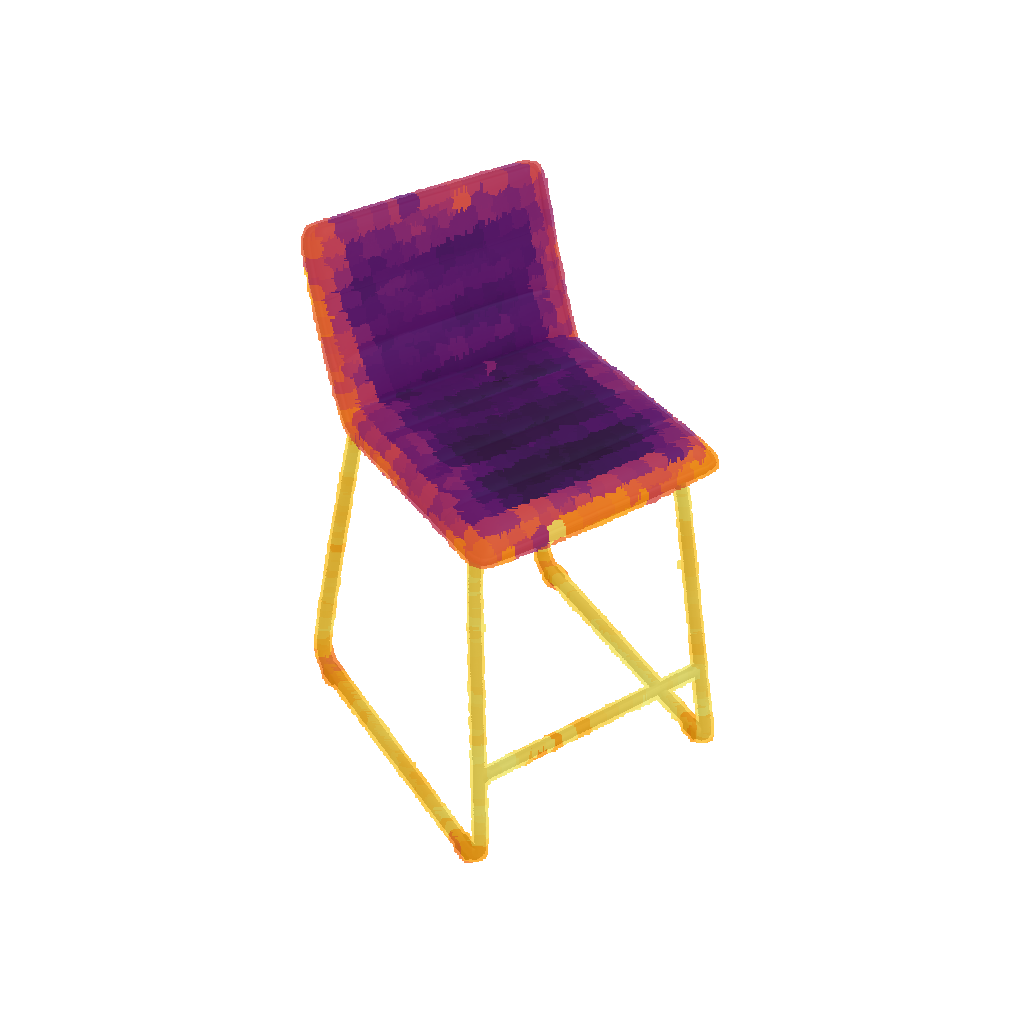}}
        }
        \\
        
         \tikz{
        \node[draw=white, line width=0mm, inner sep=0pt] 
        {\includegraphics[width=.17\linewidth, trim={4cm 5cm 4cm 3cm}, clip]{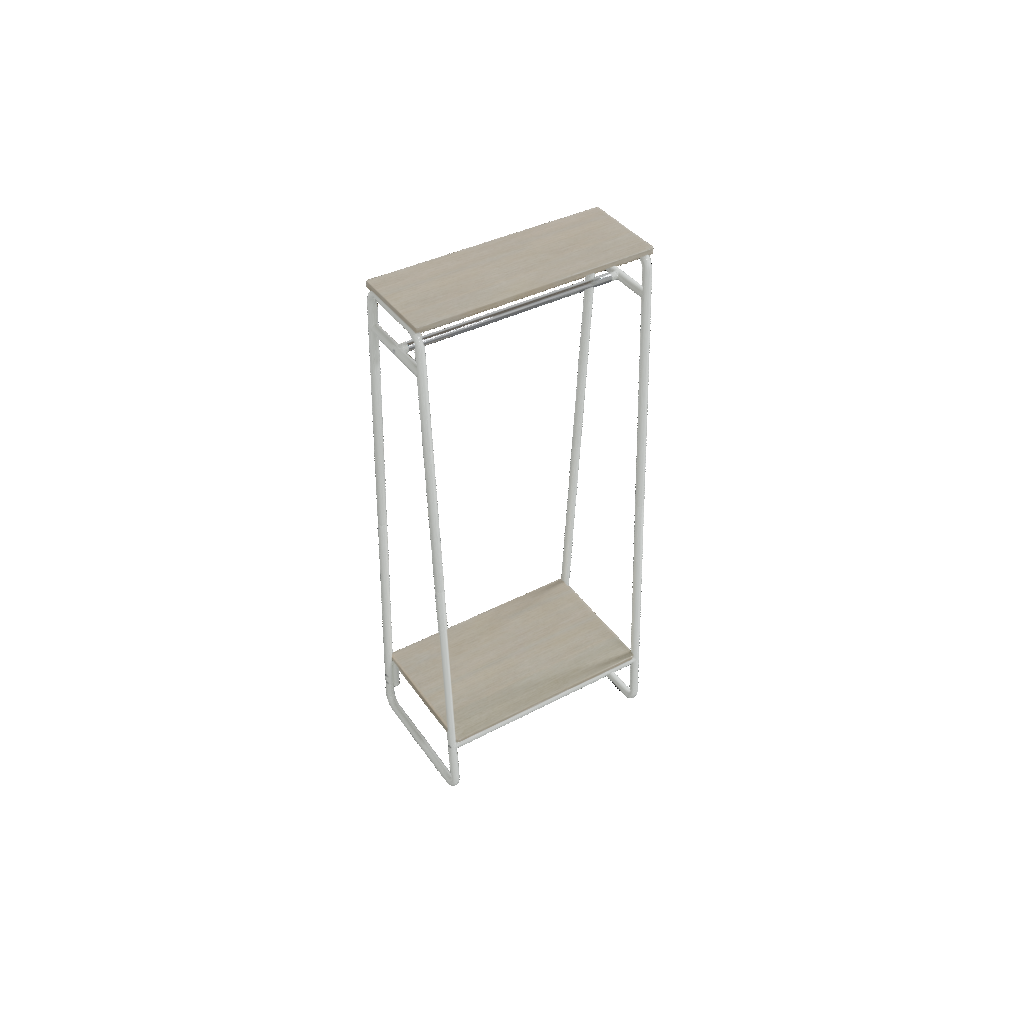}}
        }&
        \tikz{
        \node[draw=white, line width=0mm, inner sep=0pt] 
        {\includegraphics[width=.17\linewidth, trim={4cm 5cm 4cm 3cm}, clip]{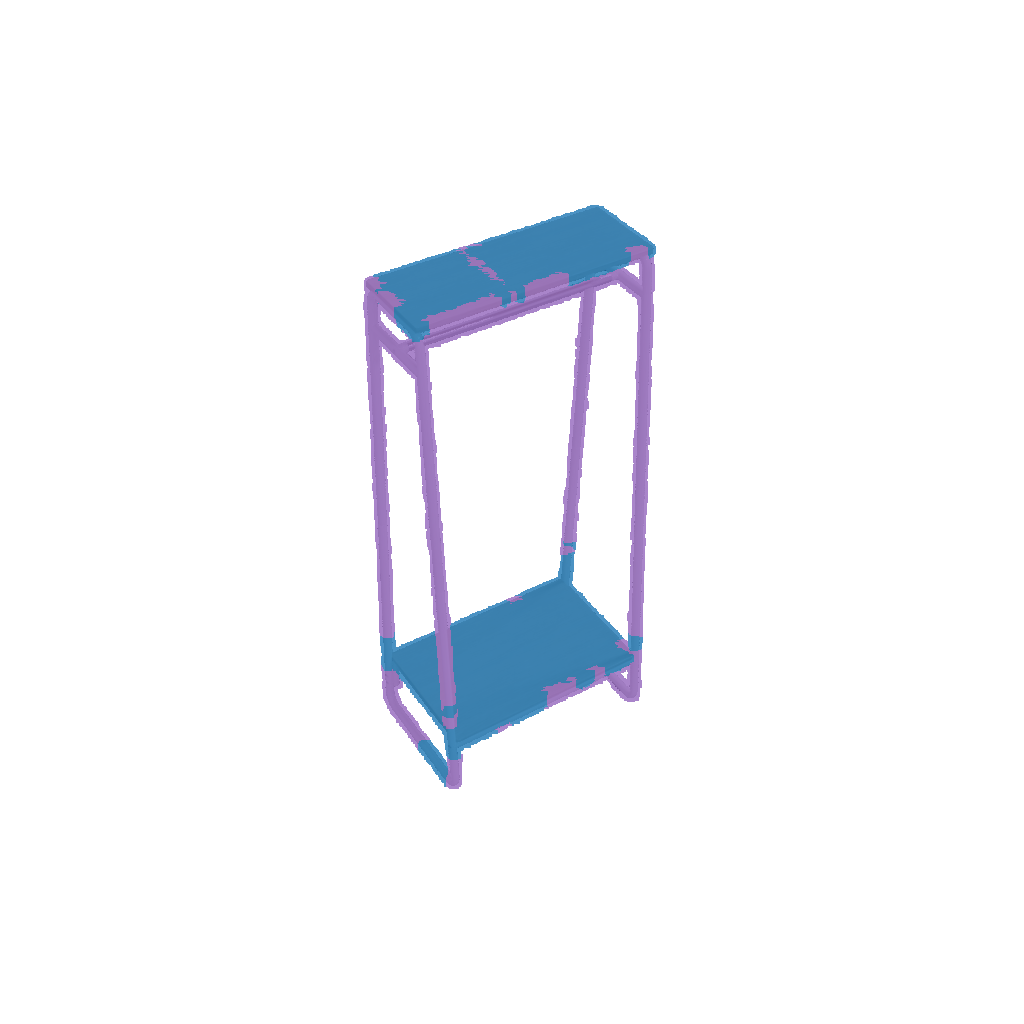} }
        }&
         \tikz{
        \node[draw=white, line width=0mm, inner sep=0pt] 
        {\includegraphics[width=.14\linewidth, clip]{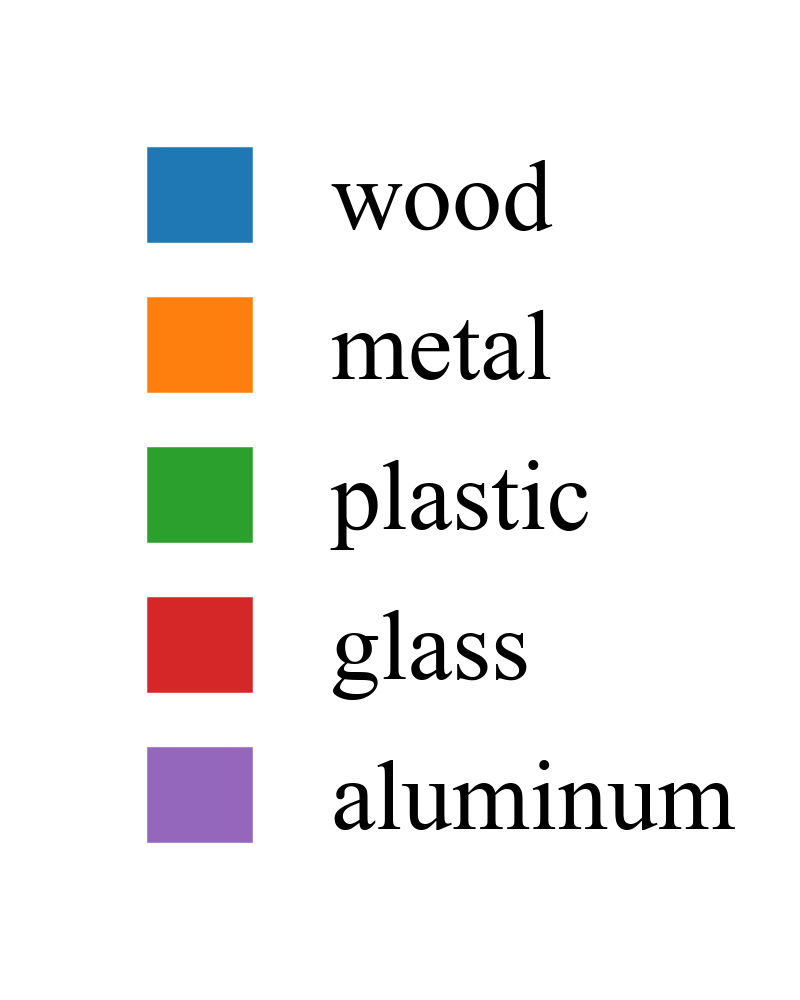}}
        }&
         \tikz{
        \node[draw=white, line width=0mm, inner sep=0pt] 
        {\includegraphics[width=.17\linewidth, trim={4cm 5cm 4cm 3cm}, clip]{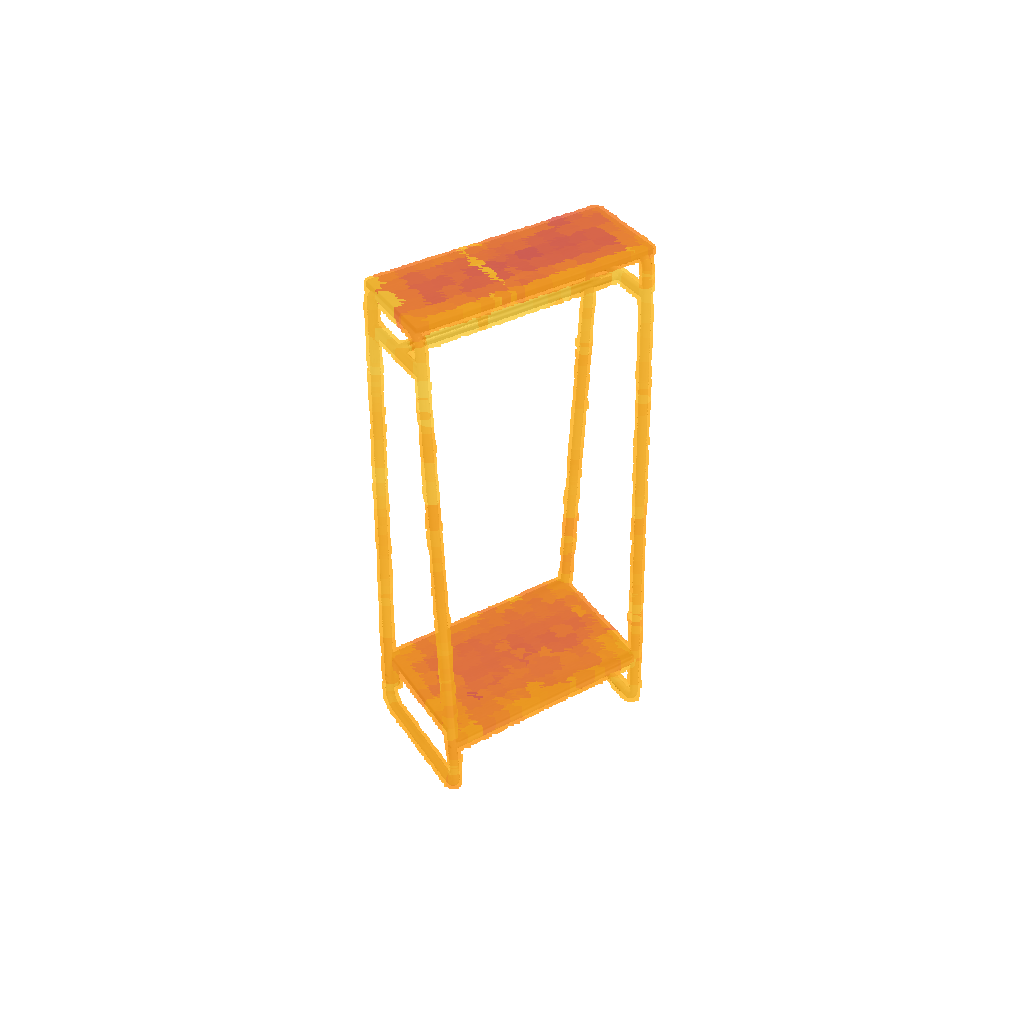}}
        }&
        \tikz{
        \node[draw=white, line width=0mm, inner sep=0pt] 
        { \includegraphics[width=.17\linewidth, trim={4cm 5cm 4cm 3cm}, clip]{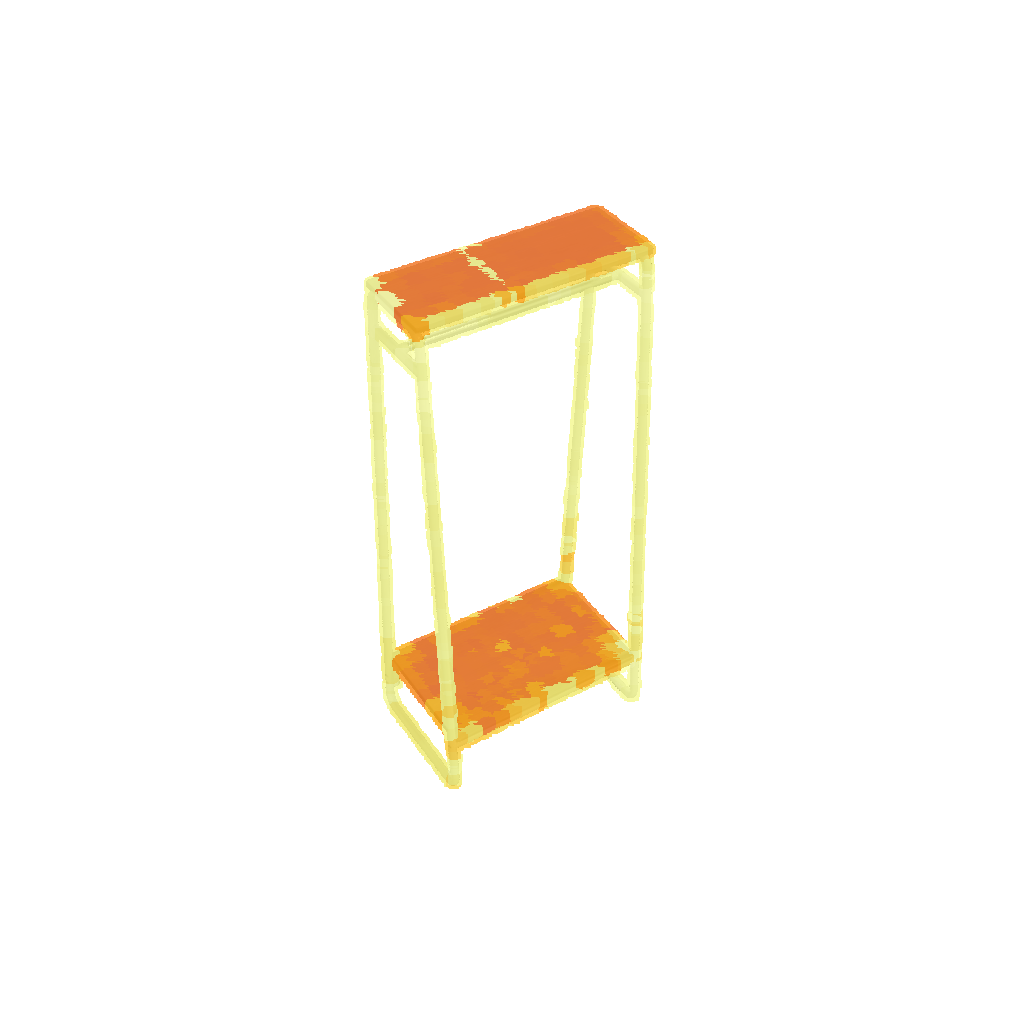}}
        }&
        \tikz{
        \node[draw=white, line width=0mm, inner sep=0pt] 
        { \includegraphics[width=.17\linewidth, trim={4cm 5cm 4cm 3cm}, clip]{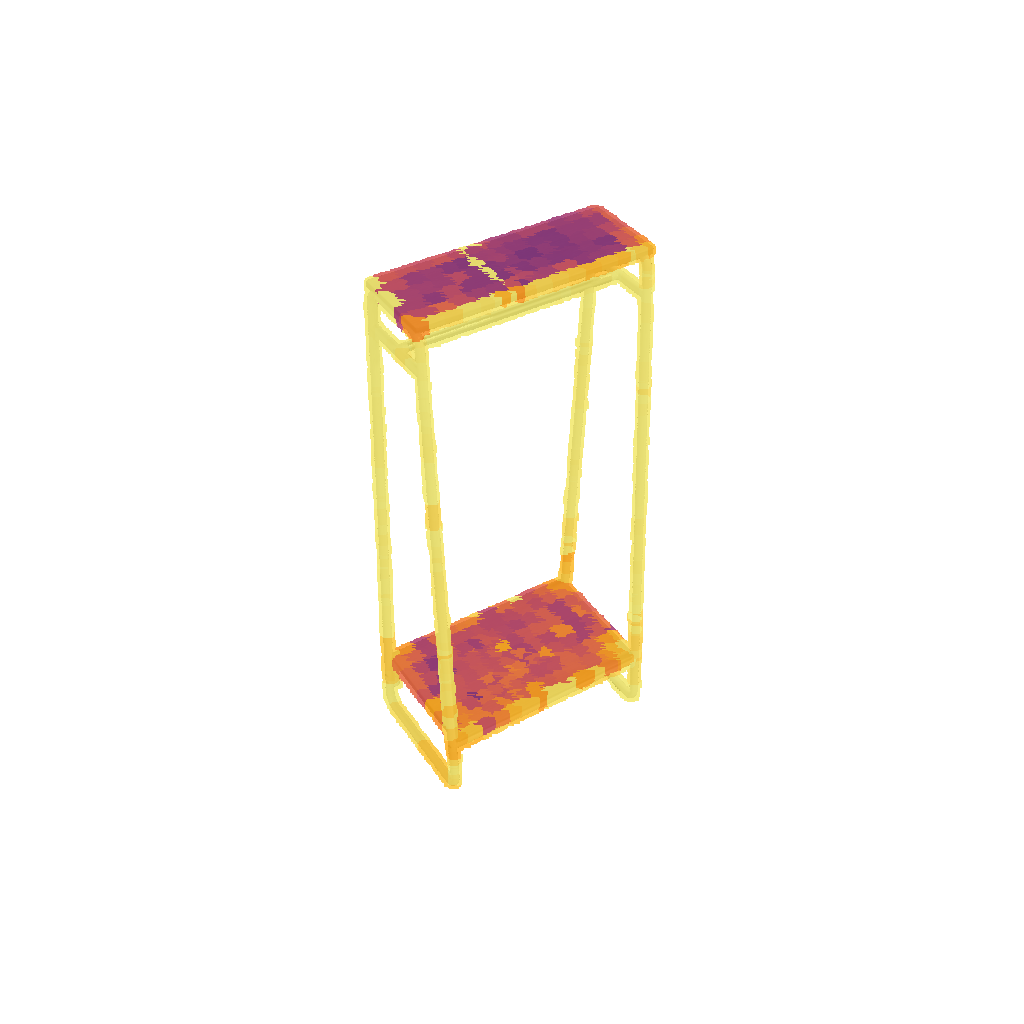}}
        }
        \\

         \tikz{
        \node[draw=white, line width=0mm, inner sep=0pt] 
        {\includegraphics[width=.17\linewidth, trim={4cm 4cm 4cm 4cm}, clip]{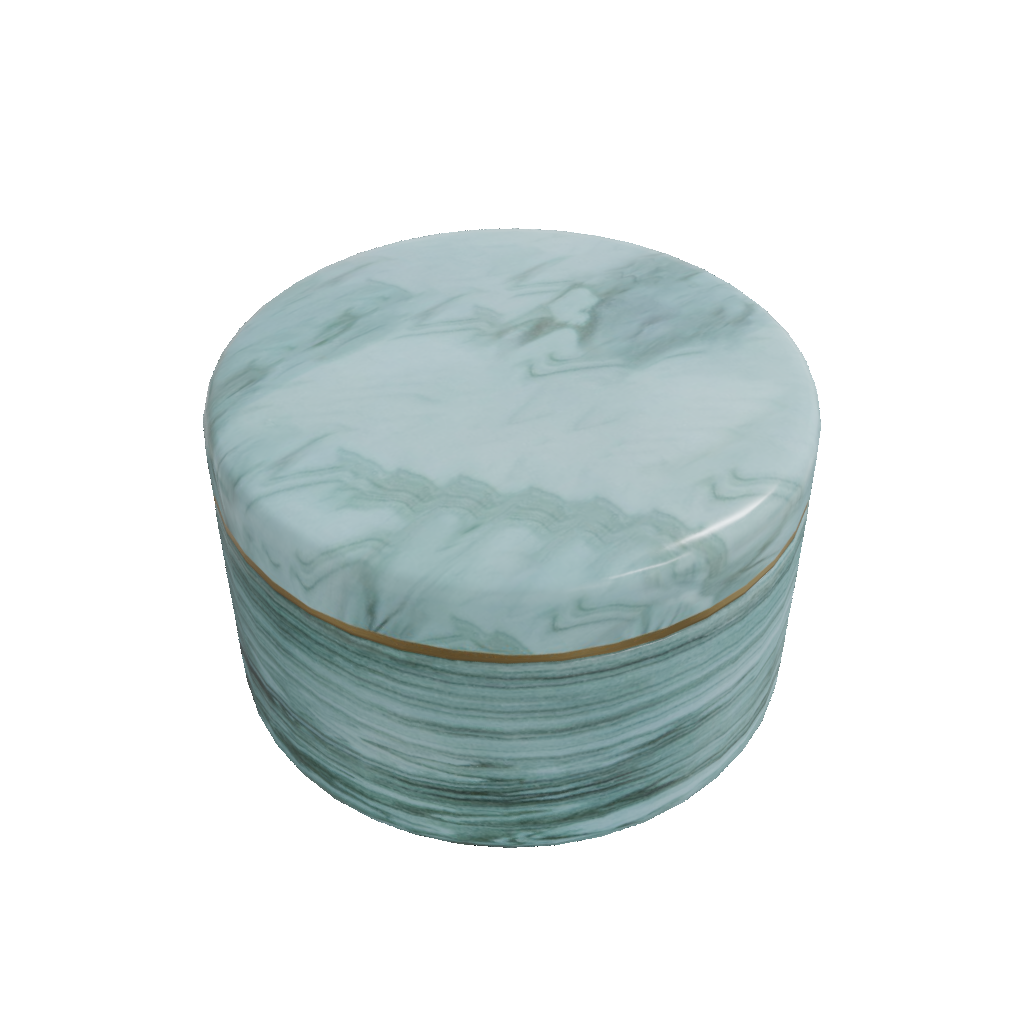}}
        }&
        \tikz{
        \node[draw=white, line width=0mm, inner sep=0pt] 
        {\includegraphics[width=.17\linewidth, trim={4cm 4cm 4cm 4cm}, clip]{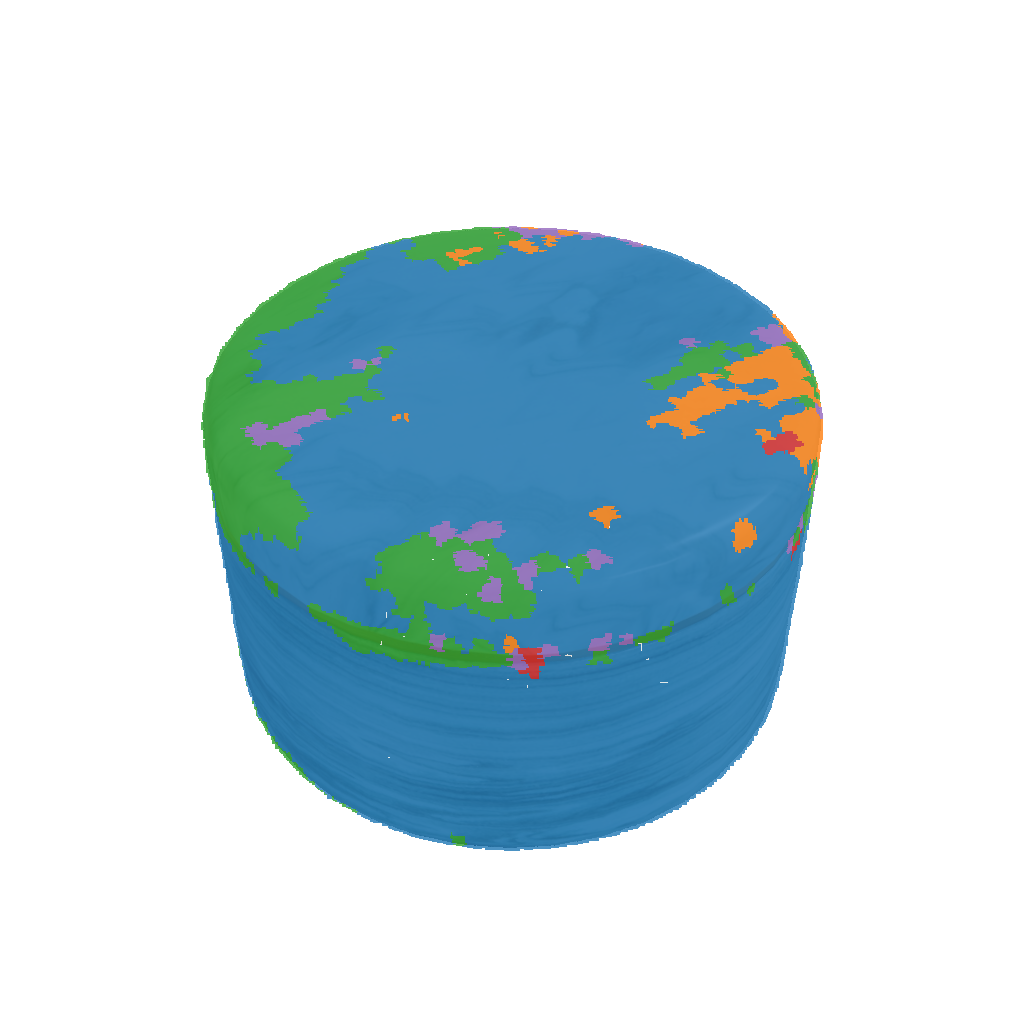} }
        }&
         \tikz{
        \node[draw=white, line width=0mm, inner sep=0pt] 
        {\includegraphics[width=.14\linewidth, clip]{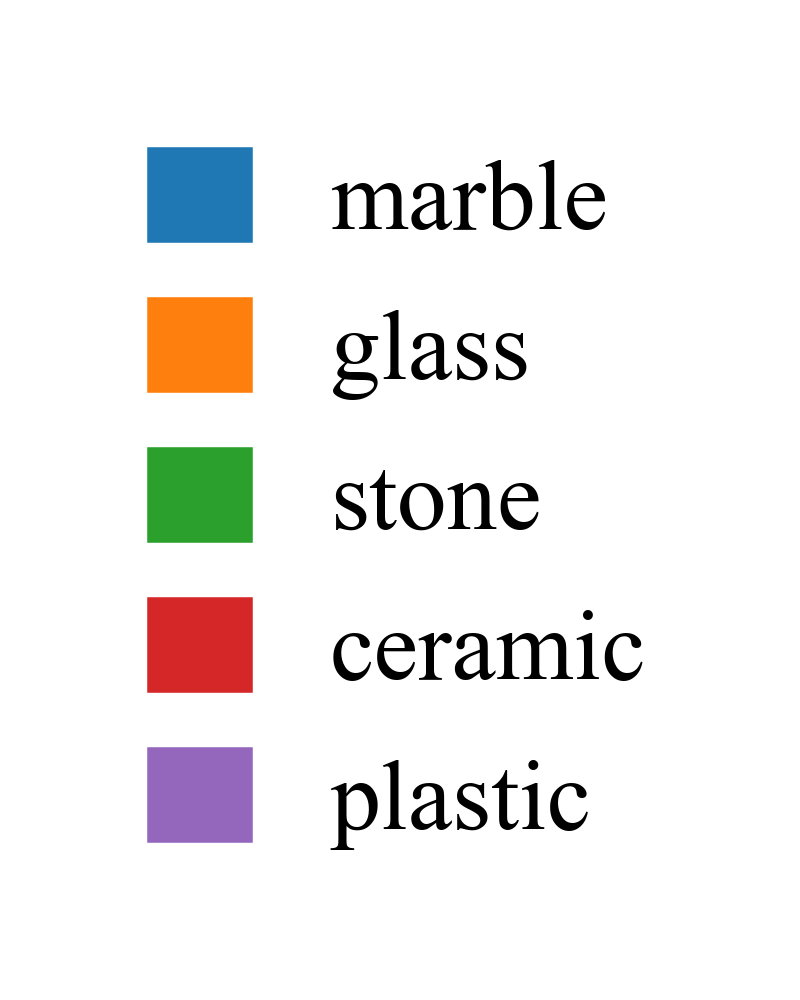}}
        }&
         \tikz{
        \node[draw=white, line width=0mm, inner sep=0pt] 
        {\includegraphics[width=.17\linewidth, trim={4cm 4cm 4cm 4cm}, clip]{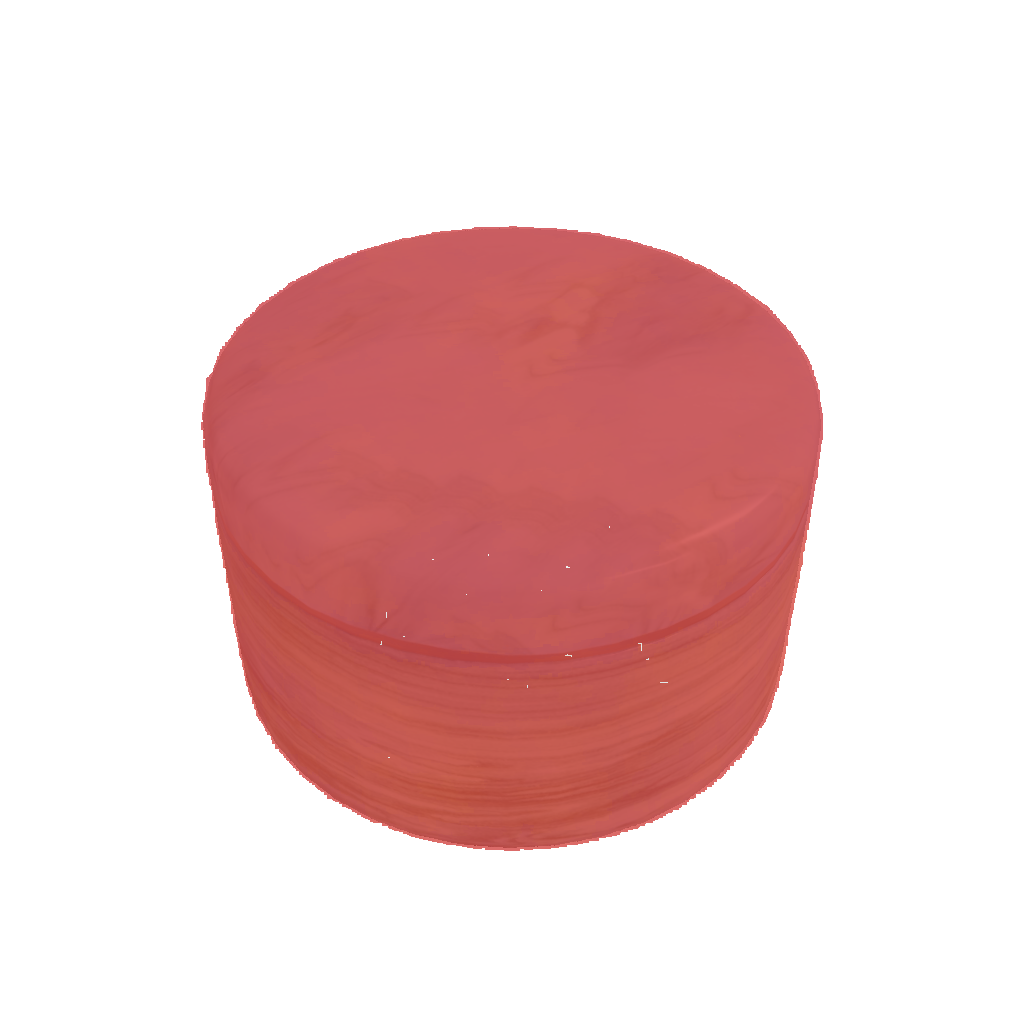}}
        }&
        \tikz{
        \node[draw=white, line width=0mm, inner sep=0pt] 
        { \includegraphics[width=.17\linewidth, trim={4cm 4cm 4cm 4cm}, clip]{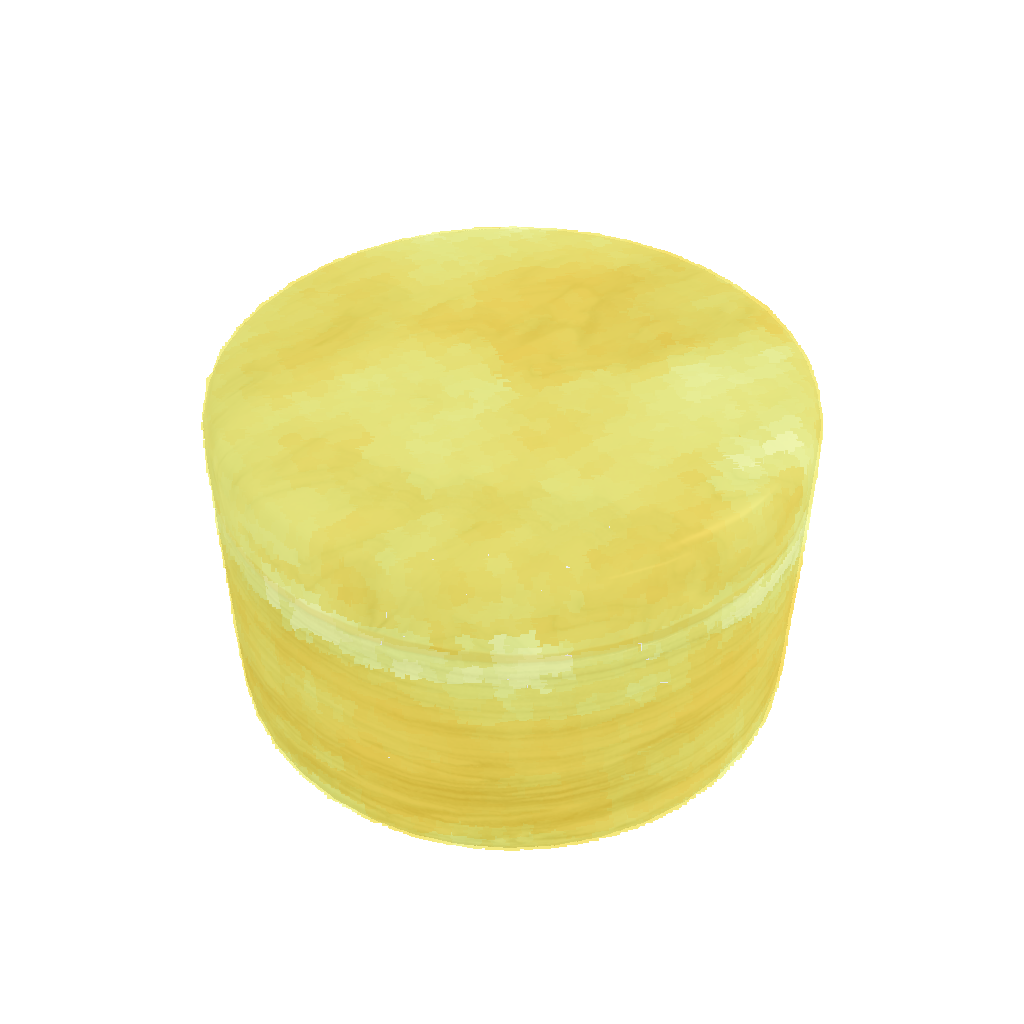}}
        }&
        \tikz{
        \node[draw=white, line width=0mm, inner sep=0pt] 
        { \includegraphics[width=.17\linewidth, trim={4cm 4cm 4cm 4cm}, clip]{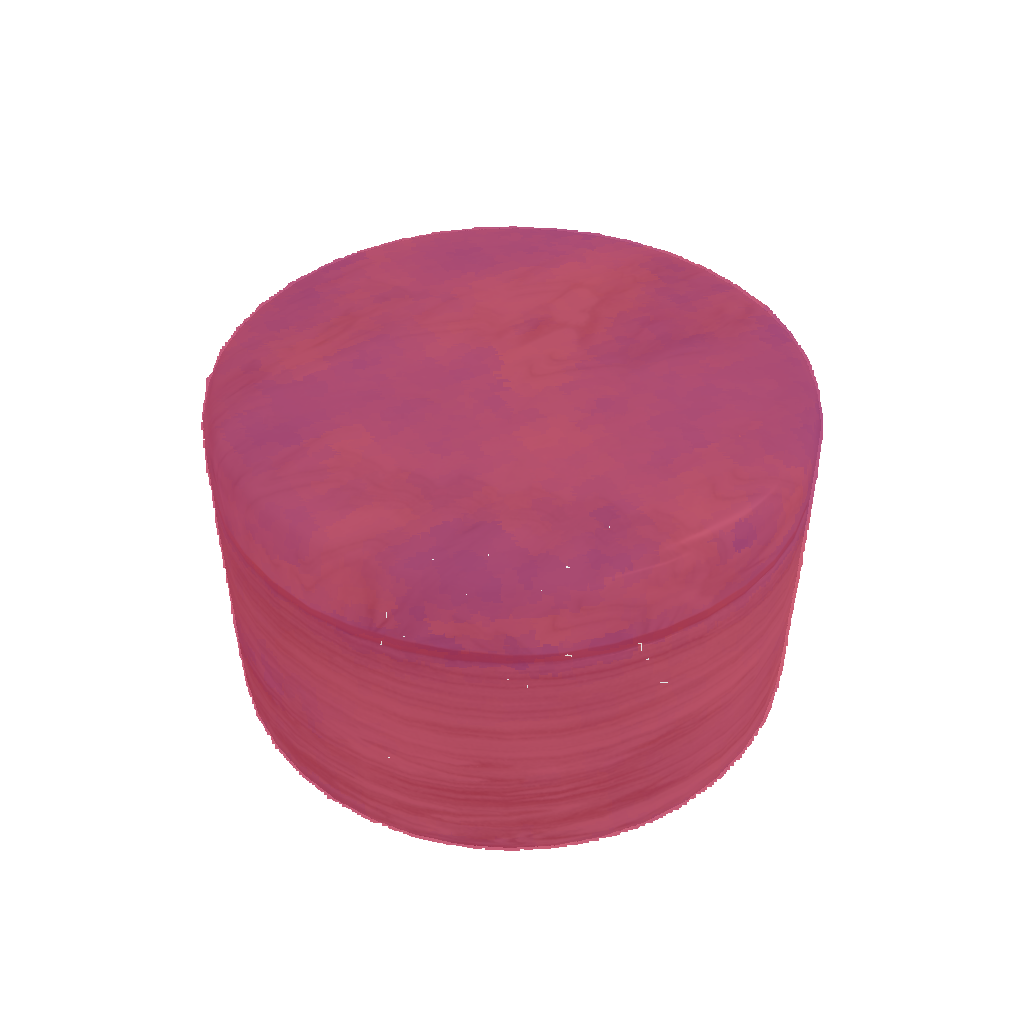}}
        }
        \\

        &
        &
        &
        \includegraphics[width=.21\linewidth, trim={0 0 0 9cm}, clip]{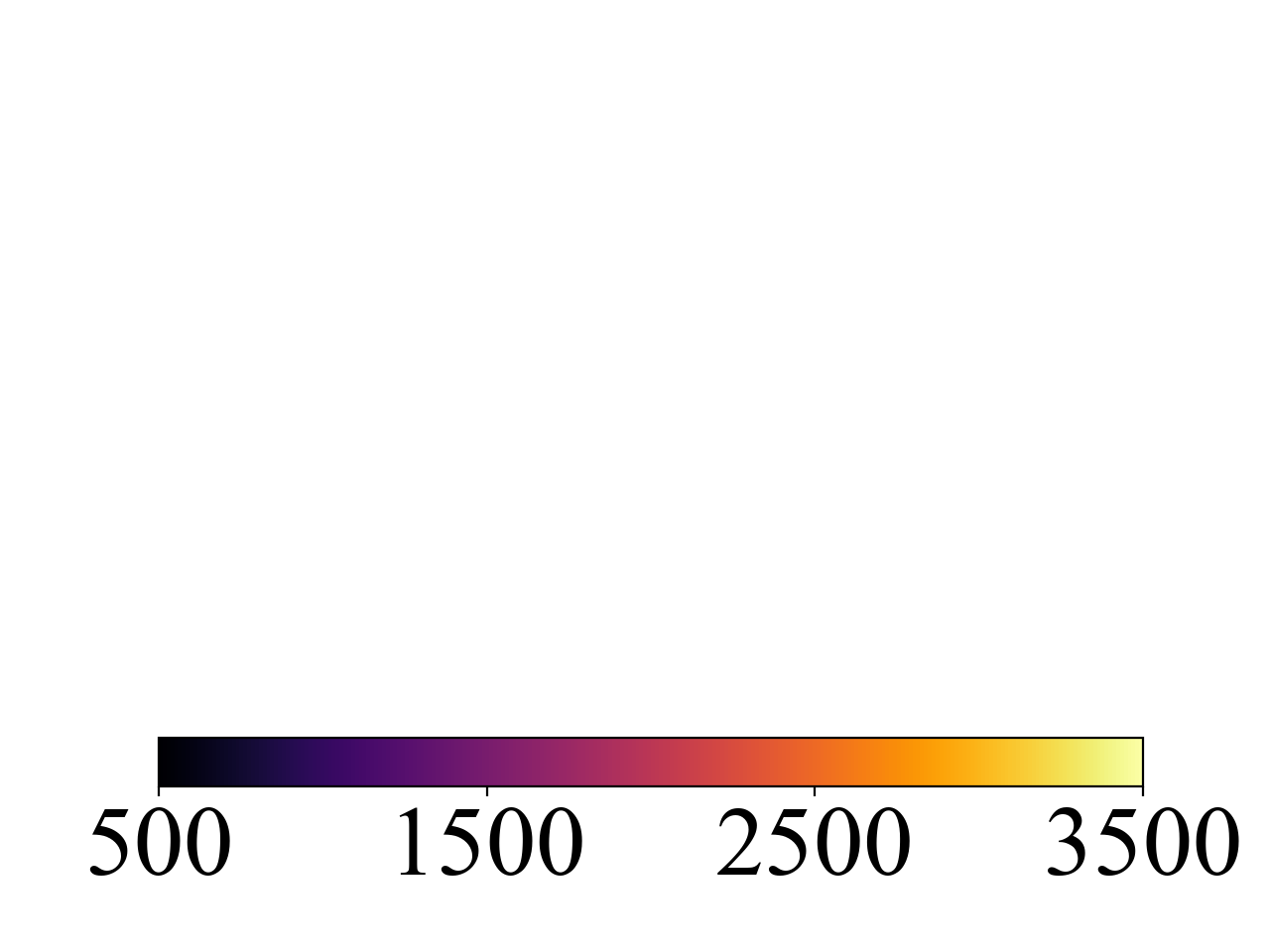}
        &
        \includegraphics[width=.21\linewidth, trim={0 0 0 9cm}, clip]{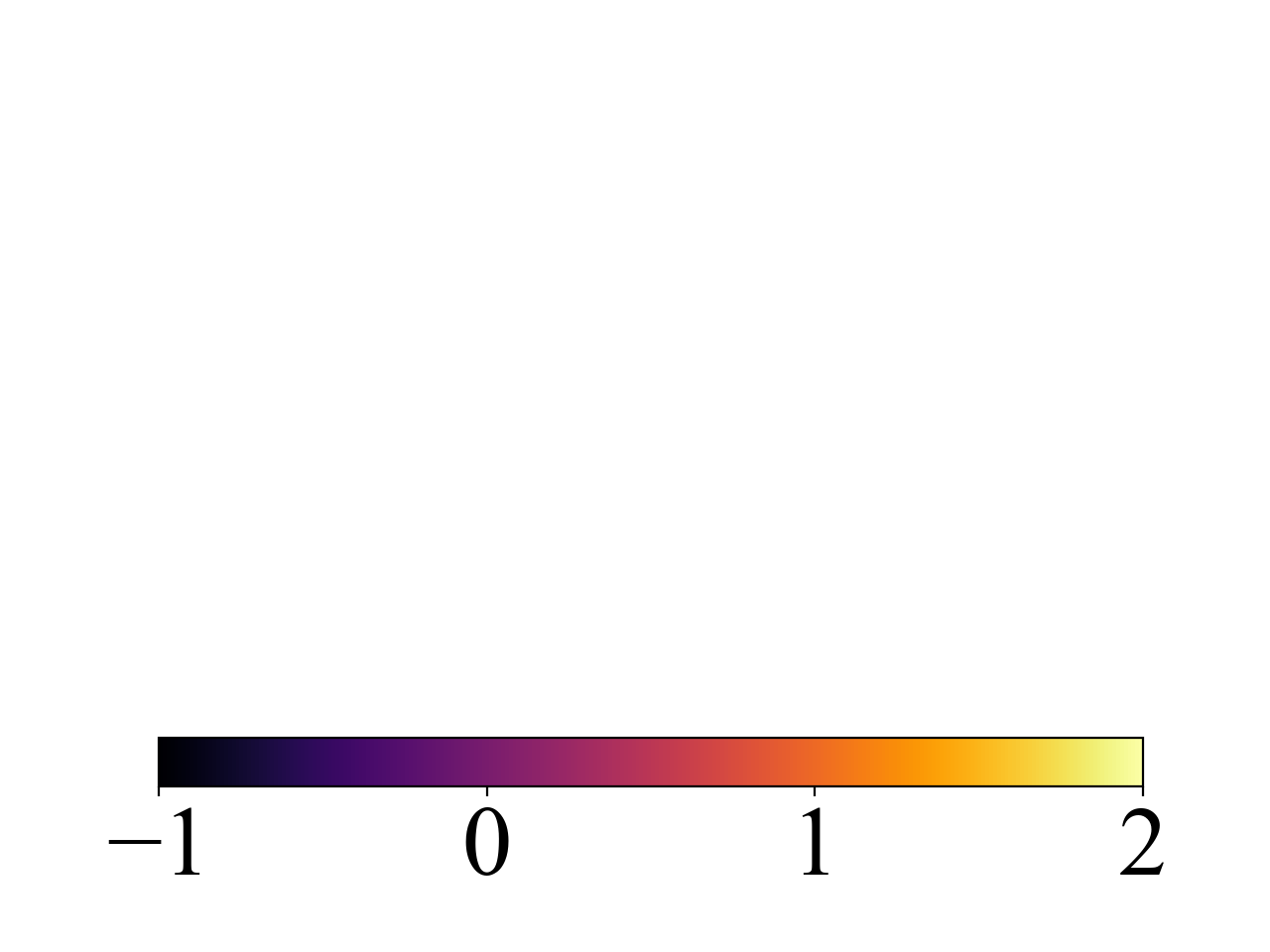}
        &
        \includegraphics[width=.21\linewidth, trim={0 0 0 9cm}, clip]{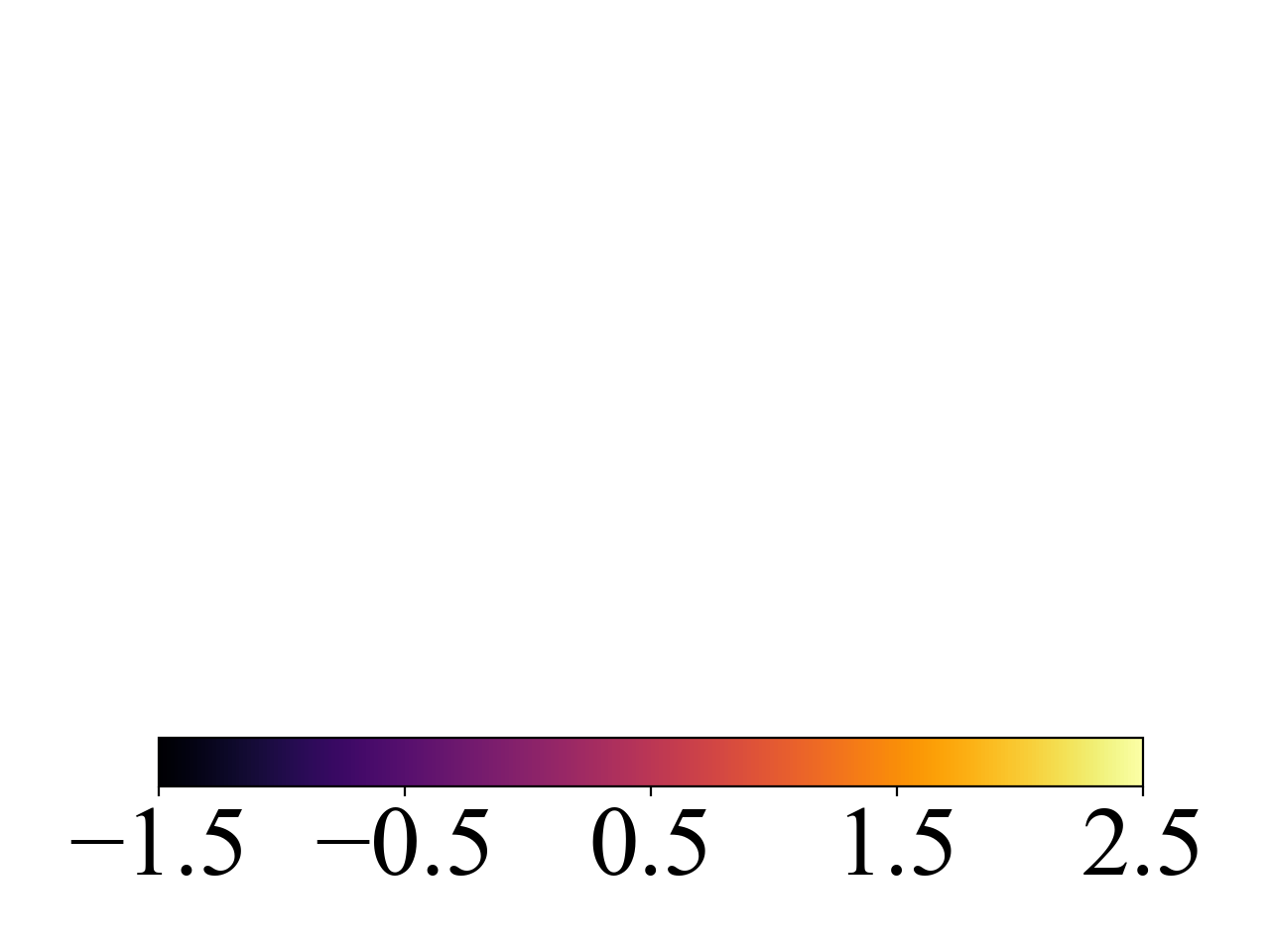}
    
    \end{tabular}
    }
    
\captionof{figure}{\textbf{Example predictions of different physical properties.}  We visualize more zero-shot predictions of mass density of objects from ABO-500, along with predictions of Young's modulus fields and thermal conductivity fields. Our method produces accurate predictions across a wide variety of objects and materials. }
\label{fig:supp_phys_props}
\end{table*}

\begin{figure}[h]
\begin{tcolorbox}[colback=white]
\footnotesize
\textbf{System:} You will be provided with captions that each describe an image of an object, along with a set of possible materials used to make the object. For each material, estimate the Young's modulus (in GPa) of that material in the object. You may provide a range of values for the Young's modulus instead of a single value.\\

Format Requirement:\\
You must provide your answer as a list of 5 (material: Young's modulus) pairs, each separated by a semi-colon (;). Do not include any other text in your answer, as it will be parsed by a code script later. Your answer must look like:\\
(material 1: low-high GPa);(material 2: low-high GPa);(material 3: low-high GPa);(material 4: low-high GPa);(material 5: low-high GPa)
\end{tcolorbox}

\caption{Prompt used for estimating Young's modulus.}
\label{fig:ym_prompt}
\end{figure}

\begin{figure}[h]
\begin{tcolorbox}[colback=white]
\footnotesize
\textbf{System:} You will be provided with captions that each describe an image of an object, along with a set of possible materials used to make the object. For each material, estimate the thermal conductivity (in W/mK) of that material in the object. You may provide a range of values for the thermal conductivity instead of a single value.\\

Format Requirement:\\
You must provide your answer as a list of 5 (material: thermal conductivity) pairs, each separated by a semi-colon (;). Do not include any other text in your answer, as it will be parsed by a code script later. Your answer must look like:\\
(material 1: low-high W/mK);(material 2: low-high W/mK);(material 3: low-high W/mK);(material 4: low-high W/mK);(material 5: low-high W/mK)
\end{tcolorbox}

\caption{Prompt used for estimating thermal conductivity.}
\label{fig:tc_prompt}
\end{figure}

\subsection{Failure Cases} 
We show example failure cases of NeRF2Physics in Fig.~\ref{fig:failure_cases}, covering the two main failure modes. In the case of the wooden box, BLIP-2 mistakes the object for a brick, causing GPT to output erroneous materials such as sand and concrete. With no correct materials in the dictionary, the CLIP-based regression is unable to produce accurate predictions. One direction for future work could be to implement a more robust view selection strategy to avoid such recognition failures. 

In the case of the black cart, the caption and materials are correct, but the CLIP-based regression mistakes the bulk of the cart as steel instead of plastic. This occurs because the local appearances of black-painted steel and black-painted plastic can look identical, and the patch-based CLIP features do not contain enough global information to accurately distinguish between them.
\begin{table*}[!t]
    \centering
    \resizebox{\linewidth}{!}{
\setlength{\tabcolsep}{0.2em} %
\renewcommand{\arraystretch}{1.}
    \begin{tabular}{cccccc}
    
    \makecell{Input RGB \\ (Display View)} &
    \makecell{Input RGB \\ (Canonical View)} &
    BLIP-2 Caption &
    \multicolumn{2}{c}{Material Segmentation} &
    Mass Density (kg/m$^3$) 
    \\

        \includegraphics[width=.17\linewidth, trim={4cm 5cm 4cm 3cm}, clip, align=c]{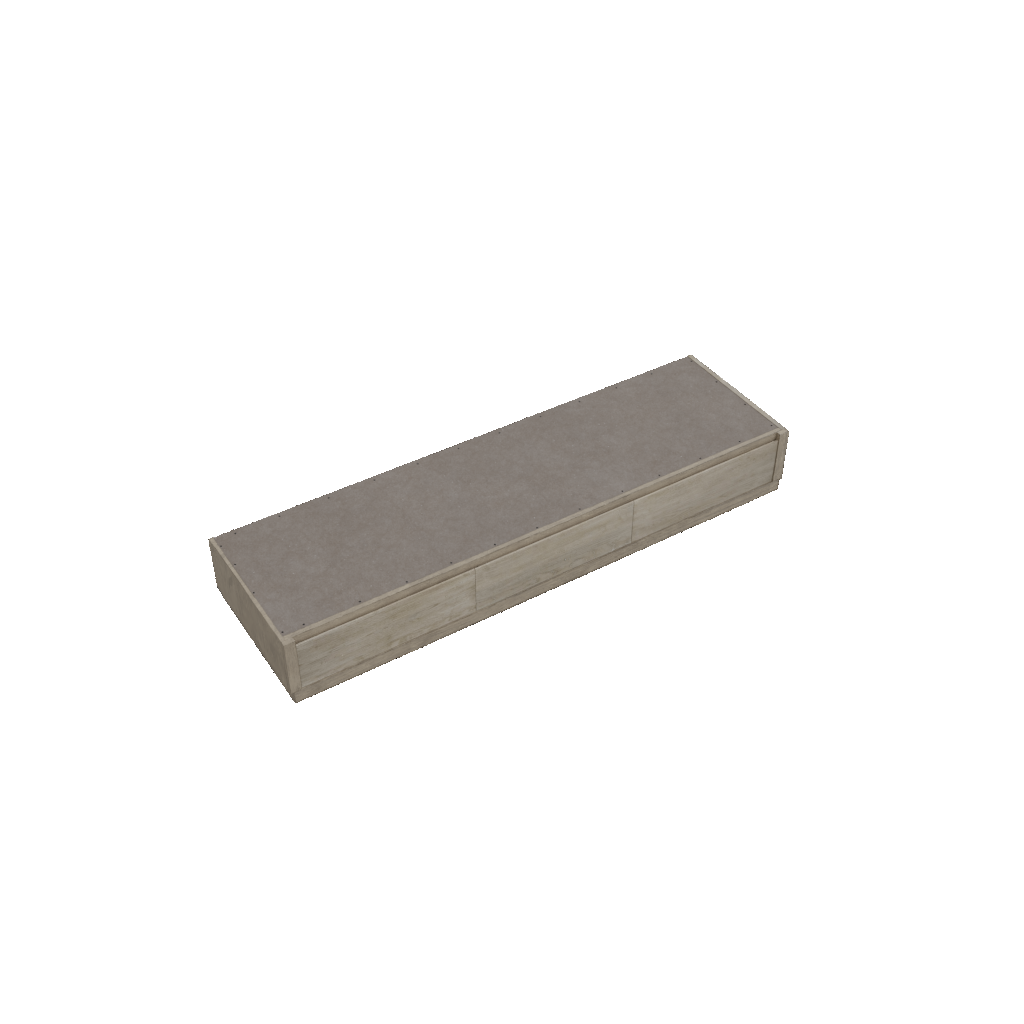}
        &
        \includegraphics[width=.17\linewidth, trim={4cm 5cm 4cm 3cm}, clip, align=c]{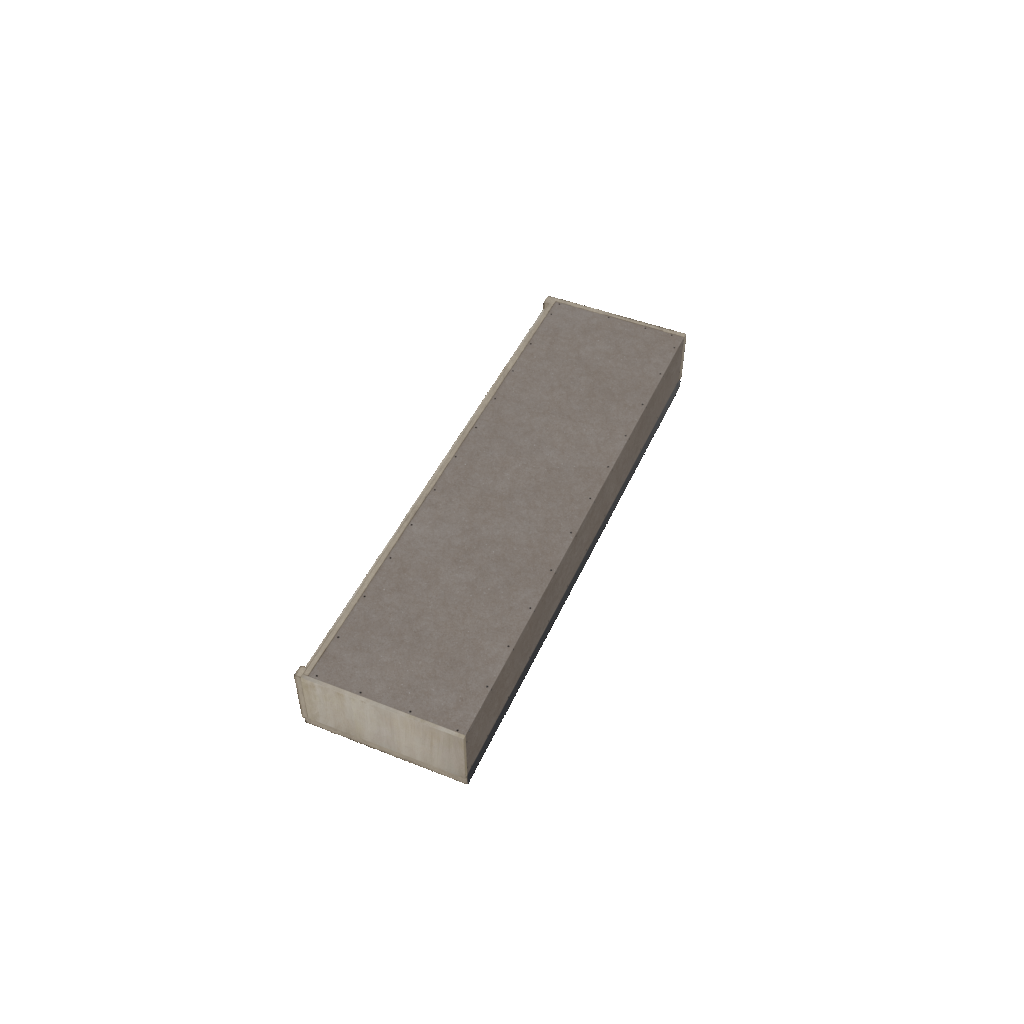} 
        &
        \makecell{
        \texttt{a grey brick on a} \\ \texttt{white background} 
        }
         &
        \includegraphics[width=.17\linewidth, trim={3cm 5cm 5cm 3cm}, clip, align=c]{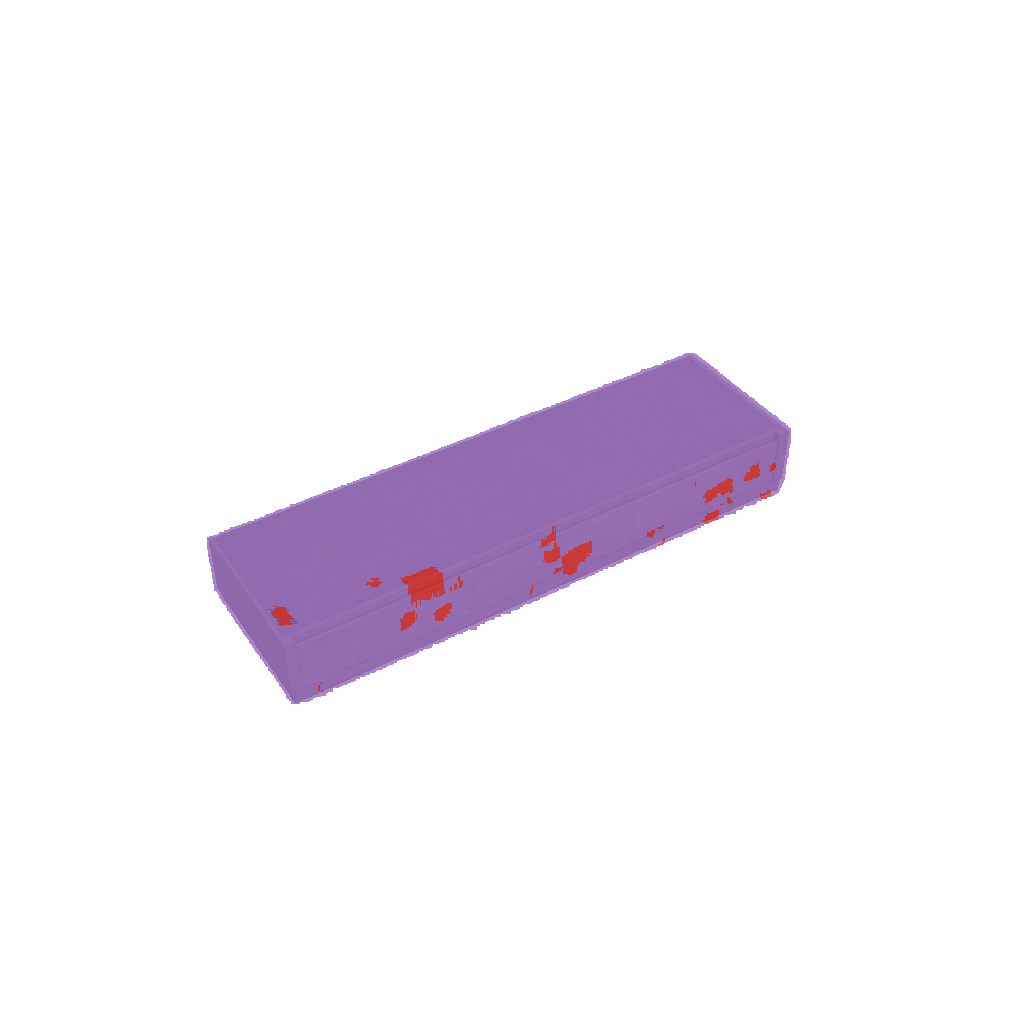}
        &
        \includegraphics[width=.14\linewidth, clip, align=c]{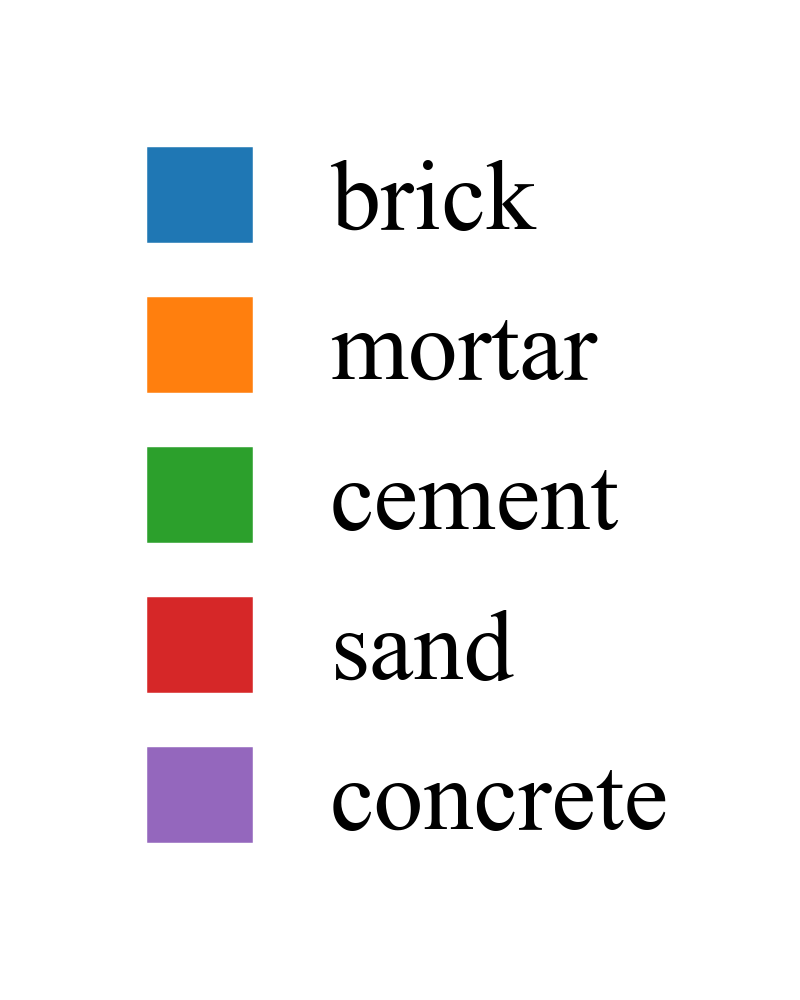}
        &
       \includegraphics[width=.17\linewidth, trim={4cm 5cm 4cm 3cm}, clip, align=c]{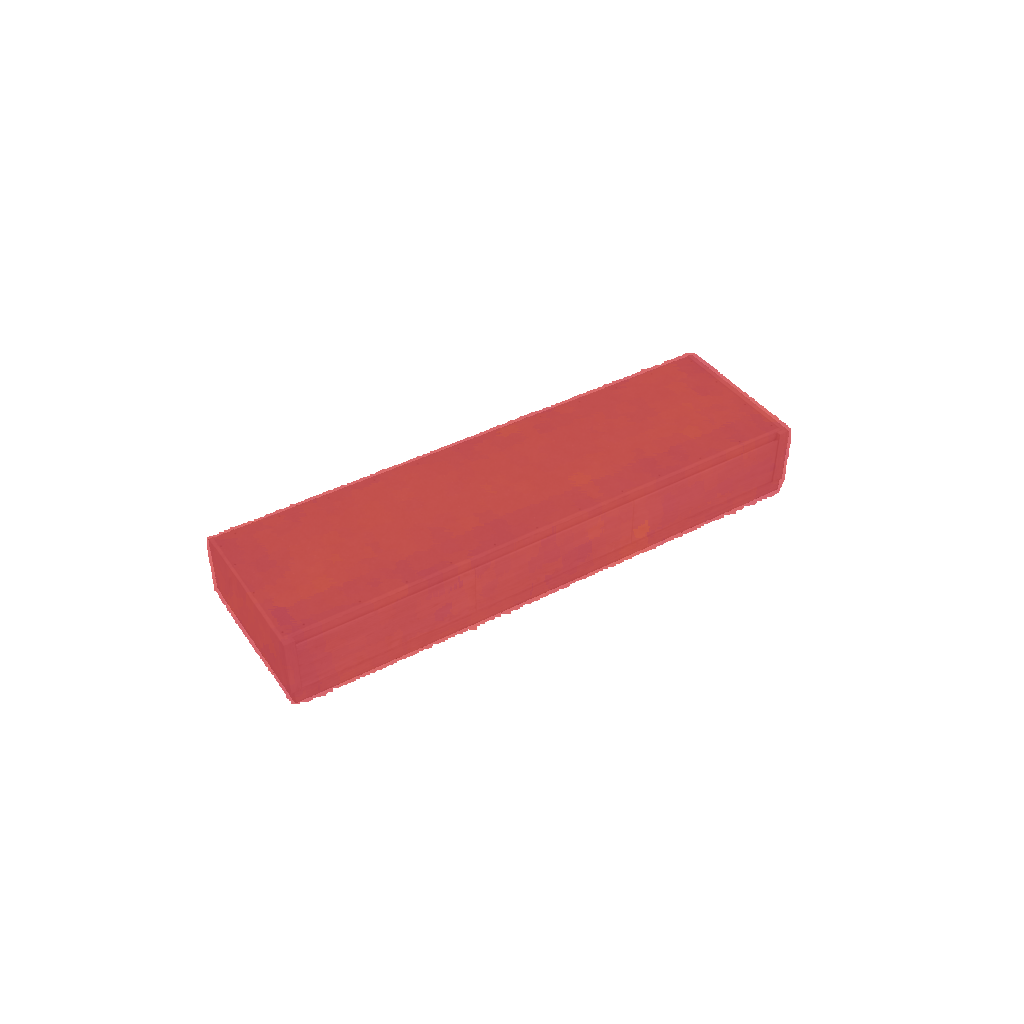}
        \\

        \includegraphics[width=.17\linewidth, trim={4cm 5cm 4cm 3cm}, clip, align=c]{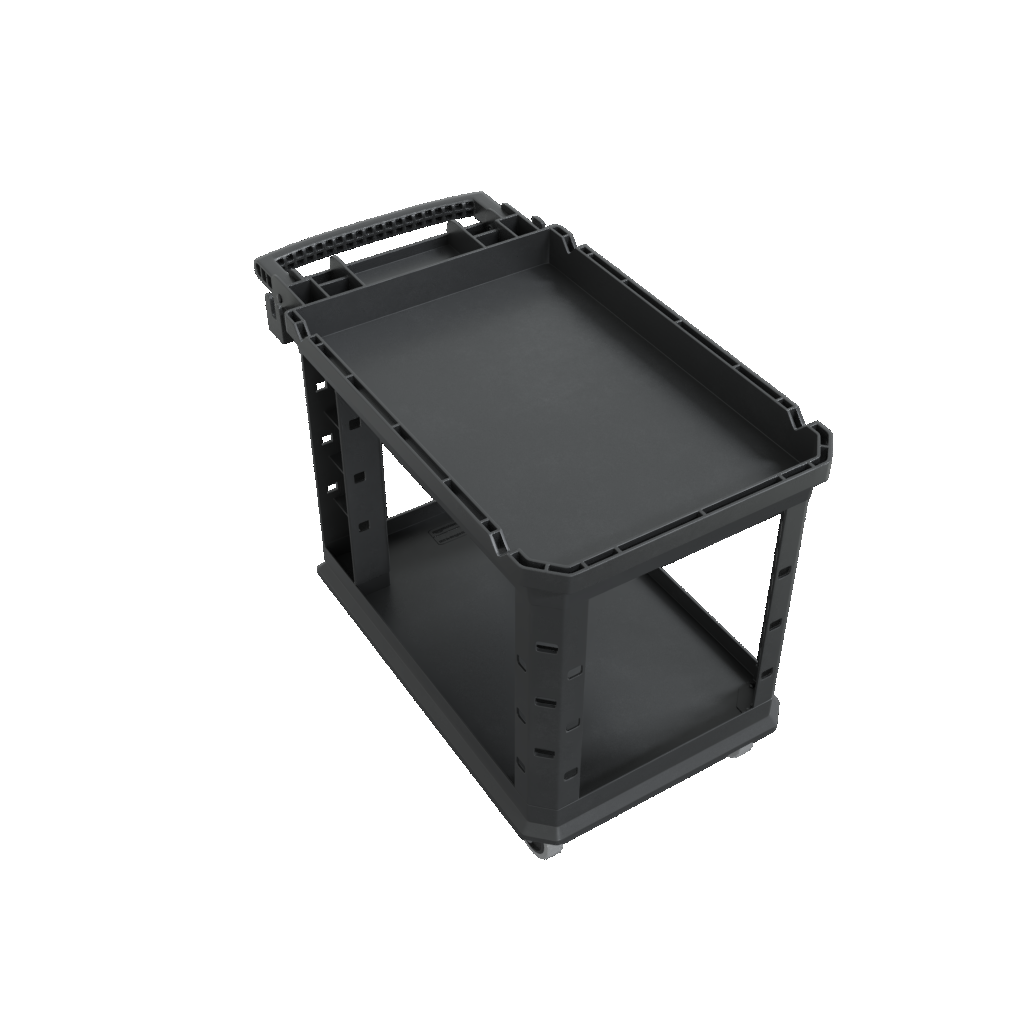}
        &
        \includegraphics[width=.17\linewidth, trim={4cm 5cm 4cm 3cm}, clip, align=c]{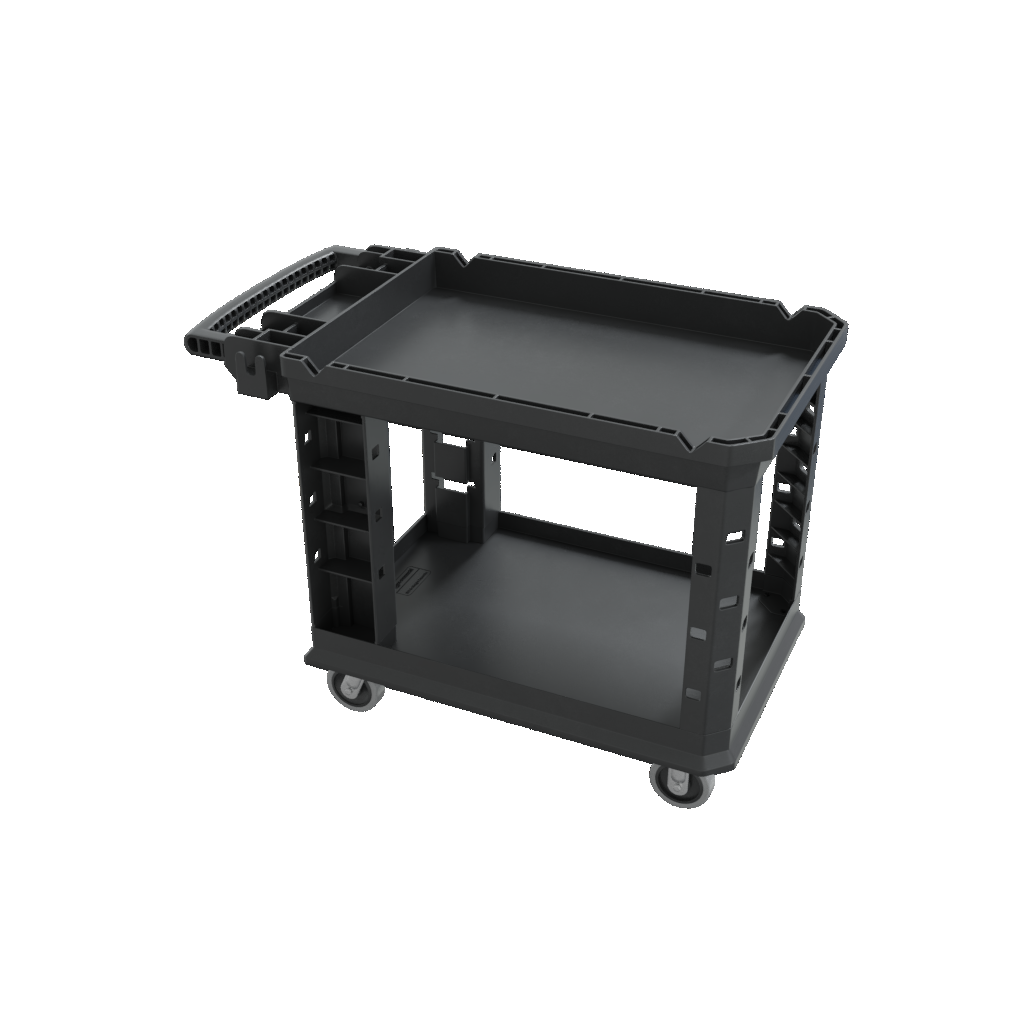} 
        &
        \makecell{
        \texttt{a black utility} \\ \texttt{ cart with two} \\\ \texttt{shelves} 
        }
         &
        \includegraphics[width=.17\linewidth, trim={3cm 5cm 5cm 3cm}, clip, align=c]{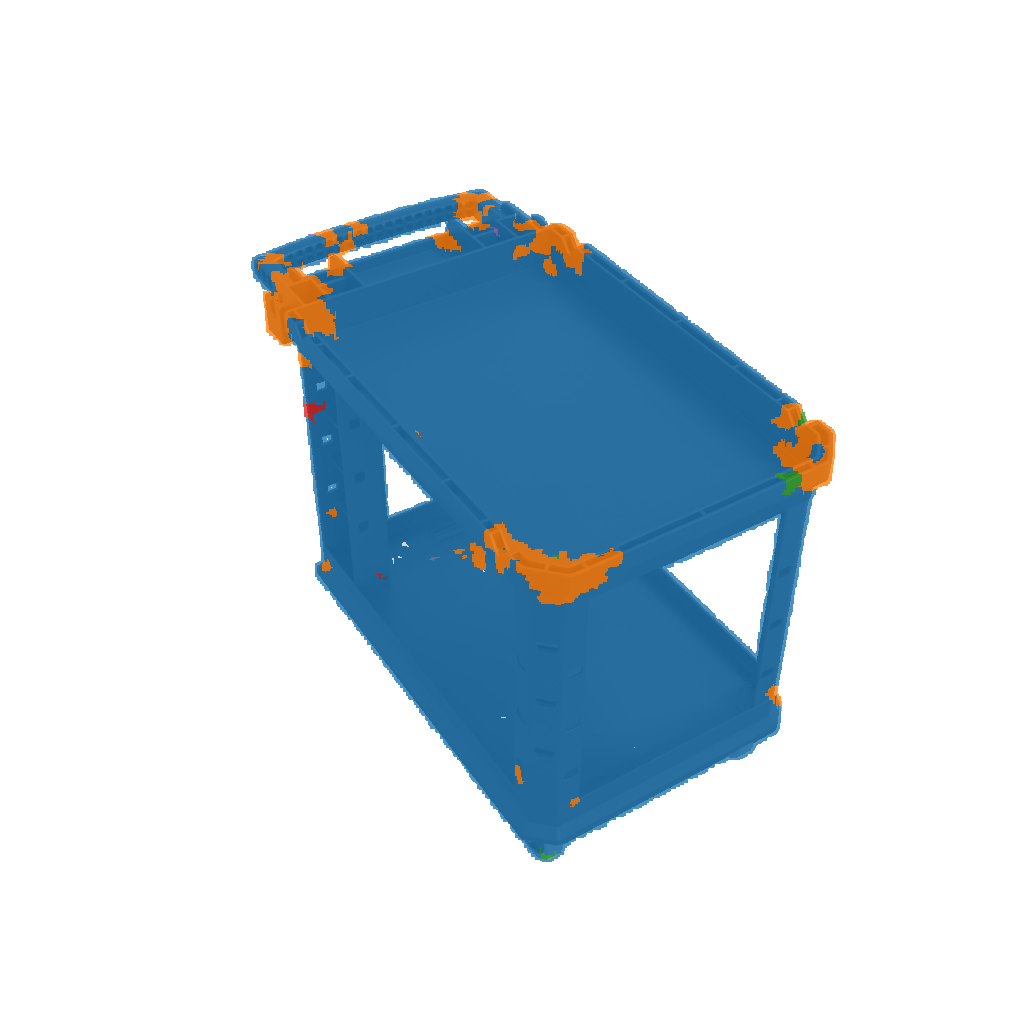}
        &
        \includegraphics[width=.14\linewidth, clip, align=c]{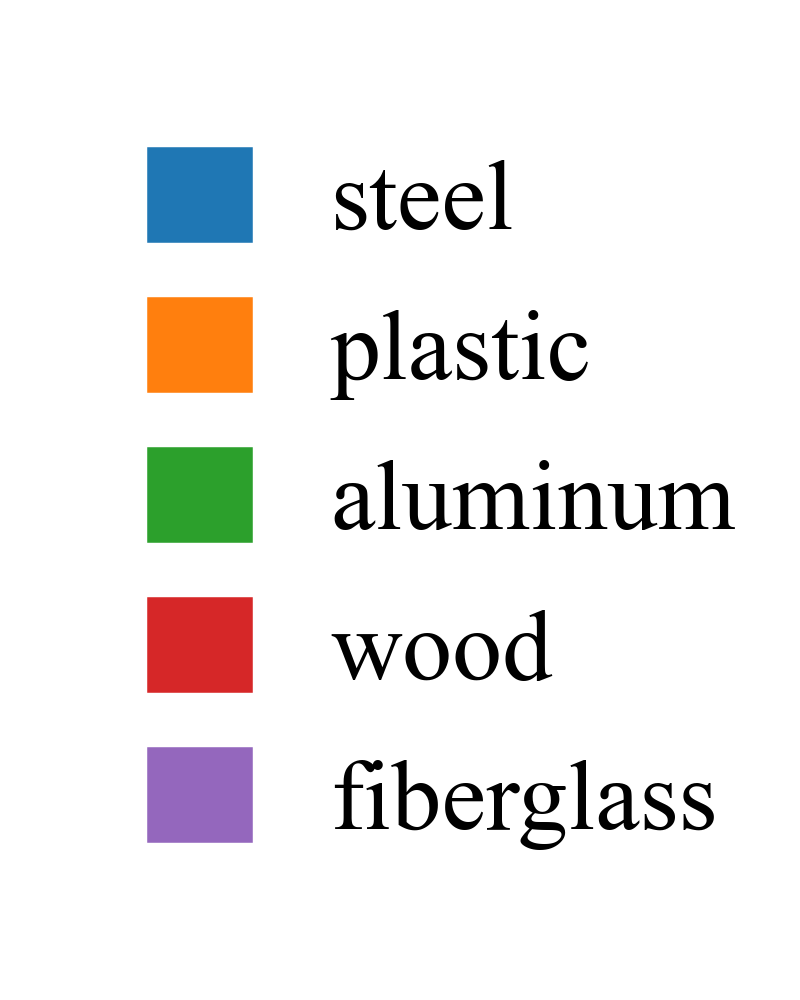}
        &
       \includegraphics[width=.17\linewidth, trim={4cm 5cm 4cm 3cm}, clip, align=c]{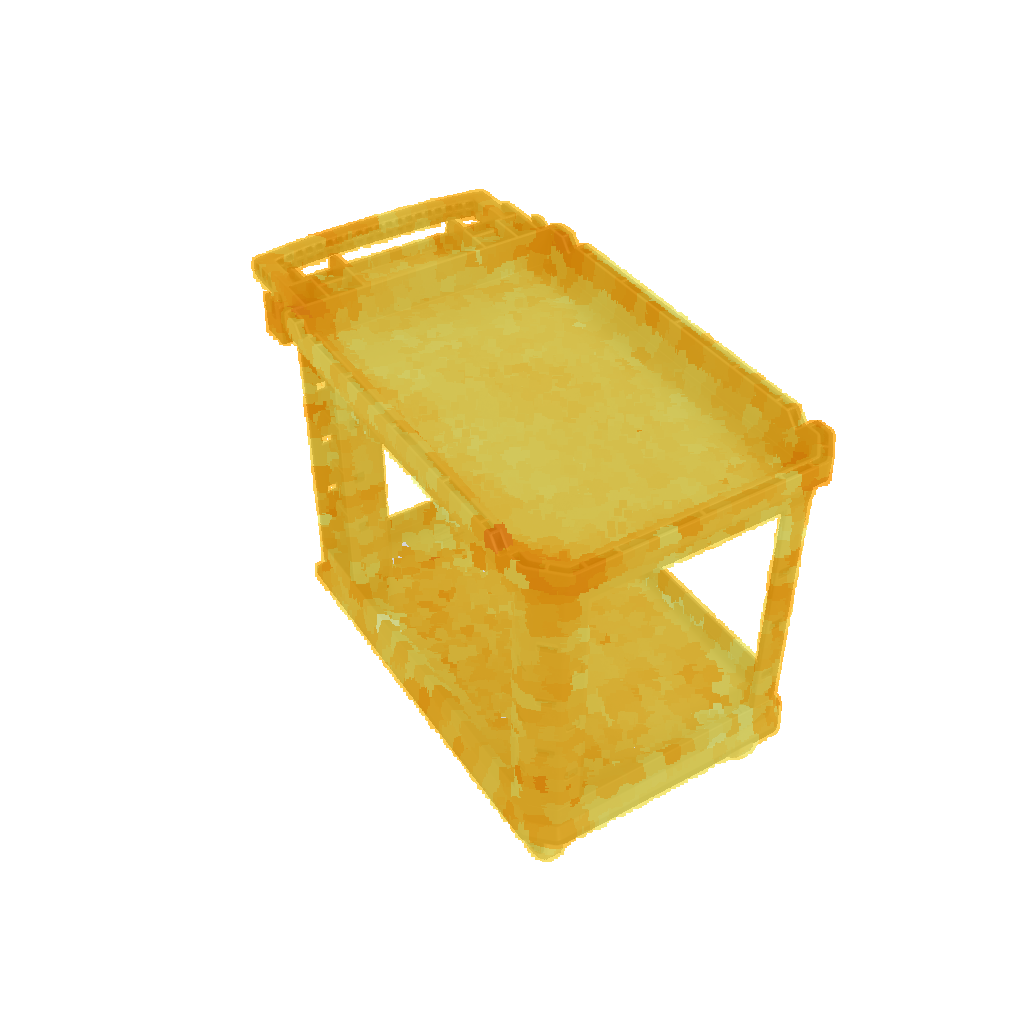}
        \\

        &
        &
        &
        &
        &
        \includegraphics[width=.21\linewidth, trim={0 0 0 9cm}, clip]{suppfig/supp_phys_props/density_cb.png}
    
    \end{tabular}
    }
    
\captionof{figure}{\textbf{Example failure cases.}  We visualize the main failure modes of NeRF2Physics in the above examples. The first example demonstrates object recognition failure at the captioning stage, and the second example demonstrates material recognition failure at the CLIP-based retrieval stage.}
\label{fig:failure_cases}
\end{table*}

\end{document}